\PassOptionsToPackage{unicode}{hyperref}
\PassOptionsToPackage{hyphens}{url}
\documentclass{article}

\usepackage{iftex}
\ifPDFTeX
  \usepackage[T1]{fontenc}
  \usepackage[utf8]{inputenc}
  \usepackage{textcomp}
  \usepackage{mathpazo}    
\else
  \usepackage{unicode-math}
  \defaultfontfeatures{Scale=MatchLowercase}
  \defaultfontfeatures[\rmfamily]{Ligatures=TeX,Scale=1}
\fi

\usepackage[margin=2.5cm]{geometry}
\usepackage{microtype}
\usepackage{setspace}
\usepackage[dvipsnames]{xcolor}
\usepackage{graphicx}
\usepackage{pdflscape}

\usepackage{booktabs,array,longtable}
\usepackage{float}
\usepackage{subcaption}

\usepackage[maxbibnames=15,
           minnames=15,
           maxcitenames=2,
           mincitenames=1,
           sorting=none 
           ]{biblatex}
\defbibheading{bibliography}{\section*{References}}

\usepackage{mdframed}
\usepackage{tcolorbox}
\tcbuselibrary{skins, breakable}

\usepackage{mdframed}
\usepackage{tcolorbox}
\tcbuselibrary{skins, breakable}

\newtcolorbox[auto counter]{namedbox}[3][]{%
    enhanced,
    oversize,
    breakable,
    colback=gray!5,
    colframe=gray!50,
    fonttitle=\bfseries,
    title={Box~\thetcbcounter: #3},
    label={box-#2},
    every float=\centering,
    boxsep=5pt,
    #1
}

\usepackage{adjustbox}

\let\oldsection\section
\renewcommand\section{\clearpage\oldsection}

\usepackage{hyperref}
\hypersetup{
    colorlinks=true,
    linkcolor=gray, 
    filecolor=Mahogany,
    urlcolor=gray, 
    citecolor=OliveGreen,
    pdftitle={NeuroAI for AI safety},
    hidelinks,
    pdfcreator={LaTeX via pandoc}
}

\DeclareGraphicsExtensions{.png,.eps}

\usepackage{enumitem}
\usepackage{authblk}
\usepackage{amsmath}
\usepackage[utf8]{inputenc}
\usepackage{textgreek}

\title{NeuroAI for AI Safety}

\author[1]{Patrick Mineault\textsuperscript{†}}
\author[1]{Niccolò Zanichelli\textsuperscript{*}}
\author[1,2,3]{Joanne Zichen Peng\textsuperscript{*}}
\author[4]{Anton Arkhipov}
\author[5]{Eli Bingham}
\author[6]{Julian Jara-Ettinger}
\author[5]{Emily Mackevicius}
\author[7]{Adam Marblestone}
\author[8]{Marcelo Mattar}
\author[9]{Andrew Payne}
\author[10]{Sophia Sanborn}
\author[5]{Karen Schroeder}
\author[5]{Zenna Tavares}
\author[10]{Andreas Tolias}
\author[11]{Anthony Zador}

\affil[1]{Amaranth Foundation}
\affil[2]{Princeton University}
\affil[3]{MIT}
\affil[4]{Allen Institute}
\affil[5]{Basis}
\affil[6]{Yale University}
\affil[7]{Convergent Research}
\affil[8]{NYU}
\affil[9]{E11 Bio}
\affil[10]{Stanford University}
\affil[11]{Cold Spring Harbor Laboratory}

\date{}

\usepackage{subfiles} 

\begin{document}
\maketitle

\def\thefootnote{*}\footnotetext{These authors contributed equally}
\def\thefootnote{†}\footnotetext{Corresponding author: patrick@amaranth.foundation}

\begin{abstract}
As AI systems become increasingly powerful, the need for safe AI has become more pressing. Humans are an attractive model for AI safety: as the only known agents capable of general intelligence, they perform robustly even under conditions that deviate significantly from prior experiences, explore the world safely, understand pragmatics, and can cooperate to meet their intrinsic goals. Intelligence, when coupled with cooperation and safety mechanisms, can drive sustained progress and well-being. These properties are a function of the architecture of the brain and the learning algorithms it implements. Neuroscience may thus hold important keys to technical AI safety that are currently underexplored and underutilized. In this roadmap, we highlight and critically evaluate several paths toward AI safety inspired by neuroscience: emulating the brain's representations, information processing, and architecture; building robust sensory and motor systems from imitating brain data and bodies; fine-tuning AI systems on brain data; advancing interpretability using neuroscience methods; and scaling up cognitively-inspired architectures. We make several concrete recommendations for how neuroscience can positively impact AI safety.
\end{abstract}

\hypertarget{introduction}{%
\section{Introduction}\label{introduction}}

AI systems have made remarkable advances in fields as diverse as
game-playing
\cite{Mnih2013-yq,Silver2016-jf,Silver2018-qz}, vision
\cite{Krizhevsky2012-uz,Dosovitskiy2020-yt}, autonomous
driving \cite{Grigorescu2019-sn}, medicine
\cite{Rajpurkar2022-wm}, protein folding
\cite{Jumper2021-gl,Watson2022-dp}, natural language
processing and dialogue
\cite{Brown2020-bk,Bubeck2023-pz}, mathematics
\cite{Trinh2024-ho}, weather forecasting
\cite{Lam2023-zc,Kochkov2024-qv} and materials science
\cite{Pyzer-Knapp2022-hy}. The advent of powerful AI has
raised concerns about AI risks, ranging from issues with today's systems
like climate impact
\cite{Rolnick2019-xd,Luccioni2023-yd}, systematic bias
\cite{Zhang2018-te,Bender2021-ew}, and mass surveillance
\cite{Peterson2022-dh}, to future-looking issues like
the misuse of powerful AI by malicious agents
\cite{Urbina2022-gd}, accidents stemming from the
misspecification of objectives
\cite{Bostrom2014-gb,Amodei2016-zg}, catastrophic risk
from autonomous systems
\cite{Bengio2024-uh,Bengio2024-va}, and race dynamics
among AI companies promoting the release of unsafe agents
\cite{Hendrycks2023-mr}.

AI safety is a field of research aimed at developing AI systems that are
helpful to humanity, and not harmful. Because AI is a general-purpose
technology \cite{Eloundou2023-bi}, AI safety research is
fundamentally interdisciplinary, cutting across research in computer
science and machine learning, mathematics, psychology, economics, law,
and social science. \emph{Technical} AI safety is a subset of AI safety that focuses on finding technical solutions to safety problems, distinguishing it from other areas such as policy work \cite{Amodei2016-zg}. 

We can broadly categorize these
safety issues into two levels, depending on the capability of AI systems
and time horizon:

\begin{itemize}
\item
  \textbf{Immediate safety concerns from today's prosaic AI systems.}
  By prosaic AI systems, we mean non-autonomous systems with limited
  capacity. This includes widely deployed systems like large language
  models and image generators \cite{Ramesh2022-jg}, as
  well as more specialized systems used in medicine, policing, and
  military applications, among others. Bias, amplification of societal issues like algorithmic policing, unequal access,
  interference in the political process and climate impacts, and the misuse of generative AI for creating fake content are commonly
  cited short-term safety concerns
  \cite{Christian2021-vb}.\item
  \textbf{Long-term safety concerns from future agentic AI systems.} By
  agentic AI systems, we mean partially or fully autonomous systems with a broad
  range of capabilities. This may include future systems with physical
  embodiments, including robotics, autonomous vehicles, aerial drones
  and wet-lab autonomous scientists \cite{Rapp2024-qm},
  as well as purely digital agents such as virtual agents interacting in
  sandboxes \cite{Park2023-fl}, virtual research
  assistants \cite{Lala2023-ot}, virtual scientists
  \cite{Lu2024-pj}, and software engineering agents
  \cite{Jimenez2023-gf,Yang2024-by}. These future
  agents, far more capable than current systems, could be far more
  valuable for society. Yet they also present dual-use concerns,
  including malicious use by state and non-state agents, use in military
  applications, organizational risks, as well as the possibility of
  losing control over advanced agents which pursue objectives and goals that can be harmful for humanity
  \cite{Russell2019-ul}. Highly capable agentic AI systems are sometimes referred to as general-purpose artificial intelligence, or artificial general intelligence (AGI).
\end{itemize}

Although immediate safety is critical for AI to benefit society \cite{Richards2023-gb}, our primary focus here is on safety concerns related to future agentic AI. As stated in a recent report chaired by Yoshua Bengio \cite{Bengio2024-uh}: 

\begin{quote}
``The future of general-purpose AI technology is uncertain, with a wide
range of trajectories appearing possible even in the near future,
including both very positive and very negative outcomes. But nothing
about the future of AI is inevitable.''
\end{quote}

Long-term AI safety is a sufficiently important
societal problem \cite{Bengio2024-va} that it deserves multidisciplinary consideration, including from neuroscientists. In this roadmap, we aim to evaluate and draw a path for how inspiration, data, and tools from neuroscience can positively impact AI safety.

\hypertarget{what-can-a-neuroscientist-do-about-ai-safety}{%
\subsection{What can a neuroscientist do about AI
safety?}\label{what-can-a-neuroscientist-do-about-ai-safety}}

Animals navigate, explore, and exploit their environments while maintaining
self-preservation. Among these, mammals, birds, and cephalopods exhibit
particularly flexible perceptual, motor, and cognitive systems
\cite{Cisek2022-qc} that generalize well
to out-of-distribution inputs, meaning they can effectively handle situations or stimuli that differ significantly from what they have previously encountered. 

Humans have evolved additional capacities for
cooperation and complex social behavior
\cite{Bennett2023-rq}, organizing themselves into societies that promote prosocial conduct and discourage harmful actions.
These capacities emerge, in part, from the neural architecture of the
brain. Evolution has shaped the brain to impose strong constraints on human behavior in order to enable humans to learn from and participate in society. By understanding what those constraints are and how they are implemented, we may be able to transfer those lessons to AI systems. We could build systems
that exhibit familiar human-like intelligence, which we have experience
dealing with. It follows that studying the brain and understanding
the biological basis of this natural alignment is a promising route
toward AI safety.

\begin{figure}
    \centering
    \includegraphics[width=0.75\linewidth]{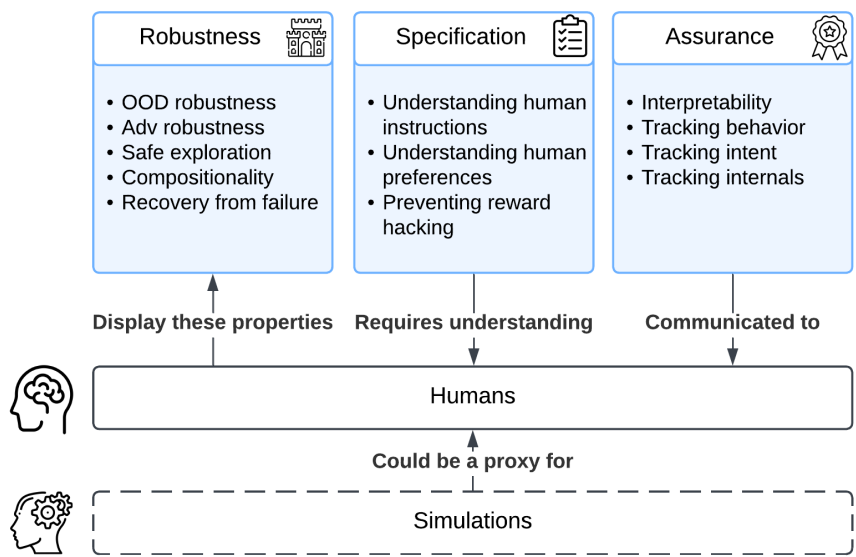}
    \caption{A framework for AI safety adapted to NeuroAI.}
    \label{fig-frameworks-for-ai-safety}
\end{figure}

We use the technical framework introduced by Deepmind in 2018
\cite{Ortega2018-bv} to more concretely categorize how studying the brain
could positively impact AI safety (Figure \ref{fig-frameworks-for-ai-safety}).

\begin{enumerate}
\def\labelenumi{\arabic{enumi}.}
\item
  \textbf{Robustness}: specifying how an agent can safely respond to
  unexpected inputs. This includes performing well or failing gracefully
  when faced with adversarial and out-of-distribution inputs, and safely exploring in
  unknown environments. This can also mean learning compositional
  representations that generalize well out-of-distribution. Robustness further implies knowing what you do not know, by maintaining a representation of uncertainty, to ensure safe and informed decision-making in novel or uncertain scenarios.
  
\item
  \textbf{Specification}: specifying the expected behavior of an AI
  agent. A pithy way of expressing this is that we want AI systems to
  ``do what we mean, not what we say''. This includes correctly
  interpreting instructions specified in natural language despite ambiguity; preventing learning shortcuts
  that generalize poorly \cite{Geirhos2020-uk}; ensuring
  that agents solve the real task at hand rather than engaging in reward
  hacking \cite{Amodei2016-zg} (i.e. Goodhart's law); and so on.

\item
  \textbf{Assurance (or oversight)}: being able to verify that AI
  systems are working as intended. This includes opening the black box
  of AI systems using interpretability methods; scalably overseeing the
  deployment of AI systems and detecting unusual or unsafe behavior; or
  detecting and correcting for bias.
\end{enumerate}

There have been a few examples where neuroscience has already positively impacted AI safety, which fit neatly into this framework. For example, interpretability methods inspired by neuroscientific approaches \cite{Lindsay2023-zr} are a form of \textbf{assurance}, while methods which seek inspiration from the brain to find solutions to adversarial attacks can enhance \textbf{robustness}
\cite{Goodfellow2014-pc,Elsayed2018-oe,Ilyas2019-ir,Guo2022-ef,Bartoldson2024-qr,Fort2024-cp, Li2019-dg, Dapello2020-af, Safarani2021-ui,Li2023-vn}. 

Throughout this roadmap, we'll encounter several more proposals which are aimed at solving specific technical issues within AI safety under one of these rubrics. Some proposals, however--for example, detailed biophysical simulations of the human brain or top-down simulations from representations learned from neural data--seek to benefit AI safety by emulating human minds and all their safety-relevant properties, thus affecting all the relevant rubrics: robustness, specification and assurance. We'll note them as primarily aiming to build simulations of the human mind in the following.

\begin{namedbox}{marrs-levels}{Marr's levels for AI safety}
At what level should we study the brain for the purpose of AI safety? The different proposals we evaluate make very different bets on which level of granularity should be the primary focus of study. Marr's levels \cite{Marr1982-xh} codify different levels of granularity in the study of the brain:

\begin{center}
\setlength{\fboxsep}{0pt}  
\setlength{\fboxrule}{0.5pt}  
    \centering
    \includegraphics[width=.9\textwidth]{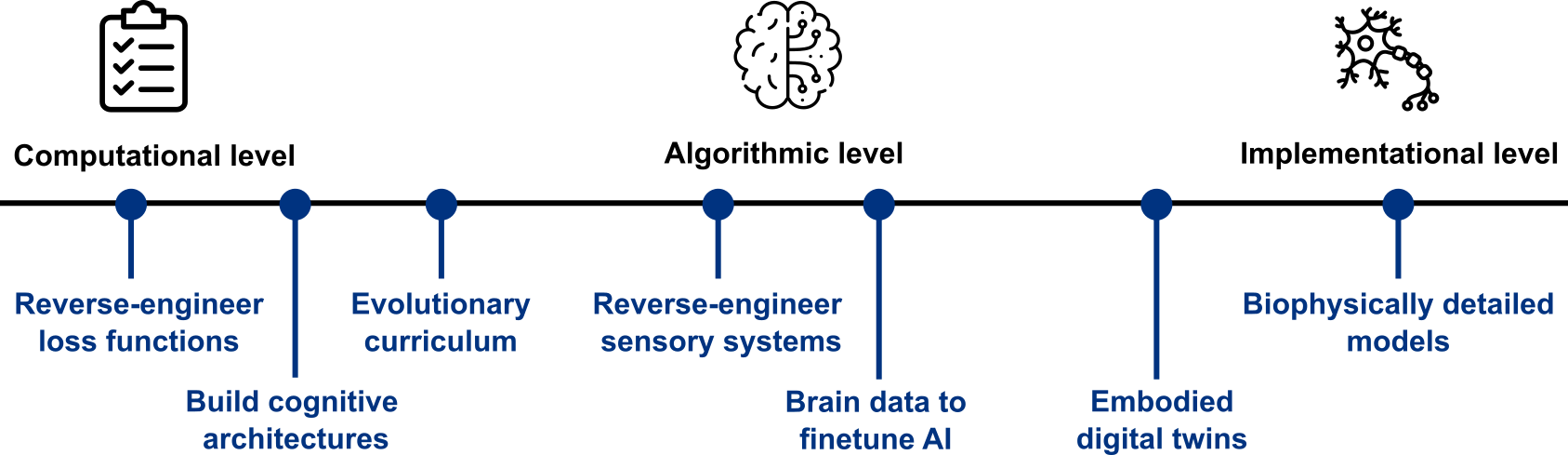}
    
\end{center}

\begin{enumerate}
    \item \textbf{The computational level}. What is the high-order task that the brain is trying to solve? Reverse engineering the loss functions of the brain, building better cognitive architectures, and building an evolutionary curriculum map onto this level.
    \item \textbf{The algorithmic (or representation) level}. What is the algorithm that the brain uses to solve that problem? Alternatively, what are the representations that the brain forms to solve that problem? Approaches including brain-informed process supervision and building digital twins of sensory systems map onto this level.
    \item \textbf{The implementation level}. How is this problem solved by the brain? Biophysically detailed whole-brain simulation falls into that category, while embodied digital twins straddle the algorithmic and implementation levels.
\end{enumerate}

For the purpose of AI safety, any one level is unlikely to be sufficient to fully solve the problem. For example, solving everything at the implementation level using biophysically detailed simulations is likely to be many years out, and computationally highly inefficient. On the other hand, it is very difficult to forecast which properties of the brain are truly critical in enhancing AI safety, and a strong bet on only the computational or algorithmic level may miss crucial details that drive robustness and other desirable properties. 

Thus, we advocate for a holistic strategy that bridges all of the relevant levels. Importantly, we focus on \emph{scalable approaches anchored in data}. All of these levels add constraints to the relevant problem, ultimately forming a safer system.

\end{namedbox}

\hypertarget{proposals-how-to-read}{%
\subsection{Proposals for neuroscience for AI safety}\label{subsec-proposals-how-to-read}}

There have been several proposals for how neuroscience can positively
impact AI safety. These span from emulating the brain's representations, information processing, and architecture; building robust sensory and motor systems by imitating brain and body structure, activity, and behavior; fine-tuning AI systems on brain data or learning loss functions from it; advancing interpretability using neuroscience methods; and scaling up cognitively-inspired architectures. We list them in Table \ref{tab-neuroscience-ai-safety}, along with which aspect of AI safety they propose to affect.

\newlength{\totalwidth}
\setlength{\totalwidth}{\textwidth}
\addtolength{\totalwidth}{.3in}  

\setlength\LTleft{-.5in}
\setlength\LTright{-.1in}
\begin{longtable}{p{0.28\totalwidth}p{0.57\totalwidth}p{0.15\totalwidth}}
\toprule
\textbf{Proposed method} & \textbf{Summary of proposition} & \textbf{Rubric} \\
\midrule
\endfirsthead

\multicolumn{3}{l}{\small\textit{Continued from previous page}} \\
\toprule
\textbf{Proposed method} & \textbf{Summary of proposition} & \textbf{Rubric} \\
\midrule
\endhead

\midrule
\multicolumn{3}{r}{\small\textit{Continues on next page}} \\
\endfoot

\bottomrule
\caption{Proposals for how neuroscience can impact AI safety}\label{tab-neuroscience-ai-safety} \\
\endlastfoot

\hyperref[sec-reverse-engineer-the-representations-of-sensory-systems]{Reverse-engineer sensory systems} & 
Build models of sensory systems (``sensory digital twins'') which display robustness, reverse engineer them through mechanistic interpretability, and implement these systems in AI & 
Robustness \\
\midrule

\hyperref[sec-embodied]{Build embodied digital twins} & 
Build simulations of brains and bodies by training auto-regressive models on brain activity measurements and behavior, and embody them in virtual environments & 
Simulation \\
\midrule

\hyperref[sec-wbs]{Build biophysically detailed models} & 
Build detailed simulations of brains via measurements of connectomes (structure) and neural activity (function) & 
Simulation \\
\midrule

\hyperref[sec-develop-better-cognitive-architectures]{Develop better cognitive architectures} & 
Build better cognitive architectures by scaling up existing Bayesian models of cognition through advances in probabilistic programming and foundation models& 
Simulation, Assurance \\
\midrule

\hyperref[sec-use-brain-data-to-finetune-ai-systems]{Use brain data to finetune AI} & 
Finetune AI systems through brain data; align the representational spaces of humans and machines to enable few-shot learning and better out-of-distribution generalization & 
Specification, Robustness \\
\midrule

\hyperref[sec-build-an-evolutionary-curriculum]{Build an evolutionary curriculum} & 
Build safety guardrails in AI by recapitulating the natural evolutionary curriculum & 
Specification \\
\midrule

\hyperref[sec-infer-the-loss-functions-of-the-brain]{Infer the brain's loss functions} & 
Learn the brain's loss and reward functions through a combination of techniques including task-driven neural networks, inverse reinforcement learning, and phylogenetic approaches & 
Specification \\
\midrule

\hyperref[sec-leverage-neuroscience-inspired-methods-for-mechanistic-interpretability]{Use neuroscience methods for interpretability} & 
Leverage methods from neuroscience to open black-box AI systems; bring methods from mechanistic interpretability back to neuroscience to enable a virtuous cycle & 
Assurance \\

\end{longtable}

Many of these proposals are in embryonic form, often in grey literature--whitepapers, blog posts, and short presentations
\cite{Bostrom2008-og,Byrnes2022-wm,Cvitkovic2023-lw,Thiergart2023-la}. Our goal here is to catalog these proposals and flesh them out, putting
the field on a more solid ground \cite{Olah2017-wj}.

\begin{figure}[htbp]
    \centering
    \includegraphics[width=6.5in]{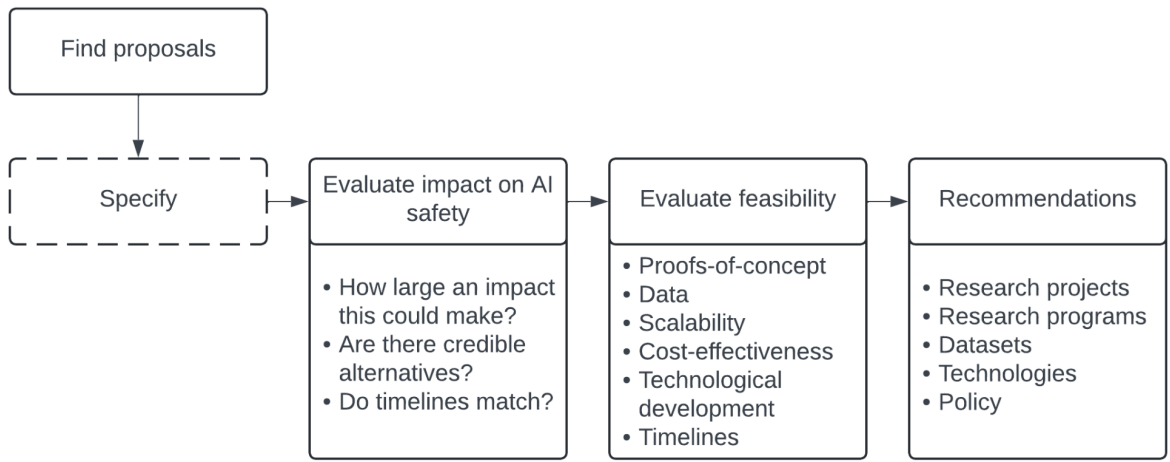}
    \caption{Process for evaluating proposals for how neuroscience can impact AI safety}
    \label{fig-neuroscience-ai-safety-process}
\end{figure}

Our process is as follows:
\begin{enumerate}
\def\labelenumi{\arabic{enumi}.}
\item
  \textbf{Specify}. Give a high-level overview of the approach.
  Highlight important background information to understand the approach.
  Where proposals are more at the conceptual level, further specify the
  proposal at a more granular level to facilitate evaluation.
\item
  \textbf{Evaluate impact}. Define the specific AI safety risks this
  approach could address. Critically evaluate the approach, with a
  preference toward approaches that positively impact safety while not
  drastically increasing capability.
\item
  \textbf{Evaluate feasibility}. Define technical criteria to make this
  proposal actionable and operationalizable, including defining tasks, recording
  capabilities, brain areas, animal models, and data scale necessary to
  make the proposal work. Evaluate their feasibility.
\item
  \textbf{Recommend}. Define the whitespace within that approach, where
  more research, conceptual frameworks, and tooling are needed to make
  the proposal actionable. Make recommendations accordingly.
\end{enumerate}

We outline our process in Figure \ref{fig-neuroscience-ai-safety-process}. We use
a broad definition of neuroscience, which includes high resolution neurophysiology in
animals, cognitive approaches leveraging non-invasive human
measurements from EEG to fMRI, and purely behavioral cognitive science.

Our audience is two-fold: 

\begin{enumerate}
    \item Practicing neuroscientists who are curious
about contributing or are already contributing to AI research, and whose research could be relevant to AI safety. For this audience, we describe technical directions that they could engage in.
\item Funders in the AI safety space who are considering neuroscience as an area of interest and
concern, and looking for an overview of proposed approaches and
frameworks to evaluate them. For this audience, we focus on describing both what's been done in this space and what yet remains.
\end{enumerate}

Our document is exhaustive, and thus quite long; sections are written to stand alone, and can be read out-of-order depending on one's interests. We have ordered the sections starting with the most concrete and engineering-driven, supported by extensive technical analysis, and proceed to more conceptual proposals in later sections. We conclude with broad directions for the field, including priorities for funders and scientists.

\begin{namedbox}{human-safety}{Are Humans Safe? A Nuanced Approach to AI Safety}
The human brain might seem like a counterintuitive model for developing safe AI systems. Our species engages in warfare, exhibits systematic biases, and often fails to cooperate across social boundaries. We display preferences for those who are closer to us--family, political affiliation, gender or race--and our judgment is clouded by heuristics and shortcuts that allow us to be manipulated \cite{Kahneman2013-aa}. Our species name, \emph{homo sapiens}--wise human--can sometimes feel like a cruel misnomer \cite{Harari2015-el}. 

Does drawing inspiration from the human brain risk embedding these flaws into AI systems? A naive replication of human neural architecture would indeed reproduce both our strengths and weaknesses. Even selecting an exemplary human mind--say, Gandhi--as a template raises complex philosophical questions about identity, values, and the nature of consciousness, themes extensively explored in science-fiction. As history shows us, exceptional intelligence does not guarantee ethical behavior; in the words of Descartes, "the greatest minds are capable of the greatest \emph{vices} as well as of the greatest \emph{virtues}" \cite{Descartes1637-gu}. Furthermore, pure replication approaches can display unintended behavior if they incorrectly capture physical details or get exposed to different inputs, as twin studies remind us that even genetically identical individuals can have very different life trajectories.

We propose instead a selective approach to studying the brain as a blueprint for safe AI systems. This involves identifying and replicating specific beneficial properties while carefully avoiding known pitfalls. Key features worth emulating include the robustness of our perceptual systems (Section \ref{sec-reverse-engineer-the-representations-of-sensory-systems}) and our capacity for cooperation and theory of mind (Section \ref{sec-develop-better-cognitive-architectures}). This approach relies on computational reductionism to isolate and understand these desirable properties.

Such selective replication is most straightforward at computational and algorithmic levels, but implementation-level approaches may also be feasible. Advances in mechanistic interpretability (Section \ref{sec-leverage-neuroscience-inspired-methods-for-mechanistic-interpretability}) offer promising tools for understanding and steering complex systems, potentially including future whole-brain biophysically detailed models (Section \ref{sec-wbs}), though this remains speculative.

In pursuing neuroscience-inspired approaches to AI safety, we must maintain both scientific rigor and ethical clarity. Not all aspects of human cognition contribute to safety, and some approaches to studying and replicating neural systems could potentially increase rather than decrease risks. Success requires carefully selecting which aspects of human cognition to emulate, guided by well-defined safety objectives and empirical evidence.
\end{namedbox}

\newpage

\tableofcontents

\newpage

\hypertarget{sec-reverse-engineer-the-representations-of-sensory-systems}{%
\section{Reverse-engineer sensory
systems}\label{sec-reverse-engineer-the-representations-of-sensory-systems}}

\hypertarget{subsec-core-idea-digital-twins}{%
\subsection{Core idea}\label{subsec-core-idea-digital-twins}}

AI systems need robust sensory systems to be safe. Sensory systems in
current AI systems are brittle: they are susceptible to adversarial
examples \cite{Goodfellow2014-pc}, they learn slowly
\cite{Zhuang2021-nm}, they can fail catastrophically
out-of-distribution \cite{Nagarajan2021-ie}, they tend
to rely on shortcuts that generalize poorly \cite{Ilyas2019-ir}, and
they don't display compositionality \cite{Lake2017-jf}.
By contrast, sensory systems in the brain are robust \cite{Geirhos2018-uf}. If we could
reverse engineer how sensory systems form these robust representations,
we could embed them in AI systems, enhancing their safety. \textit{Sensory digital twins}
are large-scale
neural networks trained to predict neural responses across a wide range
of sensory inputs \cite{wang2023towards, Bashivan2019-ha, Walker2019-bx}. We evaluate reverse engineering the representations
of sensory systems in model systems using sensory digital twins as an
intermediate.

\hypertarget{subsec-why-does-it-matter-for-ai-safety-and-why-is-neuroscience-relevant-digital-twins}{%
\subsection{Why does it matter for AI safety and why is neuroscience
relevant?}\label{subsec-why-does-it-matter-for-ai-safety-and-why-is-neuroscience-relevant-digital-twins}}

\hypertarget{subsubsec-adversarial-robustness}{%
\subsubsection{Adversarial robustness}\label{subsubsec-adversarial-robustness}}

\begin{figure}[htbp]
    \centering
    \includegraphics[width=6.5in]{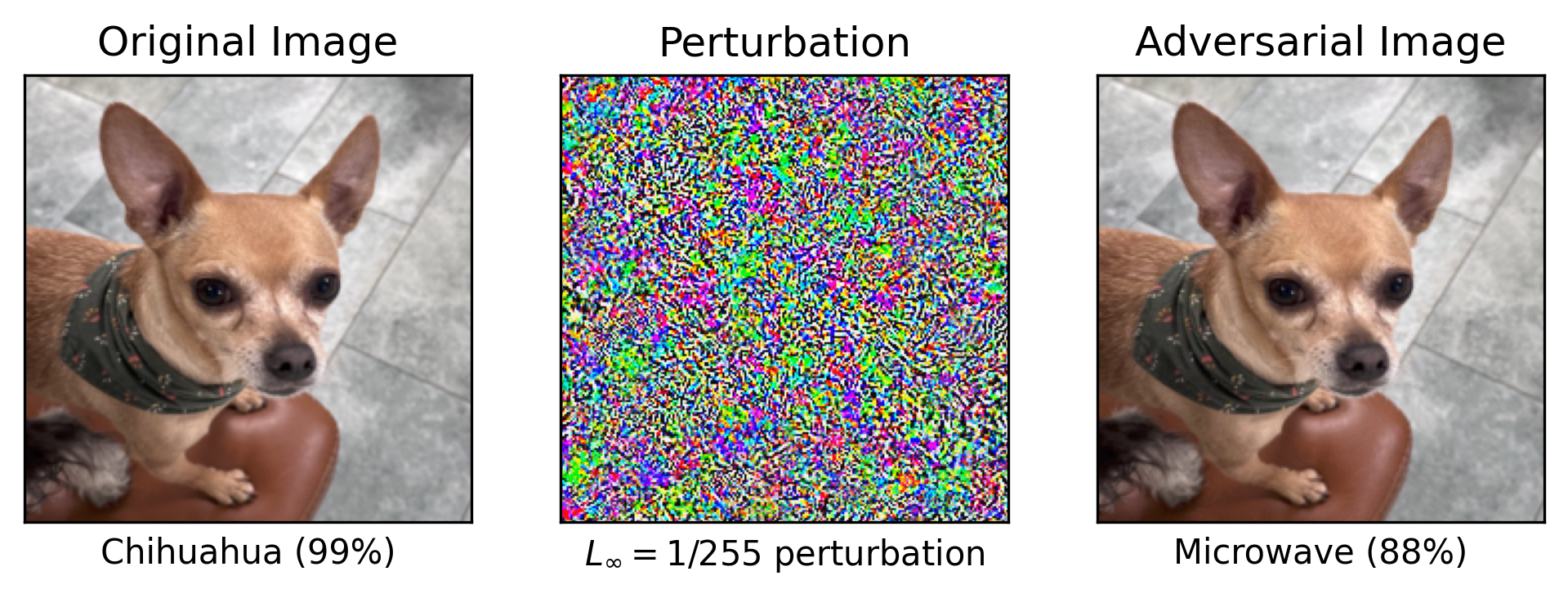}
    \caption{An example of an adversarial attack on an image recognition model (ResNet-50 from torchvision). An imperceptible change in the input image leads the model to confidently apply the label microwave to this image of a chihuahua. Inspired by \cite{Goodfellow2014-pc}.} 
    \label{fig-adversarial-attack-example}
\end{figure}

Adversarially crafted examples can cause AI systems to misbehave, and
create an attack vector for malicious actors. Small perturbations to
input data that are imperceptible to humans can cause otherwise highly
capable models to make catastrophic errors in their predictions. While
the most potent adversarial attacks leverage access to model internals, vulnerabilities still remain in the black-box setting
\cite{Gu2023-vr}. Attackers can leverage decision-based boundary attacks, which iteratively adjust inputs based on model outputs to approximate decision boundaries in black-box settings \cite{Brendel2017-nv}. These methods expose AI to vulnerabilities even without internal access, highlighting the risks of adversarial exploitation. Vulnerabilities of this kind have
justifiably raised serious concerns, particularly at a time when AI
systems are increasingly deployed in sensitive real-world contexts, including in autonomous settings
\cite{Athalye2017-jx}.

The problem of adversarial robustness is not confined to any single
domain of AI. Early results focused on image classification models,
following the discovery that deep neural networks could be fooled
through adversarial input perturbations
\cite{Goodfellow2014-pc}, but similar vulnerabilities
have been discovered in large language models
\cite{Howe2024-xl} and in superhuman game-playing
systems like KataGo \cite{Wang2022-fg}. Much effort has
gone into developing more robust AI systems, and progress has been made
through dedicated adversarial training. Scaling up models and datasets
helps, but far less than one might hope. The most capable adversarially
robust networks on CIFAR-10 and CIFAR-100 have required GPT-3.5-levels
of compute to train, and still lag behind humans
\cite{Bartoldson2024-qr}. Scaling up adversarial
robustness to human level on ImageNet is predicted to take multiple
orders of magnitude compared to GPT-4. This has led to calls for
preparing for a future where AI systems may remain inherently vulnerable
to such attacks \cite{Gleave2023-fh}.

Adversarial robustness is a human phenomenon
\cite{Ilyas2019-ir}: adversarial examples are defined by
their inscrutability to humans, and it is natural to think that their solution may lie in studying the sensory systems of humans, and
potentially primates and rodents. A potential framework to understand
these issues \cite{Ilyas2019-ir} is in terms of robust
and non-robust features. Both types of features are equally predictive
in-distribution, but robust features remain predictive under varying
amounts of distribution shift, unlike non-robust ones, which are highly
sensitive to any such change. Since there are many more potential
non-robust features than robust ones, deep neural networks are likely to
exploit non-robust features as a result of their optimization process,
which aims to maximize in-distribution performance.
Biological perceptual systems are the result of an incomprehensibly
large evolutionary search that has selected inductive biases to robustly
generalize in real-world environments
\cite{Zador2019-gf,Hasson2020-uj}.

Thus, adversarial robustness is an unsolved, human-centric AI safety
problem, and studying neural systems is a plausible route to progress.

\hypertarget{subsubsec-out-of-distribution-ood-generalization}{%
\subsubsection{Out-of-distribution (OOD)
generalization}\label{subsubsec-out-of-distribution-ood-generalization}}

Out-of-distribution inputs refer to data that are significantly
different from what a machine learning model has been trained on. These
inputs are especially prevalent in autonomous agents that navigate and
explore environments on their own. If not properly designed, such agents
can fail catastrophically when encountering unfamiliar inputs. In
contrast, humans display adaptive behaviors even in entirely new
situations, including those requiring zero-shot (no prior exposure) or
few-shot (minimal prior exposure) learning. This adaptability is often
attributed to our ability to recognize and combine familiar components
in new ways, a capability known as finding and exploiting compositional
representations \cite{Lake2015-kb}, and to learn representations of the world that disentangle its causal variables. For example, an agent that has disentangled texture, color, and shape can correctly classify an object with unique combinations of variables it has never encountered before, such as a pink elephant.

Foundation models have alleviated some of these concerns by pretraining
on massive, internet-scale datasets, effectively incorporating
previously unseen scenarios into their training distribution. However,
this approach is impractical for covering all possible
out-of-distribution scenarios because it would require training on an
exponentially large number of cases \cite{Filos2020-ow}. For example, training a self-driving car to handle every possible road condition, weather scenario, pedestrian behavior, and vehicle type would require simulating or collecting data from an unimaginably vast number of situations. Rare edge cases, such as a child running into the street during a snowstorm while an autonomous car encounters a malfunctioning traffic light, are extremely costly and difficult to capture comprehensively in a dataset. Similar to the challenge of achieving adversarial robustness, ensuring robustness to out-of-distribution inputs remains an unsolved problem with important safety implications.

\hypertarget{subsubsec-specification-alignment-and-simulating-human-sensory-systems}{%
\subsubsection{Specification alignment and simulating human sensory
systems}\label{subsubsec-specification-alignment-and-simulating-human-sensory-systems}}

Current systems view the world and sense the world in different ways
than humans. This can pose a safety risk in out-of-distribution
situations, where specifications written by a human are misinterpreted
by an AI that lacks the primitives of the human mind (e.g., understanding causality, context, or social norms)
\cite{Sucholutsky2023-fx}. Autonomous AI agents will
need to be able to simulate how their actions affect the world to safely
explore the world and act inside of it
\cite{Dalrymple2024-sf}. AI agents will thus need to
engage in perspective-taking, simulating how a human would react to a
particular set of sensory inputs in a model-based fashion
\cite{Linsley2024-ee}. Again, building good models of
human sensory systems is a stepping stone toward safe human-AI
interactions.

\hypertarget{subsubsec-bridging-neuroscience-and-ai-through-digital-twins}{%
\subsubsection{Bridging neuroscience and AI through digital
twins}\label{subsubsec-bridging-neuroscience-and-ai-through-digital-twins}}

To solve the problem of reverse engineering sensory processing, a natural intermediate milestone is to build a model that can account for responses of neurons to arbitrary stimuli. A \textit{sensory digital
twin} is trained to learn the relationship between stimuli and the resulting neural response as well as how these sensory representations are modulated by motor variables and brain states~\cite{sinz2018stimulus, Walker2019-bx, franke2022state}. If trained with enough data, the model can be used to simulate the neural response to data never seen by the animal~\cite{wang2023towards}, allowing researchers to simulate, parallelize and scale experiments \emph{in silico} that would be impossible \emph{in vivo}. Sensory digital twins have been used to uncover how neurons in the visual cortex adapt their tuning selectivity to changing brain states~\cite{franke2022state}; how they integrate local and contextual information to optimize information processing~\cite{fu2023pattern}; and to systematically characterize single-neuron invariances~\cite{ding2023bipartite}.

Critically, digital twins are built using artificial neural networks, allowing the use of mechanistic interpretability (Section \ref{sec-leverage-neuroscience-inspired-methods-for-mechanistic-interpretability}) to understand how they function. For example, if we were to create a digital twin of the entire primate visual brain, we could investigate how it constructs adversarially and distributionally robust representations that are useful for behavior. In the following, we turn our attention to the feasibility of
building digital twins given current technology.

\hypertarget{subsec-details-digitaltwins}{%
\subsection{Details}\label{subsec-details-digitaltwins}}

\hypertarget{subsubsec-digital-twins}{%
\subsubsection{Sensory digital twins}\label{subsubsec-digital-twins}}

\begin{figure}[htbp]
    \centering
    \includegraphics[width=.8\textwidth]{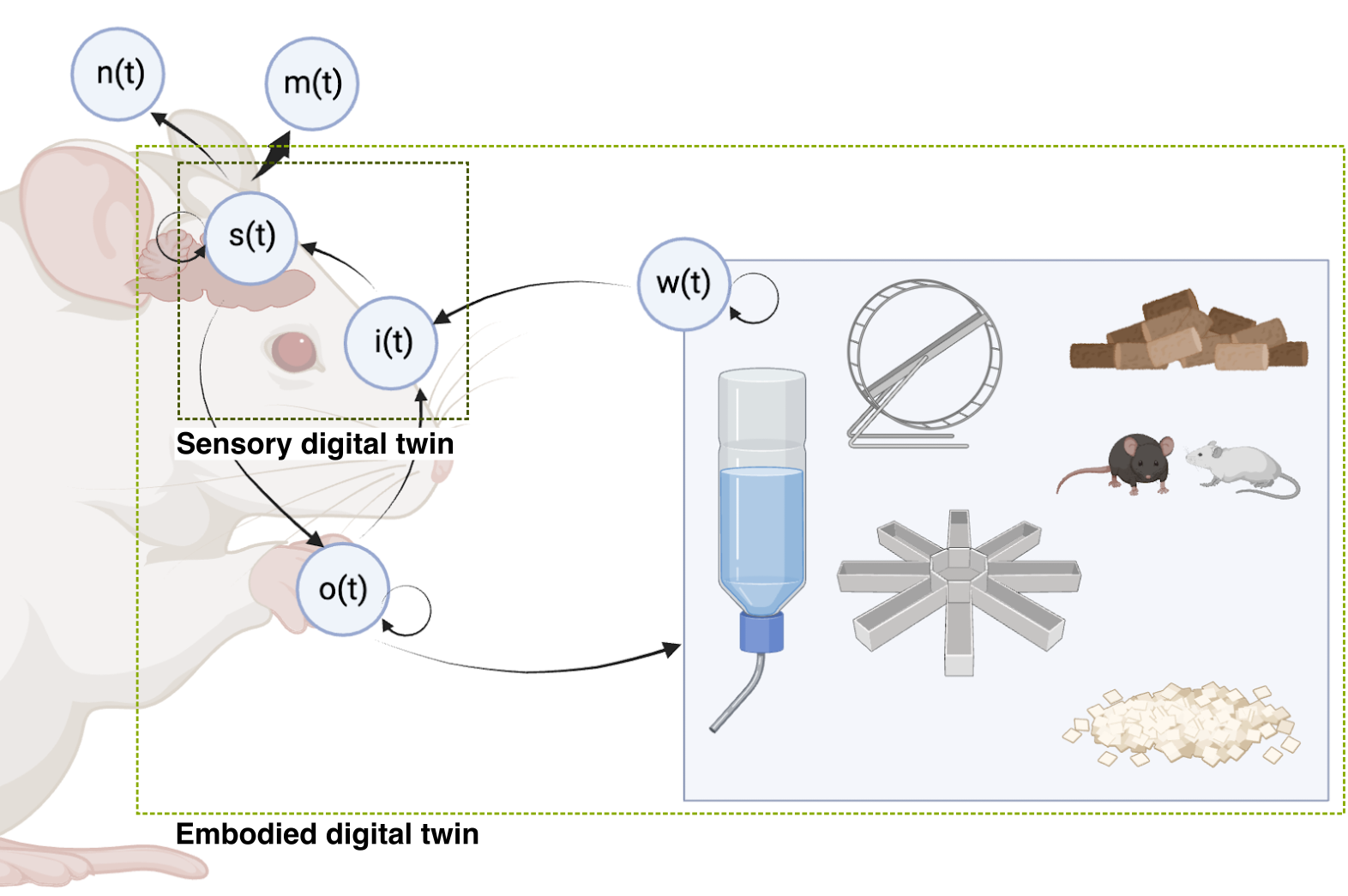}
    \caption{A digital twin of a sensory system seeks primarily to account for the relationship between input $i(t)$ and the state of the brain $s(t)$ as reflected by high-dimensional measurements $m(t)$, potentially taking into account. By contrast, an embodied digital twin (Section \ref{sec-embodied}) must account for the full relationship between input, brain state, motor output $o(t)$, as well as the world $w(t)$.}
    \label{fig-digital-twin-sensory-system}
\end{figure}

Foundation models of the brain, whole-brain simulations, and digital twins are
sometimes conflated. For the purpose of this discussion, we use the following definitions (Figure \ref{fig-digital-twin-sensory-system}):

\begin{enumerate}
    \item \textbf{Sensory digital twins} seek to model an animal's sensory systems, potentially with auxillary task-relevant inputs such as cognitive and motor state, at the level of representations. These are the topic of this section.
    \item \textbf{Embodied digital twins} seek to model an entire animal, including its sensory, cognitive and motor systems, its body, and its relationship to the environment. An embodied digital twin could contain a sensory digital twin. We cover embodied digital twins in Section \ref{sec-embodied}.
    \item \textbf{Biophysically detailed models} seek to model nervous systems from the bottom-up, with detailed simulations that may include biophysically detailed neuron models and connectomes. These are covered in Section \ref{sec-wbs}.
    \item \textbf{Foundation models of the brain} are AI models trained at a large scale, usually using self-supervised or unsupervised learning, that seek to find good representations of neural data--which may include single-neuron data, local field potentials, electrocorticography, EEG, fMRI, and optionally behavior. Note that foundation models are a general approach that could be used as past of the construction of the types of models above. They are the subject of Section \ref{subsubsec-foundation-models-for-neuroscience}.
\end{enumerate}

The boundary between these types of models can be diffuse, but we use them here to facilitate concrete discussion. We use the following working definition for a sensory digital twin:

\begin{itemize}
\item They are models that learn the relationship between stimuli and neural responses $f(i(t)) \to s(t)$, where $i(t)$ are stimuli and $s(t)$ are neural response vectors. Sensory inputs can potentially include visual, auditory, somatosensory, proprioceptive or olfactory inputs.
\item
  They are typically trained directly on single neuron response data, either from
  scratch or via fine-tuning after task-relevant pre-training, such as
  image recognition.
\item
  They are optimized to maximize their success at predicting neural responses, using metrics such as $R^2$ or log-likelihood \cite{Wu2006-rd}.
\item
  They are primarily focused on predicting responses of neurons to
  sensory stimuli, although they can also leverage other task-relevant
  inputs such as cognitive state, or motor inputs such as eye position and pupil dilation.
\item
  They use scalable architectures such as convolutional neural networks
  (CNNs), transformers or state-space models (SSMs) that can leverage
  arbitrarily large datasets.
\end{itemize}

As their predictive accuracy and ability to generalize improves, these
models capture more and more of the computational principles and
representations that underlie biological perception. 

Much recent work has focused on the building of digital twins of the
primary visual cortex (V1) in mice
\cite{Walker2019-bx,Lurz2020-cs,Cobos2022-lo,wang2023towards,Du2024-ho,Ding2023-af}
and macaques \cite{Cadena2019-xp,Du2024-ho,Miao2023-tk}.
This line of work has been extended to other visual macaque area V4
\cite{Bashivan2019-ha,Willeke2023-ae,Cowley2023-jb,Cadena2024-vg,Pierzchlewicz2023-rj,Wang2024-ie}.
This follows from a long tradition, predating the deep learning
revolution, of characterizing visual systems through nonlinear systems
identification methods
\cite{Wu2006-rd,Rust2006-yw,Nishimoto2011-op,Willmore2010-cn,Mineault2012-hu,Pasupathy2002-wv}.
Although vision is by far the most well-studied modality,
\cite{Rancon2024-dx} have developed digital twins for
auditory processing in rats and ferrets, and
\cite{Marin_Vargas2024-xp} for proprioception in
the macaque brainstem and somatosensory cortex.

\begin{figure}[htbp]
    \centering
    \includegraphics[width=5.5in]{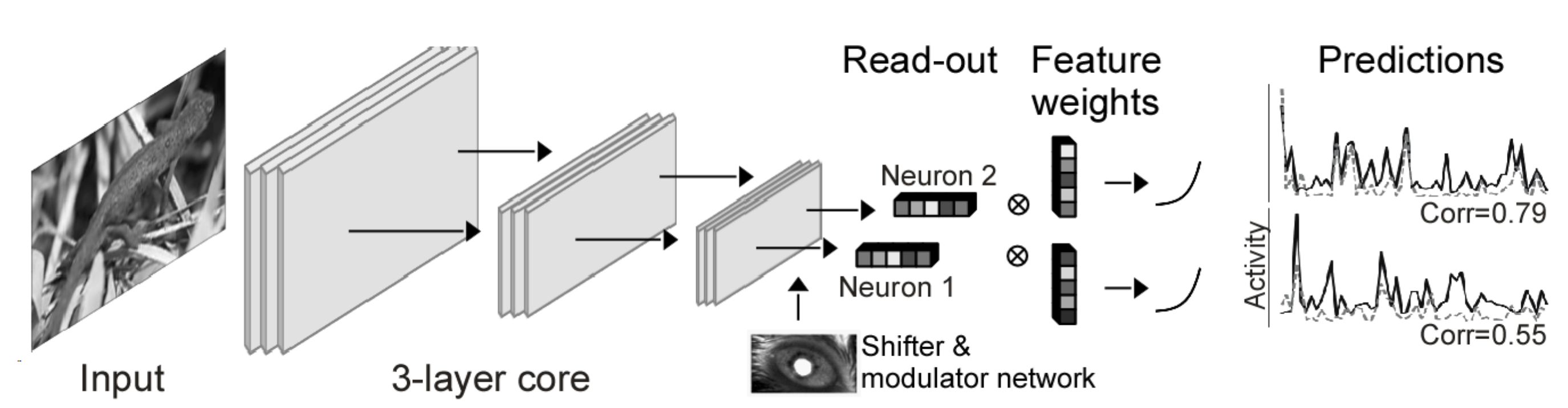}
    \caption{Typical layout of a visual digital twin trained on neural data. Images are processed by a common core, which gets read out by a series of weights to predict the firing rate of a set of neurons. This visual input-neuronal output mapping may be modulated by external drives, for example the position of the eye. Both the common core and the readout feature weights are learned. In the following, we distinguish between scaling laws for the core and the readout. From \cite{Fu2023-pe}, under a CC-BY-NC-ND 4.0 license.}
    \label{fig-visual-twin-layout}
\end{figure}

As a concrete example of how one might build a digital twin, consider a deep CNN trained from scratch on a
large-scale dataset consisting of thousands of paired natural images
(sensory stimuli) and simultaneous neural responses collected from
thousands of neurons in the primary visual cortex (V1) of multiple mice (Figure \ref{fig-visual-twin-layout}).
The model receives as input an image, and optionally a set of auxiliary
information, including the mouse\textquotesingle s eye position and pupil dilation.
The \emph{core} of the model is a CNN which finds a
good latent representation of the visual input; neuron-specific
\emph{readouts} perform a linear weighting of the latents to predict the
response of each neuron. The entire network is trained end-to-end to
predict neuron responses using a learning objective such as Poisson
likelihood or mean-squared error.

Once a core has been trained on a sufficient amount of neural data, the
hope is that it can be used as a proxy for sensory systems in a wide
range of scenarios: that it captures a kind of average sensory system.
An ideal core would transfer to other unrecorded neurons in the same
area and same animal; to neurons in other animals; to stimuli outside
the input distribution; or even to other species. It could also serve as
a basis for understanding higher-level areas in the same species.

\hypertarget{subsubsec-feasibility-of-building-digital-twins}{%
\subsubsection{Feasibility of building digital
twins}\label{subsubsec-feasibility-of-building-digital-twins}}

Despite 70+ years of research on the primary visual cortex (V1) \cite{Hubel1959-zo}, and consensus by neuroscientists that it is one of the most well-understood brain areas, Olshausen \& Field \cite{Olshausen2006-fc} famously estimated that we still do not know what 85\% of V1 is doing. They attributed this to a combination of nonlinearities, low spike rates,
inaccessible neurons, and low variance accounted for. Now that large-scale recordings
and powerful nonlinear function approximators are ubiquitous, is it feasible
to capture the response properties of all neurons in V1--or perhaps the
entire sensory system--using digital twins? How much data would be needed to train models to capture its variance? 

\begin{figure}[htbp]
    \centering
    \includegraphics[width=6.5in]{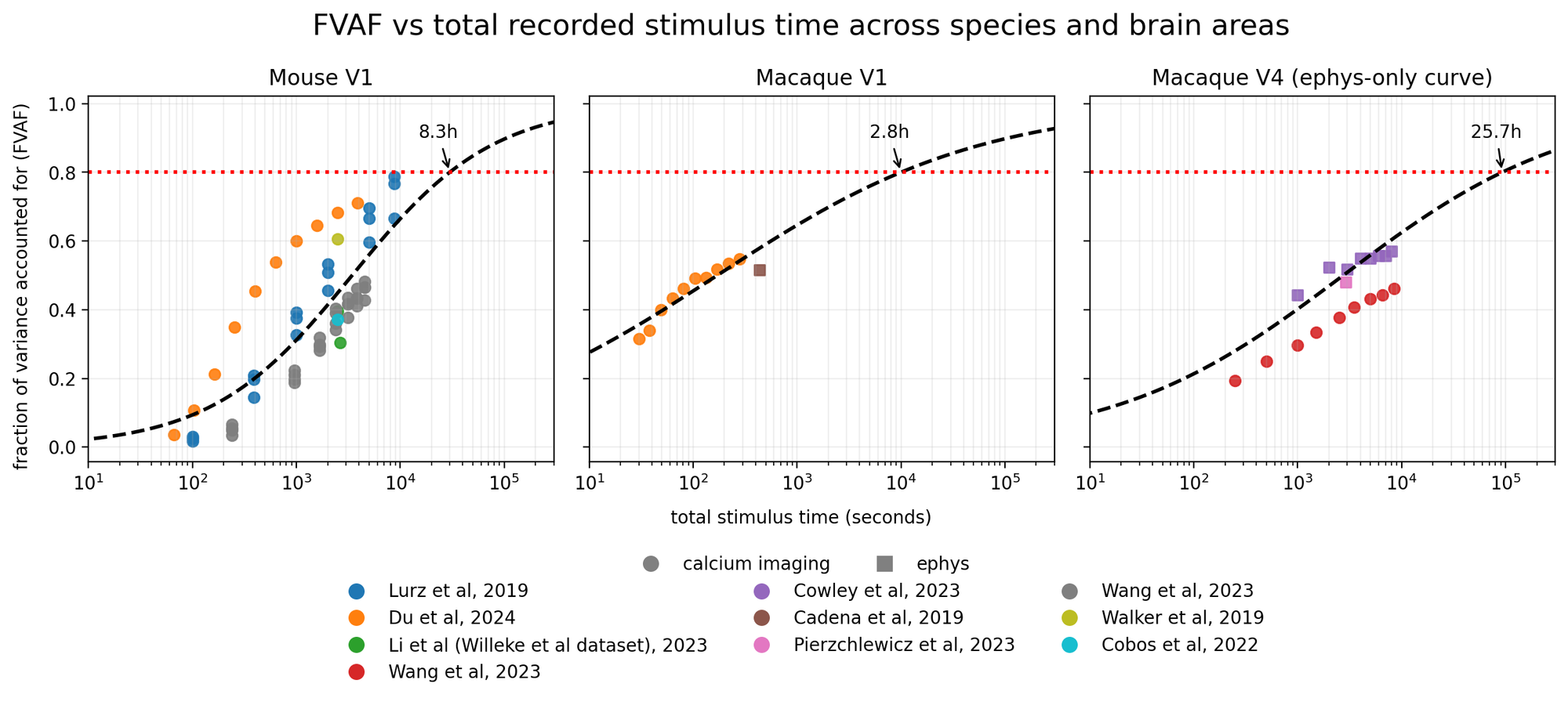}
    \caption{Empirical scaling laws for digital twins of visual areas across 10 different studies. Data from \cite{Lurz2020-cs, Cowley2023-jb, wang2023towards, Du2024-ho, Cadena2019-xp, Walker2019-bx, Li2023-iz, Pierzchlewicz2023-rj, Cobos2022-lo, Wang2024-ie}.}
    \label{fig-visual-scaling-laws}
\end{figure}

One way to answer this question is to examine and estimate scaling laws for digital
twins of the visual system, drawing from existing results in the literature.
Several studies that have built digital twins of visual areas in response to natural images and movies have
documented the average performance of their models in predicting
held-out data. A select few
\cite{Lurz2020-cs,Du2024-ho,wang2023towards,Wang2024-ie,Cowley2023-jb}
have systematically varied the amount of training data that goes into
fitting the models and measured goodness-of-fit on a held-out dataset. We
aggregated all of these studies in Figure \ref{fig-visual-scaling-laws}, rescaling the goodness-of-fit to fraction of explainable variance explained (FEVE) to allow consistent comparison across all studies (see Methods for details). 

The shape of these curves across three areas (mouse V1; macaque V1; macaque V4) is well-described by a sigmoid in log-recording time:

$$\textrm{FEVE} = \sigma(a \log(t) + c)$$

We extrapolated these scaling laws to estimate the total amount of
recording time necessary to obtain an average validated fraction of
explainable variance explained (FEVE) across all recordings greater than
80\% for their training distribution. This yielded a projected recording
time to 80\% FEVE of 3 hours in macaque V1, 8 hours in mouse V1, and 26
hours in macaque V4. 

We emphasize here that these scaling laws were derived from presenting natural images, except for \cite{wang2023towards}, which used natural movies, and thus only represent one potential stimulus class; that the stimulus sets and preprocessing steps were different across the different studies, such that the results are not directly comparable across studies; and that the distributions of stimuli were generally similar across train and test sets, thus representing a best-case scenario during evaluation--for out-of-distribution model evaluation see \cite{wang2023towards}. Despite these caveats, it is interesting and noteworthy that digital twins appear to follow similar scaling laws across a range of studies \cite{Lurz2020-cs, Kaplan2020-dj}.

At face value, this indicates that obtaining good digital twins of
single neurons in the visual system to natural images is at the edge of feasibility given
current technological limitations with acute recording technologies like
Neuropixels (maximum recording length of \textasciitilde3-4 hours in a
single session). In particular, in macaque V4, one would likely need
chronic electrophysiological recordings--or recording for multiple days with a technology,
like calcium imaging, that allows one to stitch recordings--to get to
the projected 26 hours of recordings. We note, however, that some stimulus types can be more informative than others \cite{Talebi2012-aa, Cowley2020-vs}, and that closed-loop techniques could significantly improve upon these scaling laws \cite{Walker2019-bx, Bashivan2019-ha, Ponce2019-ll}. 

\hypertarget{subsubsec-scaling-laws-for-cores-vs-readouts}{%
\subsubsection{Scaling laws for cores vs. readouts}\label{scaling-laws-for-cores-vs-readouts}}

The scaling laws in the previous section focused on the feasibility of building a digital twin \emph{of a single neuron}, which depends on effectively learning the readout.  However, the artifact we really care about is the \emph{core}, a distillation of the processing within an area, which can be learned by stitching data together across neurons, animals and sessions. In the particular implementation of digital twins we discussed so far, the tuning of the neurons is embedded in a latent space that learns the non-linear relationships between neural responses and sensory input and other variables like motor responses. Each neuron’s response is then modeled as a linear readout of the core, where the readout is learned separately for each neuron. How do cores scale with data? In particular, how much data do we need to learn good cores, and how does this affect the scaling laws for learning accurate models of individual neurons?

To disentangle core and readout scaling, we turn to simulations. We start
with a simulation where \emph{the correct core} is known \textit{a priori}: it is simply the identity function over the inputs. In
this case, all we need to learn are the weights of the readouts, and
we're in a situation equivalent to multivariate Poisson regression. We
generate random design matrices and random weights for
linear-nonlinear-Poisson (LNP) neurons, estimate maximum a posteriori
weights (MAP) under Tikhonov regularization constraints, and estimate
the FEVE in a validation dataset.

We find that scaling laws for these models are qualitatively and
quantitatively well explained by a sigmoid as a function of log
recording time and log number of parameters in the readout:

$$\textrm{FEVE} = \sigma(a \log(t) + b \log(\textrm{readout params}) + c)$$

On a log-linear scale, increasing the dimensionality of the core, and consequently of the readout weights, shifts
the curves to the right, as more data is needed to fit the readout
weights. We derive a mathematical expression for the functional form of
scaling laws for linear regression which matches the log-sigmoid scaling
laws we empirically find here (see Appendix).

What happens when the core is \emph{incorrect}? Imagine, for example,
that a neuron in V1 is selective for the sign of an edge, but displays
some translation invariance. In other words, this neuron displays
properties that place it somewhere between the classic
orientation-and-phase-selective simple cell and a phase-invariant
complex cell. If we tried to learn a linear mapping from an image's
luminance values to this neuron's responses, we would not be able to
fully account for the responses of the neuron no matter how much data we
train on; we'd need a nonlinear mapping.

\begin{figure}[htbp]
    \centering
    \begin{subfigure}[b]{0.48\textwidth}
        \centering
        \includegraphics[width=\textwidth]{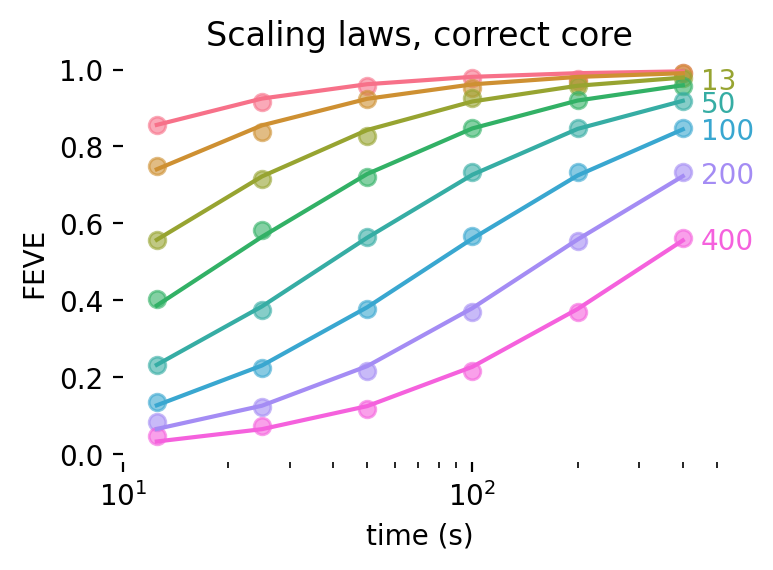}
    \end{subfigure}
    \hfill
    \begin{subfigure}[b]{0.48\textwidth}
        \centering
        \includegraphics[width=\textwidth]{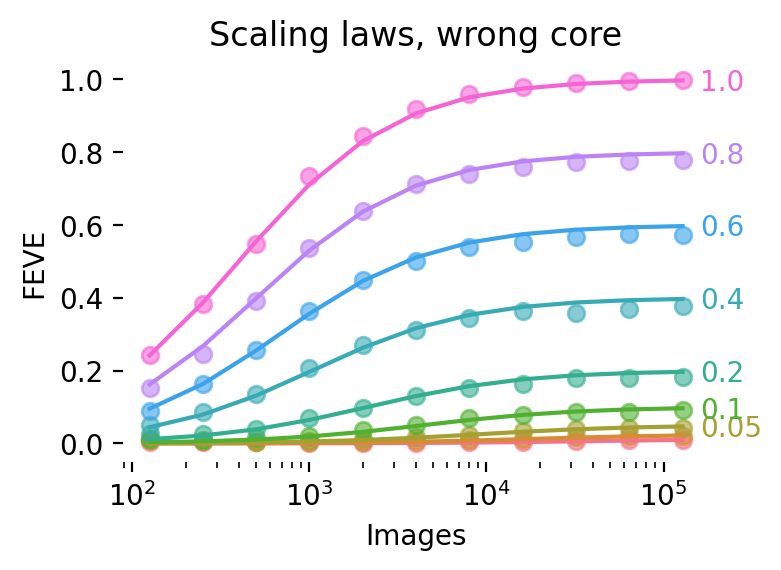}
    \end{subfigure}
    \caption{(Top) simulated scaling laws for the correct core as a function of recording time and number of parameters in the core (colored labels). An FEVE of 1 corresponds to a perfect fit. All the curves are parallel, but cores with larger number of parameters are shifted to the right, and they require more data to fit. (Bottom) Simulated scaling laws for incorrect cores. Cores which are incorrect have scaled down scaling curves--even with infinite training data, their performance asymptotes below 1. Furthermore, they display delayed learning compared to the correct core.}
    \label{fig-scaling-laws-comparison}
\end{figure}

We simulate this effect by partitioning the design matrix into two
components: one part which is known, and for which we can estimate
linear weights as before; and a second component which is unknown, and
for which we cannot estimate weights by \emph{construction}. The effect
is shown in Figure \ref{fig-scaling-laws-comparison}. The takeaway is that using the wrong core
\emph{scales down} the entire curve, capping the maximum attainable
FEVE. A secondary effect is that using a bad core delays learning, as
the unaccounted component of a neuron's response decreases its effective
signal-to-noise ratio.

These effects are captured by the following scaling law:

$$\textrm{FEVE} = R^2_{core} \sigma(a \log(t) + b \log(R^2_{core}) + c)$$

Note that this scaling law predicts that with bad cores, learning is
delayed indefinitely. We obtain excellent fits with three free
parameters (R\textsuperscript{2} \textgreater{} .999 in both cases) with
this functional form.

Does real neural data follow the same scaling laws as these simulations?
We turn to Lurz et al. (2021) \cite{Lurz2020-cs} to verify that this is the case. Their figure 5
explores how single neuron fits in the mouse V1 scale with data for
different \emph{fixed cores} that are matched to mouse V1 to varying degrees. This comparison corresponds exactly to our \emph{wrong core}
simulations, and indeed we obtain excellent fits using the same
functional form we used in simulations (R\textsuperscript{2}=.997).

\begin{figure}[htbp]
    \centering
    \begin{subfigure}[b]{0.48\textwidth}
        \centering
        \includegraphics[width=\textwidth]{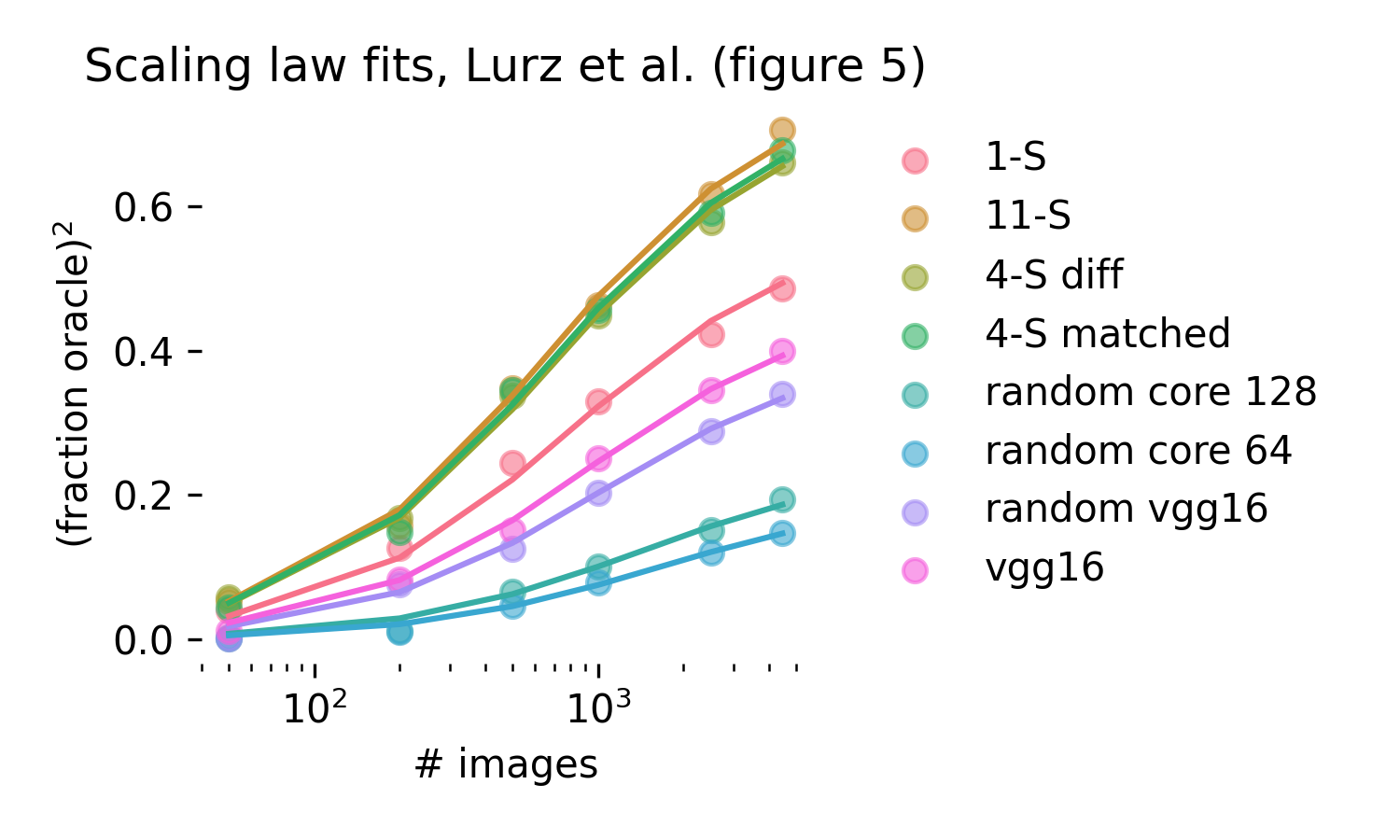}
    \end{subfigure}
    \hfill
    \begin{subfigure}[b]{0.48\textwidth}
        \centering
        \includegraphics[width=\textwidth]{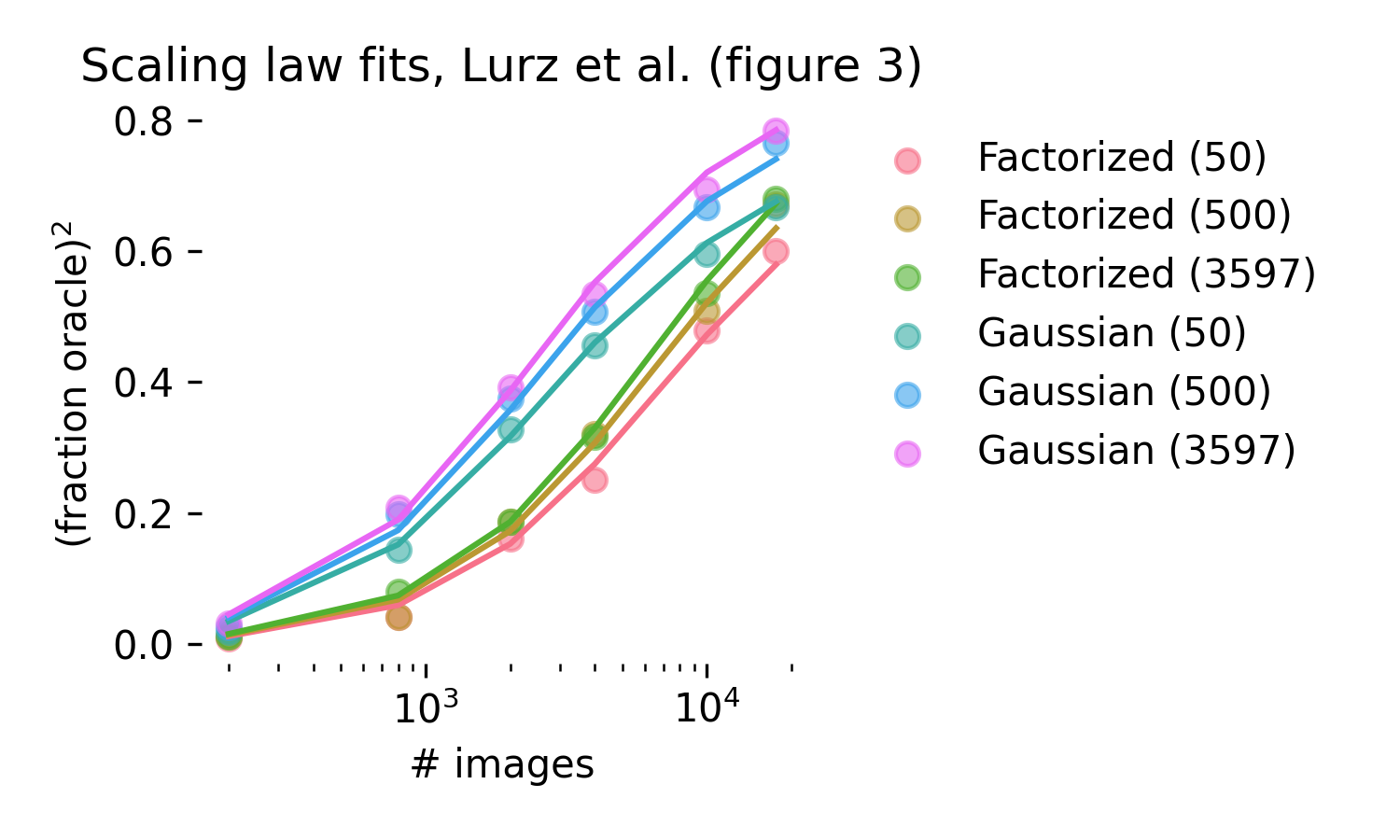}
    \end{subfigure}
    \caption{Scaling laws are well-described by sigmoid in log recording time for both core and readout. Data from \cite{Lurz2020-cs}. Top: scaling laws for different fixed cores. Bottom: scaling laws for core and readout.}
    \label{fig-sigmoid-scaling-laws}
\end{figure}

What happens when cores are \emph{learned}? We assume that the maximal
attainable R\textsuperscript{2} by a core scales similarly to how single
neuron training scales, but with different scaling
parameters\footnote{We consider a single scaling parameter
  $a_{core}$, but one could easily extend this framework to
  have different scaling parameters for, e.g. recording time, number of
  neurons, number of animals, entropy of the stimuli, etc.}:

$$R^2_{core} = \sigma(a_{core} \log(t \cdot \textrm{neurons}) + c_{core})$$

Figure 3 in Lurz et al. (2021) quantifies how single neuron fits scale
as more data is used to train \emph{both the core and the readout},
allowing us to test this functional form. We obtain excellent fits,
R\textsuperscript{2} = .996 for a 5 parameter scaling law. Importantly,
scaling laws for the core are far slower than for the readout; the
parameter a is estimated to be 1.13, while $a_{core}$ is
estimated at .15. The core, in these model fits, is far more data hungry
than the readout, but better cores can be learned by pooling data together across experimental sessions and animals--that is, by sampling different neurons from the same area.

Thus, scaling curves for individual neurons are the product of the
effect of the maximal variance attainable by the core and the readout
scaling law for that neuron \cite{Canatar2023-ya}. We
summarize how different factors will affect the scaling law for a single
neuron:

\begin{itemize}
\item
  \emph{Different readout mechanisms}. Higher dimensional dense readouts will
  shift learning curves right, and lower dimensional readouts will shift
  learning curves left, provided they don't discard crucial variance in
  the core. Sparse readouts, whose scaling should depend on the number
  of \emph{non-zero} readout parameters rather than total parameters,
  are an underexplored but promising avenue to increase readout
  efficiency \cite{Du2024-ho,Cowley2023-jb}.
 \item
  \emph{Different cores}. Fixed, suboptimal cores cause two effects: scaling
  learning curves down, and moving learning curves to the right. Cores
  should be learned from the data. Data-efficient architectures trained on large numbers of images and neurons can shift the scaling curves to the left. Congruent with this, Wang and colleagues \cite{wang2023towards} showed that a more powerful core trained on eight mice mice and around 66,000 neurons could reach the same performance with 16 minutes of recording time in a new mouse than with 76 minutes of data if trained on the neurons from the new mouse alone.
 \item
  \emph{Different areas}. Areas higher up the visual hierarchy, which accumulate multiple stages of nonlinear processing, likely require learning a more elaborate core. This makes it more challenging to learn the response function of a single neuron. A promising research
  avenue is to use a core initialized from a lower-level area to
  bootstrap learning of a core for a higher-level area, and to use deep learning architectures that can better model internal brain states~\cite{bashiri2021flow}, which are more dominant in areas higher up in the visual hierarchy.
 \item
  \emph{Different estimation schemes for maximum variance explainable}:
  Misestimating maximum explainable variance should cause a
  multiplicative scaling of learning curves.
 \item
  \emph{Different SNR thresholds for inclusion}: Including neurons with low
  spike rates is expected to drag average prediction performance down,
  shifting learning curves to the right. Using low SNR modalities like
  single-photon imaging should have the same effect
  \cite{Wang2024-ie}.
 \item
  \emph{Different stimulus sets}. A less well-explored phenomenon is the effect
  of different types of stimuli \cite{Talebi2012-aa}.
  Using spatiotemporally varying stimuli (movies) rather than static
  stimuli (images) is expected to shift learning curves to the right, as
  more parameters are needed to describe the readout.
\end{itemize}

While we have focused on scaling laws for single neurons, the framework
proposed here allows one to estimate the quality of a \emph{core}:
the asymptotic $R^2$ of a core, the maximum that could be obtained if one had access to infinite single neuron data, can be extrapolated from fitting single-neuron scaling laws. Obtaining a high ranking on this
metric is a prerequisite for making a claim that a digital twin has been
achieved. However, it is by no means the only metric that matters
\cite{Williams2021-uu,Duong2023-ej,Kriegeskorte2021-vy}.

Given the bulk of the evidence, we tentatively conclude that it should
be feasible to learn good models of individual neurons in visual areas
of mice and macaques using single session, acute recordings (2-3 hours)
by distilling rich cores from multiple sessions and animals. We leave
open the possibility that higher-level areas may require chronic
recordings to go past the 3-hour barrier
\cite{Yasar2024-eu}. Of course, such models will be
static snapshots stitched together from multiple animals; dynamic models
incorporating learning, or personalized models are likely to still
require a larger amount of data, although it's difficult to forecast at
this time. While we have not established that this holds for modalities
other than vision, we see no obvious additional technical barriers to
doing this for audition; somatosensory and olfactory maps will likely
require better, more comprehensive actuators than is currently feasible.

Although macaques have been used as stand-ins for humans in vision research for decades, and there
is high homology, their visual representations and brain organization do differ from
humans \cite{Tootell2003-oy}. Furthermore, the human
visual cortex is rarely recorded with electrophysiology, although this may
change through a push toward cortical visual prosthesis
\cite{Liu2022-vp}. Fine-tuning macaque pretrained models
on human data collected noninvasively is one potential route to human digital twins.
This also appears more feasible than directly training on fMRI data,
where each voxel averages over tens of thousands of neurons, and where a
wide variety of models can account for the data equally well
\cite{St-Yves2022-pa}.

Beyond vision, audition is greatly expanded in humans and specialized
for language. Given the push toward novel speech neuroprosthesis
\cite{Card2023-hx,Moses2021-ki}, and the sustained
interest in understanding audition due to its relationship to language
\cite{Bouchard2013-qz}, it seems plausible that human
auditory cortex data will be plentiful enough in the near future to
build a digital twin \cite{Mathis2024-yk} directly from human data, although it could benefit from pre-training on macaque auditory cortex. This is in
addition to complementary advances in noninvasive decoding from auditory
and speech-related areas
\cite{Defossez2022-ws,Vaidya2022-gu,Schrimpf2020-uu}.

\hypertarget{subsubsec-feasibility-of-transferring-robustness-from-brains-to-models}{%
\subsubsection{Feasibility of transferring robustness from brains to
models}\label{subsubsec-feasibility-of-transferring-robustness-from-brains-to-models}}

We have seen that it's practically feasible to build digital twins that
accurately predict neural responses within their training distribution.
Since primates and humans are robust to out-of-distribution shifts, and
they are not sensitive to adversarial examples
\cite{Elsayed2018-oe,Zhou2018-kd,Veerabadran2023-ov,Gaziv2023-kj,Guo2022-ef,Bartoldson2024-qr},
it would stand to reason that distilling neural data should lead to
adversarially and distributionally robust neural networks. If this was
the case, one could potentially use these robust digital twins either as
adversarially robust networks for classification purposes, or as a means
of reverse engineering adversarial robustness.

To the best of our knowledge, a direct approach--simply training a
neural network to imitate neural data at scale and testing its
adversarial robustness--has not been tried. However, several references
instead use neural data to regularize networks trained for image
classification. They report that regularization with neural data leads
to higher adversarial or distributional robustness in the domain of
natural images (Table \ref{tab-neural-data-augmentation} with summary of results). We refer to these
methods collectively as neural data augmentation (NDA). These prior
results give valuable insight into whether distilling robust
representations from neural data to digital twins is feasible.

\clearpage

\setlength\LTleft{-.5in}
\setlength\LTright{-.1in}

\begin{longtable}{
    p{0.10\textwidth}  
    p{0.23\textwidth}  
    p{0.15\textwidth}  
    p{0.12\textwidth}  
    p{0.13\textwidth}  
    p{0.14\textwidth}  
    p{0.13\textwidth}  
    }

\toprule
\textbf{Paper} & \textbf{Neural dataset} & \textbf{Regularization method} & \textbf{Target network} & \textbf{Evaluation dataset} & \textbf{Attack Type} & \textbf{Accuracy gain vs baseline} \\
\midrule

\cite{Li2019-dg} & 
Responses from 8k mouse V1 neurons obtained with 5.1k grayscale ImageNet images & 
Cosine Similarity Matching loss & 
ResNet-18 & 
Grayscale CIFAR-10 & 
Gaussian noise ($\sigma = 0.08$) & 
+22.5\% \\

~ & 
~ & 
~ & 
ResNet-34 & 
Grayscale CIFAR-100 & 
Gaussian noise ($\sigma = 0.08$) & 
+12.9\% \\

\hline

\cite{Dapello2022-gg} & 
Responses from 188 macaque IT neurons obtained with 2880 grayscale HVM images & 
Centered Kernel Alignment (CKA) loss & 
CORnet-S & 
HVM (IID) & 
PGD $L_\infty (\epsilon = 0.001)$ & 
+20.9\% \\

~ & 
~ & 
~ & 
CORnet-S & 
HVM (IID) & 
PGD $L_2 (\epsilon = 0.25)$ & 
+13.3\% \\

~ & 
~ & 
~ & 
CORnet-S & 
COCO (OOD) & 
PGD $L_\infty (\epsilon = 0.002)$ & 
+11.1\% \\

~ & 
~ & 
~ & 
CORnet-S & 
COCO (OOD) & 
PGD $L_2 (\epsilon = 0.2)$ & 
+9.6\% \\

\hline

\cite{Dapello2020-af} & 
Responses from 102 macaque V1 neurons obtained with 450 naturalistic textures and noise samples & 
Biologically-constrained V1 front-end (VOneBlock) & 
ResNet50 & 
ImageNet & 
PGD (high strength) & 
+37.1\% \\

~ & 
~ & 
~ & 
CORnet-S & 
ImageNet & 
PGD (high strength) & 
+36.4\% \\

~ & 
~ & 
~ & 
AlexNet & 
ImageNet & 
PGD (high strength) & 
+18.1\% \\

\hline

\cite{Safarani2021-ui} & 
Responses from 458 macaque V1 neurons obtained with 24075 ImageNet images & 
Multi-task learning with macaque V1 response prediction & 
VGG-19 & 
TinyImageNet-TC & 
ImageNet-C corruptions (without blur) & 
+9.0\% \\

\bottomrule
\caption{Summary of neural data augmentation approaches
and their
results.}
\label{tab-neural-data-augmentation}
\end{longtable}

Despite these promising results, we don't have a current theory for why
NDA networks are robust, and if they could ever be competitive with
state-of-the-art defenses. Some studies suggest that low-frequency features can to some extent enhance robustness \cite{li2023robust}, and NDA might help learn better features more generally. The most successful class of defense against
adversarial attacks remains adversarial training together with massive
data augmentation \cite{Bartoldson2024-qr}. This relies
on training a network on \emph{adversarial examples}. Yet, NDA only accounts for responses to \emph{clean
examples}, not to adversarial examples. Why does training on clean
examples improve the performance of neural networks on corrupted
(adversarial, out-of-distribution) examples?

\begin{figure}[htbp]
    \centering
    \includegraphics[width=3.83854in]{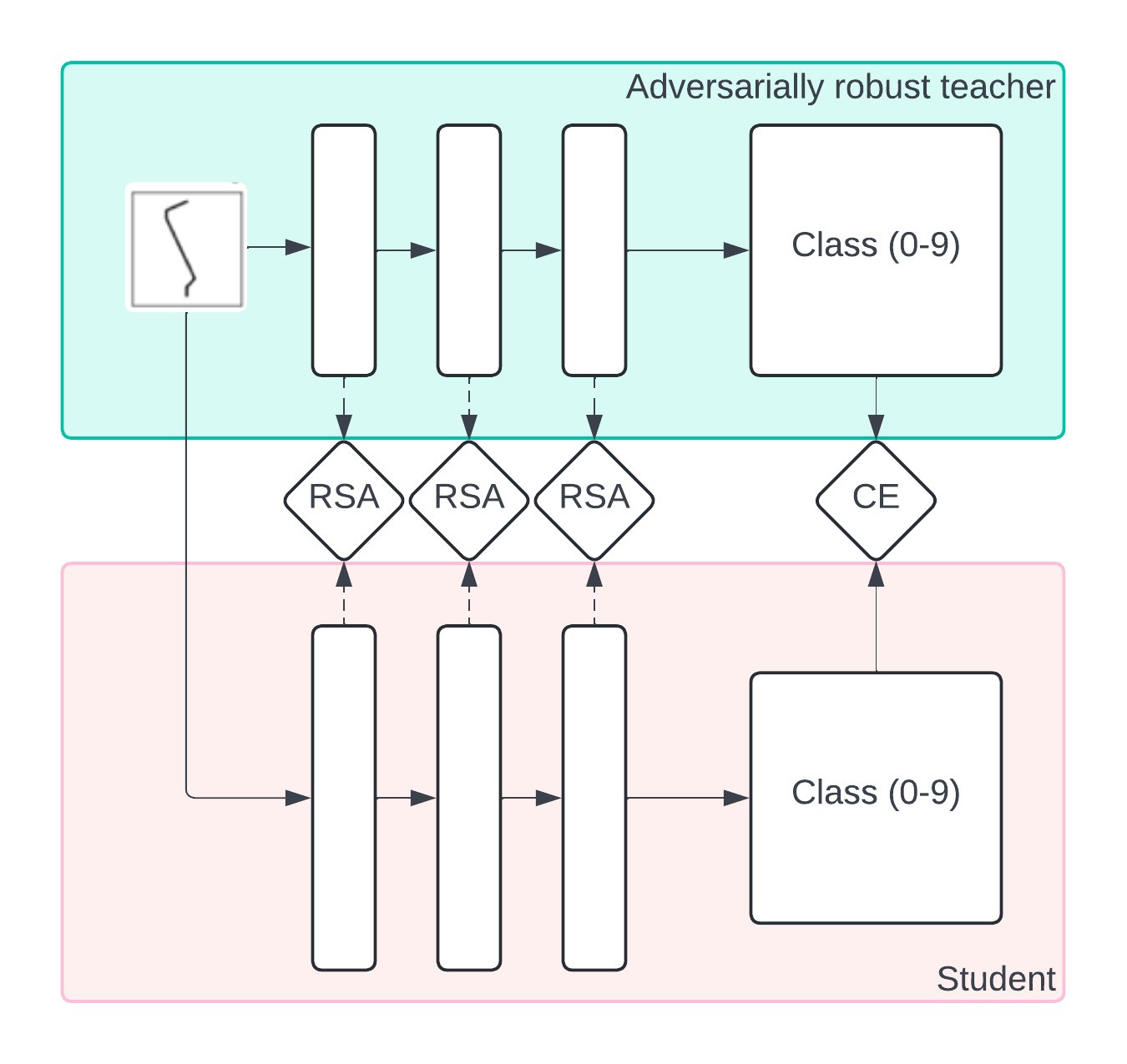}
    \caption{Setting for teacher-student simulations. We use an adversarially robust teacher and train a student to match either class labels only (outcome supervision) or both class labels and intermediate representations via RSA (outcome + process supervision).}
    \label{fig-teacher-student-simulation}
\end{figure}

To make progress on these questions, we turn to teacher-student
simulations, in which we test whether training a student to imitate a
robust teacher leads to robust representations. Here, the robust teacher
is a stand-in for the brain; however, unlike real brains, we can
generate arbitrary amounts of data from the teacher.

We leverage MNIST-1D, an algorithmically generated dataset that allows
us to generate arbitrarily large numbers of training examples and train
many networks very rapidly. We first train a small three-hidden-layer
CNN adversarially on large amounts of generated data, obtaining a robust
network. We then use this robust network as the teacher for a student
network. We consider two scenarios:

\begin{itemize}
\item
  The student is only trained on the teacher's labels. This is
  conceptually similar to the business-as-usual scenario of training
  only on labels.\item
  The student is trained on the teacher's labels and intermediate
  activity. We compute representational similarity matrices and match
  the intermediate activity of the teacher and student at their three
  convolutional layers on clean examples--we refer to this as a representational
  similarity analysis loss (RSA). This is conceptually similar to using
  neural data augmentation to match intermediate activations of teacher
  and student ~\cite{Li2019-dg}.
\end{itemize}

\begin{figure}[htbp]
   \centering
   \begin{subfigure}[b]{0.49\textwidth}
       \centering
       \includegraphics[width=3.2in]{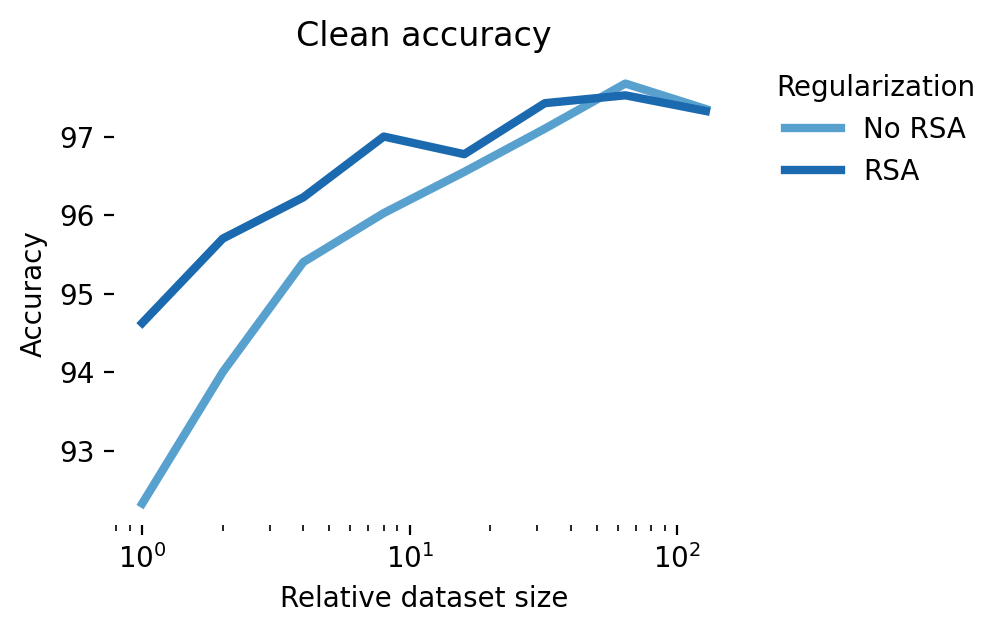}
   \end{subfigure}
   \hfill
   \begin{subfigure}[b]{0.49\textwidth}
       \centering
       \includegraphics[width=3.2in]{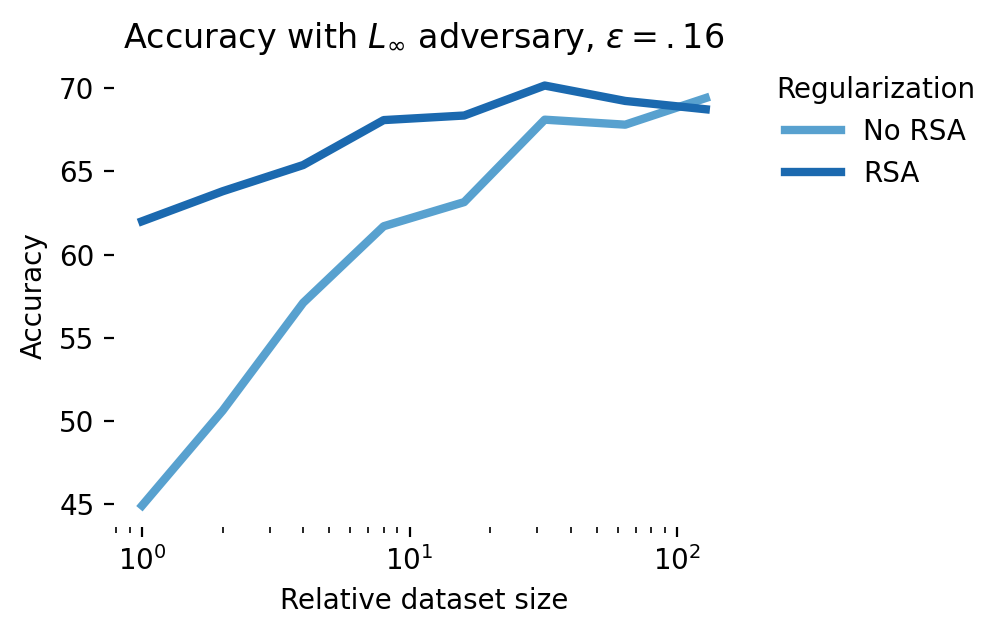}
   \end{subfigure}
   \caption{MNIST-1D accuracy under standard training with or without RSA regularization. Top: clean accuracy, bottom: Adversarial accuracy under an $L_\infty$ adversary. Matching inner representations improves clean and adversarial accuracy.}
   \label{fig-accuracy_rsa_vs_no_rsa}
\end{figure}

Both clean accuracy and adversarial accuracy under an adversary are
noticeably increased under these scenarios across all dataset sizes (Figure \ref{fig-accuracy_rsa_vs_no_rsa}). The
representation similarity loss forces the intermediate activations to
match those of the robust network, essentially forcing the student to
follow not just the labels of the robust network but also its strategy,
boosting in-distribution and adversarial generalization.

\begin{figure}[htbp]
    \centering
    \begin{subfigure}[t]{0.49\textwidth}
        \centering
        \includegraphics[width=3.2in]{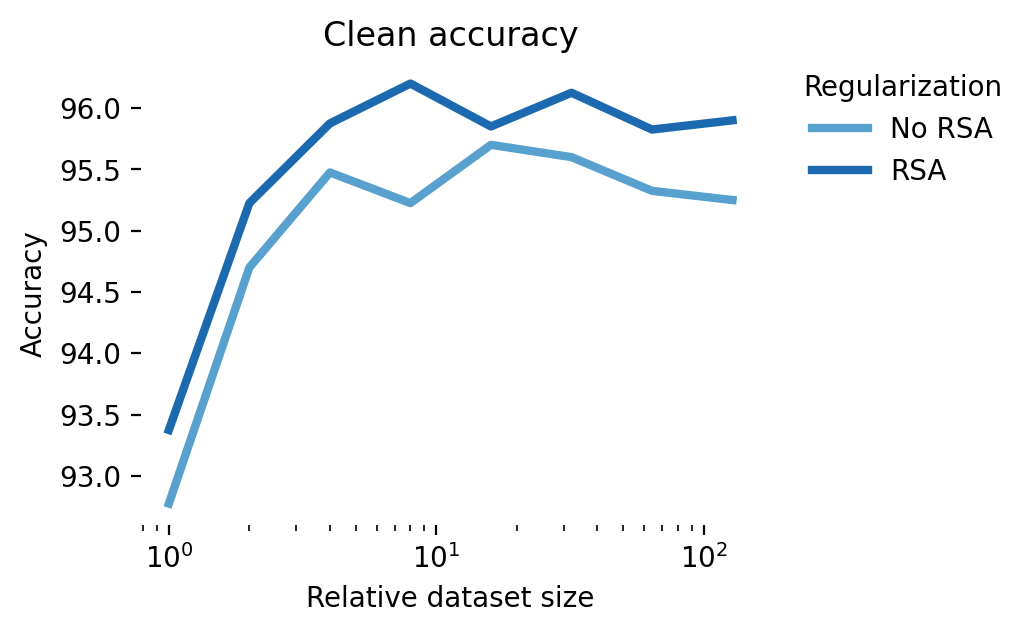}
    \end{subfigure}
    \begin{subfigure}[t]{0.49\textwidth}
        \centering
        \includegraphics[width=3.2in]{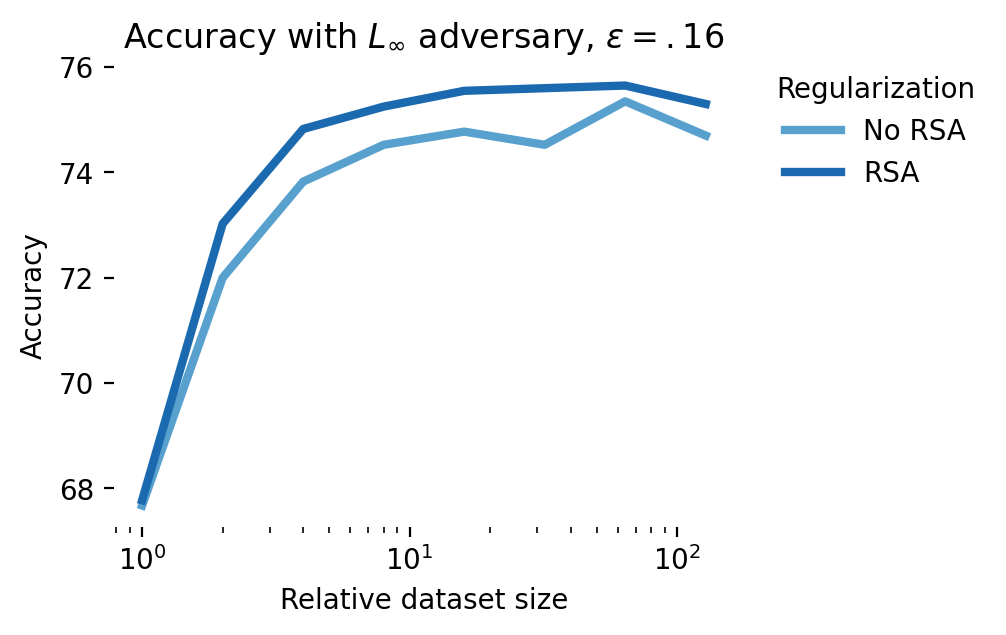}
    \end{subfigure}
    \caption{MNIST-1D accuracy under adversarial training, with or without RSA regularization. Top: clean accuracy, bottom: adversarial accuracy under an $L_\infty$ adversary. Matching representations is additive to adversarial training.}
    \label{fig-accuracy_adv_rsa_vs_no_rsa}
\end{figure}

These results are promising and in line with the NDA literature. To move
beyond a proof-of-concept, however, and positively impact AI safety,
we'd need to show that it's possible to do better than the
state-of-the-art, adversarial training. The results shown in Figure \ref{fig-accuracy_adv_rsa_vs_no_rsa} show that while NDA does better than standard training,
adversarial training offers a more direct route to robust students: at
the smallest dataset size, with an $\epsilon=0.16$ adversary, no
augmentation gives 45\% accuracy, RSA gives 62\% accuracy, and
adversarial training obtains 68\%. Combining RSA and adversarial
training in these simulations gives a small but consistent performance
boost of up to 1\% over adversarial training only. In the regime of
scaling laws for adversarial training, a 1\% improvement in adversarial
robustness could be matched with a \textasciitilde2X increase in FLOPs
\cite{Bartoldson2024-qr}.

\begin{figure}[htbp]
   \centering
   \begin{subfigure}[b]{0.44\textwidth}
       \centering
       \includegraphics[height=2.2in]{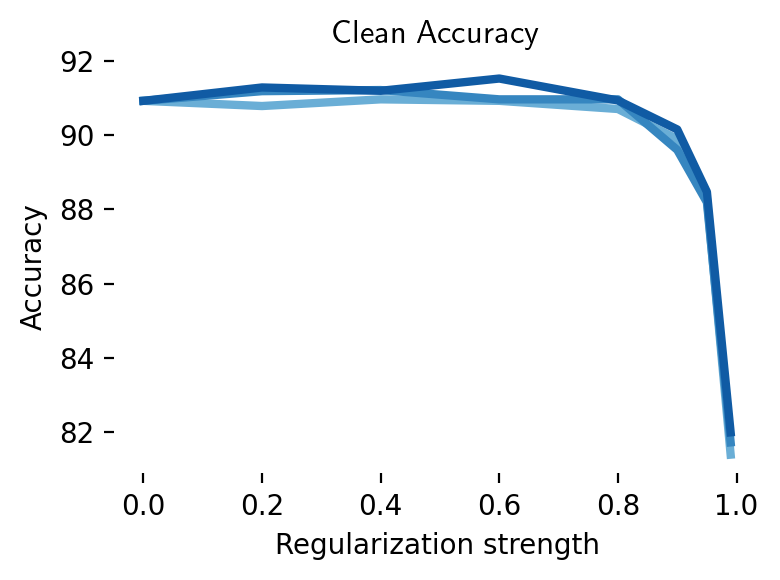}
   \end{subfigure}
   \hfill
   \begin{subfigure}[b]{0.53\textwidth}
       \centering
       \includegraphics[height=2.2in]{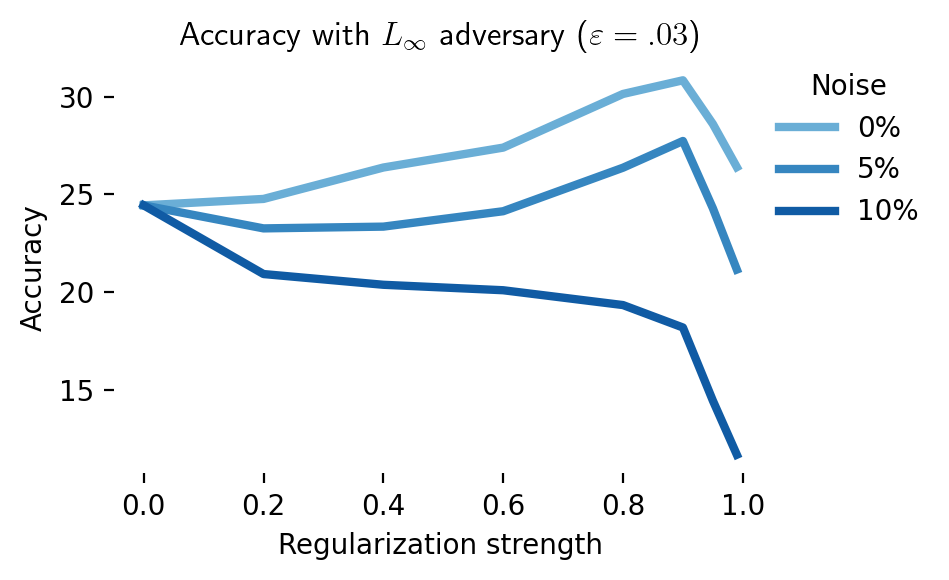}
   \end{subfigure}
   \caption{CIFAR-10 accuracy with RSA regularization at different noise regimes. Top: clean accuracy, bottom: adversarial accuracy under an $L_\infty$ adversary. Training the network with a loss that encourages the student to match the latent representation of the teacher can lead to more robust representations, provided that the activations are measured without noise.}
   \label{fig-cifar10-teacher-results}
\end{figure}

We extended these simulations to CIFAR-10 (Figure \ref{fig-cifar10-teacher-results}). Here again,
aligning latent representations with those of a robust teacher leads to
an improvement in robustness. However, this improvement in robustness is
diminished and eventually eliminated when activations are noisy. This
shows a potential role for digital twins trained on neural data in
improving adversarial robustness: a digital twin trained at multiple
stages of a sensory hierarchy can be used to essentially denoise latent
activations \cite{Li2019-dg}.

These results show there is a regime in which data from the brain could
help adversarial robustness, especially if future brain augmentation
methods can leverage brain data more effectively
\cite{Pirlot2022-jz}. It does, however, highlight a gap
in the literature: most papers surveyed here compare brain-augmented
training against vanilla non-robust networks, not robust networks. The
latter comparison is, in our opinion, more relevant for AI safety
applications. Direct engagement with state-of-the-art benchmarks
\cite{Croce2020-nk} is desirable. Furthermore, if the
solution to adversarial examples lies in the brain, one would want to
measure how \emph{adversarial} examples, as opposed to \emph{clean}
examples, affect representations at different levels in the sensory
hierarchy \cite{Guo2022-ef}.

While these results indicate a rather narrow \emph{direct} path for
neural data directly improving robustness in artificial neural networks, we remain optimistic that better understanding the computational and information processing principles of how robust representations are formed in the brain could \emph{inspire} new adversarially robust algorithms \cite{Fort2024-cp, Attias2024-pw}.

\hypertarget{subsec-evaluation-digitaltwins}{%
\subsection{Evaluation}\label{subsec-evaluation-digitaltwins}}

Digital twins of perceptual systems could potentially contribute towards
safer AI systems by:

\begin{enumerate}
    \item Allowing one to predict a subject's perception in reaction to a physical stimulus, which could facilitate human-AI interactions
    
    \item Distilling more robust representations than is currently feasible with data augmentation and adversarial training alone
    
    \item Enabling virtual and closed-loop experiments that tease apart the circuit mechanisms, inductive biases, and representation geometry underlying robust perception (see also Section \ref{sec-leverage-neuroscience-inspired-methods-for-mechanistic-interpretability} for a discussion on mechanistic interpretability)
\end{enumerate}

Scaling laws indicate that building digital twins that account for a
large fraction of the explainable variance in a particular area is
likely feasible with current technology in model species, while the path
to scale up to humans remains more speculative. The data used to train
digital twins may be useful to enhance adversarial robustness in
existing neural networks; furthermore, digital twins as an intermediate
step to denoise activations in neural data augmentation may lead to more
effective regularization than the direct approach of using noisy neural
data directly as a regularizer. To move beyond proofs of concepts, we
identify one bottleneck, the measurement of neural responses to
adversarial stimuli. We also single out better understanding the
geometry of human robustness to adversarial stimuli using large-scale
psychophysics as an important bottleneck to improving adversarial
robustness \cite{Bartoldson2024-qr}.

\hypertarget{subsec-opportunities-digitaltwins}{%
\subsection{Opportunities}\label{subsec-opportunities-digitaltwins}}

\begin{itemize}
\item
  Create large-scale neural recordings from sensory areas in response to
  rich spatiotemporal stimuli\begin{itemize}
  \item
    Focus on bringing down the required single-neuron recording times to
    reach high FEVE within a 3-4 hour recording session\begin{itemize}
    \item
      Improve efficiency of transfer function estimation by learning
      compact cores, or by using sparse readout mechanisms\end{itemize}
  \item
    Scale promising chronic recording technologies to allow recording
    beyond the 3-4 hour single-session limit\end{itemize}
\item
  Build sharable, composable digital twins of sensory systems\begin{itemize}
  \item
    Measure and report scaling laws on an apples-to-apples basis, for
    both single-neuron transfer function estimation and the ceiling
    attainable by a core\item
    Share pretrained models in a standardized format to bootstrap the
    creation of fine-tuned models\item
    Expand beyond vision to other modalities, including audition and
    somatosensation and more motor variables\end{itemize}
\item
  Build robust digital twins\begin{itemize}
  \item
    Measure in-vivo neural responses to adversarial stimuli in a
    closed-loop fashion\item
    Build robust digital twins through adversarial training and
    anchoring to measured neural responses to adversarial stimuli
  \item
    Estimate the geometry of adversarial robustness in humans through
    large-scale, online psychophysics\item
    Track and report robustness of digital twins to adversarial and
    out-of-distribution stimuli across a range of relevant attack
    dimensions (e.g. $L_\infty$ and $L_2$
    robustness, distributional shifts)\begin{itemize}
    \item
      Compare against state-of-the-art defenses rather than against
      non-robust training\end{itemize}
  \end{itemize}
\end{itemize}

\clearpage

\begin{namedbox}{scaling-trends}{Scaling trends for neural recordings}

Many of the approaches to NeuroAI safety that we discuss are dependent
on the continued creation of ever-larger datasets with more capable
recording techniques. For instance, under log-linear scaling laws, to
obtain linear improvements in performance over time, one would need to
obtain datasets that grow exponentially over the same time. Stevenson
\cite{Stevenson2011-sr} first documented a
Moore's-law-like relationship for recording capabilities in
electrophysiology, estimating that simultaneously recorded neurons
double every 7 years; Stevenson recently updated his estimate at 6
years. Urai et al. \cite{Urai2022-eh} extended these
results to calcium imaging; while the growth was not quantified, the
trend pointed toward exponentially faster improvements in calcium
imaging.

We extended these results to update these scaling trends using an
LLM-based pipeline to identify relevant papers (see methods for details)
and joining them with previous databases collected by Stevenson and
Kording and Urai et al. Fitting a Bayesian linear regression to the log
number of neurons for recordings after 1990, we obtained an estimated
5.2±0.2 year doubling time for electrophysiology and 1.6±0.2 year
doubling time for imaging.

Out of the 10 electrophysiology studies reporting the largest
simultaneous neuron numbers recorded, 9 were performed with multi-shank
Neuropixel recordings. Light bead microscopy is currently the
state-of-the-art for calcium imaging
\cite{Manley2024-xn}. If current trends continue, we
predict that in 2035, 10,000 neuron electrophysiological recordings will
be commonplace, whereas calcium imaging should reach 10M neurons. A concerted investment in neurotechnology could break these incremental trends, in particular for electrophysiology.  

These trends are important to contextualize how digital twins will
evolve; indeed, scaling laws for digital twins suggest linear
improvements in our ability to predict neural activity with exponential
increases in channel count. However, this analysis does not assess the
signal-to-noise ratio (SNR) of different recording technologies. While
this is more or less constant in electrophysiology, SNR can vary widely
in imaging depending on the amount of out-of-focus light and the size of
the focus area (e.g. one-photon vs. multiphoton imaging approaches).
Thus, it's likely that we'll see optimizations in terms of
signal-to-noise ratio rather than increasing channel count as we
approach whole-brain coverage with imaging.

\begin{center}
\setlength{\fboxsep}{0pt}  
\setlength{\fboxrule}{0.5pt}  
\colorbox{white}{
    \centering
    \includegraphics[width=3.0in]{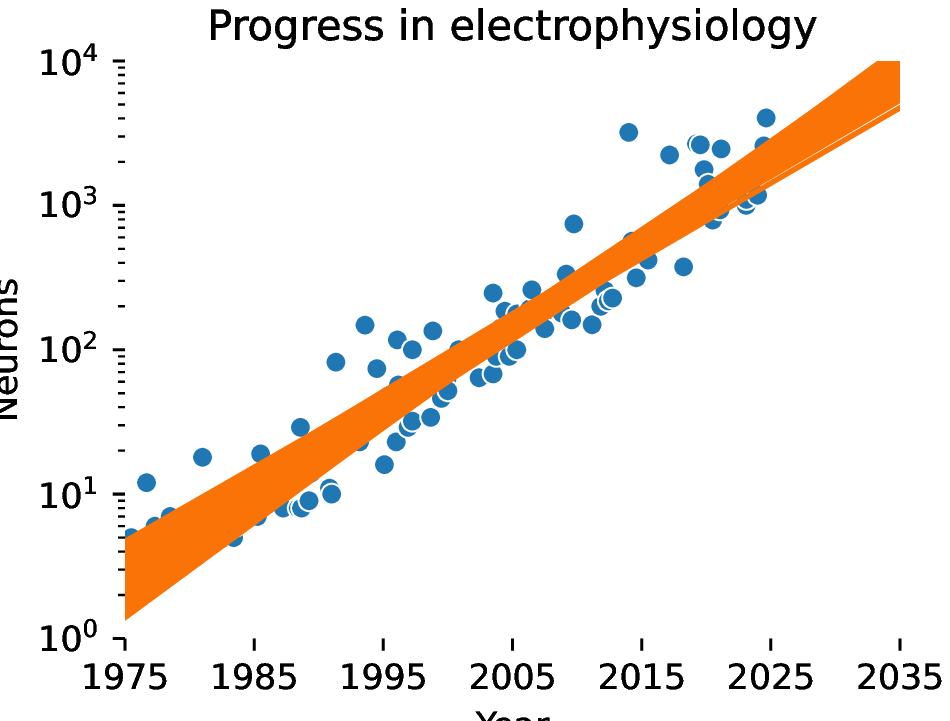}
    \includegraphics[width=3.0in]{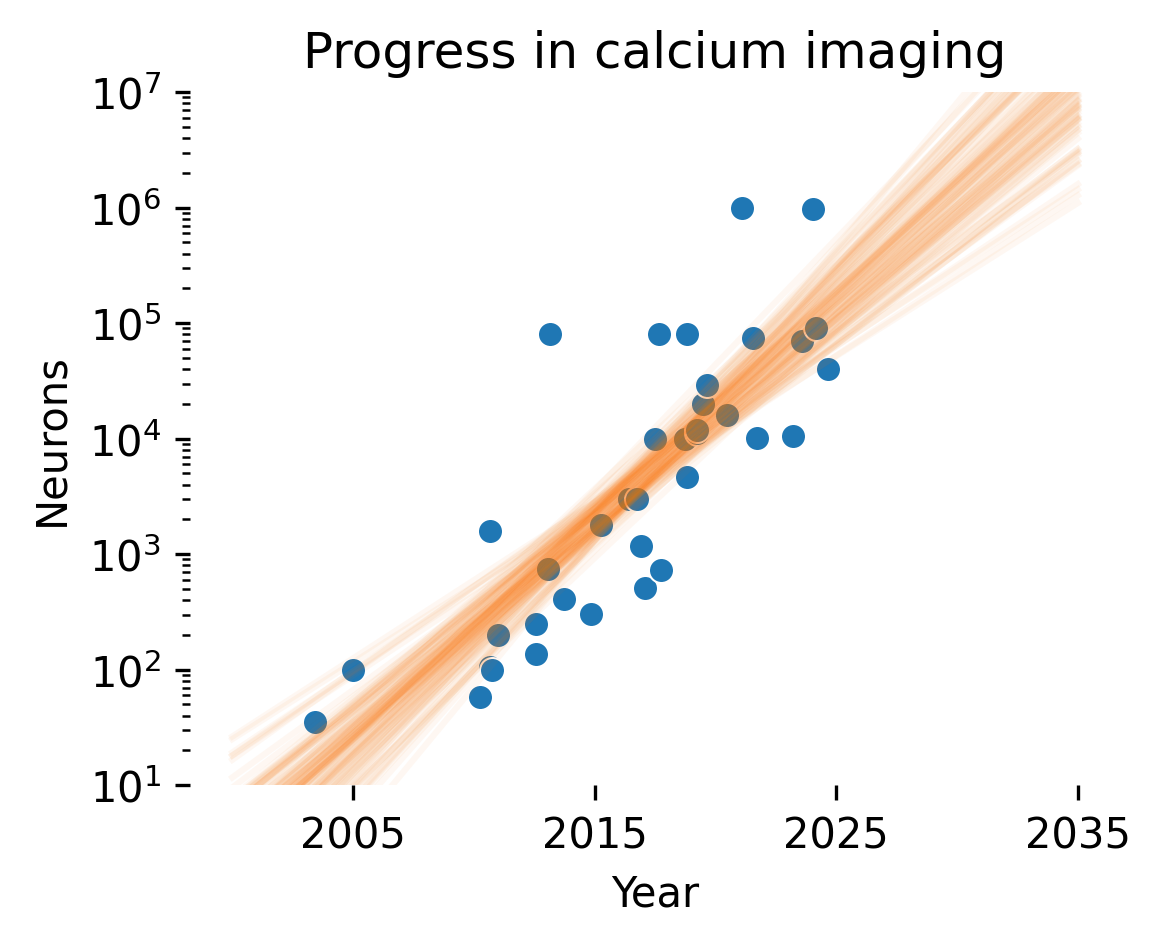}%
}
\end{center}
\end{namedbox}

\hypertarget{sec-embodied}{%
\section{Build embodied digital twins}\label{sec-embodied}}

\hypertarget{sec-core-idea-embodied}{%
\subsection{Core idea}\label{sec-core-idea-embodied}}

In the previous section, we saw a path toward building sensory digital twins that learn the transfer function between sensory inputs and brain state. Could we leverage similar ideas to build a digital twin of the entire brain and body at the functional level? \textbf{Embodied digital twins} aim to model human behavior, embodiment and neural activity at a
coarse level without necessarily replicating intricate neural circuits. If they are accurate models of the humans which they aim to imitate, we may derive relevant safety properties from them. This includes the ability to safely explore the world and control their bodies, as well as display the same inductive biases as humans.

An embodied digital twin could be built with a patchwork of approaches: leveraging a pre-existing sensory digital twin; learning a model for the brain and body through self-supervised learning on neural activity and behavior, acting as a base controller; embodying that model in a virtual environment; and fine-tuning the controller in silico in virtual environments. Embodied digital twins would be built with many of the same building blocks as conventional AI systems, which could facilitate synergistic interactions between AI and neuroscience \cite{Richards2019-by}. We evaluate the technical feasibility of embodied digital twins here.

\hypertarget{subsec-why-does-it-matter-for-ai-safety-and-why-is-neuroscience-relevant-embodied}{%
\subsection{Why does it matter for AI safety and why is neuroscience
relevant?}\label{subsec-why-does-it-matter-for-ai-safety-and-why-is-neuroscience-relevant-embodied}}

Building a biophysically detailed simulation of people at the brain and body level has long been entertained as a speculative, brute-force path toward artificial general intelligence \cite{Bostrom2008-og, Bostrom2014-gb}. In the AI safety community, biophysically detailed bottom-up simulations of neural activity are often termed WBEs--whole brain emulations \cite{Bostrom2008-og, Bostrom2014-gb}. Bostrom (2014) lists three reasons why biophysically detailed simulations are a safer path to AGI than alternatives:

\begin{enumerate}
    \item \emph{Mutual understanding}. Because simulations are derived from humans, with whom we have extensive experience, we should be able to reason about their behavior, in the same way we leverage theory-of-mind and empathy to reason about humans. Furthermore, they should be able to reason about humans.
    \item \emph{Inheriting human values}. Because simulations are derived from humans, they should have similar motivations and values as humans.
    \item \emph{Slower takeoff}. Because biophysically detailed simulation is a brute-force approach to AGI, it requires massive investments in neural recording and scanning hardware. We should be better able to forecast when biophysically detailed simulations are likely to arrive and prepare accordingly, compared with alternative paths which could be massively accelerated through scientific insight.
\end{enumerate}

It's easy to find flaws in each of these lines of reasoning. We may not be able to reason about simulations that are run at much faster speeds than humans. Inheriting human motivations and values can be a double-edged sword, as these may include undesirable properties like aggression and power-seeking. And takeoff could be faster than anticipated if coarser-grained simulations can accelerate subsequent steps in building the next generation of simulations. 

However, simulations need only be \emph{relatively} safer than alternative paths toward artificial general intelligence for them to represent an improvement over the status quo \cite{Sandbrink2022-nm}. Other desirable properties of human brains that simulations could inherit include the ability to cooperate, theory of mind, and out-of-distribution robustness. Economists have fleshed out how a society with human emulations would function \cite{Hanson2016-ng}. A 
recent workshop identified emulation as potentially the most impactful
neuroscience-related approach to AI safety, yet it also identified it as
the least technically feasible \cite{Thiergart2023-la}.

This state of affairs--high impact, but low feasibility--has motivated
some to look for more tractable alternatives. To a certain extent, large language
models (LLMs) are a proof-of-existence that imitation learning of human behavior followed by fine-tuning can display emergent capabilities
\cite{Bubeck2023-pz}. Could imitating neural data, body and behavior lead to an effective simulation of human capabilities and safety properties, thus leading to an alternative path toward highly capable AI?
\emph{Embodied digital twins} describes brain and body simulations learned through imitation learning of brain data
and behavior. In the following, we define the technical challenge in building an embodied digital twin, defining its scope and associated success criteria.

\hypertarget{subsec-details-embodied}{%
\subsection{Details}\label{subsec-details-embodied}}

\hypertarget{subsubsec-defining-embodied-wbe-and-related-concepts}{%
\subsubsection{Defining embodied digital twins}\label{subsubsec-defining-embodied-wbe-and-related-concepts}}

What do we mean by an embodied digital twin? In a footnote, Bostrom and
Sandberg \cite{Bostrom2008-og} define an effective simulation of an embodied nervous system 
as being able to predict the future state of the brain and body $x(t)$ at all future times $t > T_0$ given
the state of the system at the current time, $x(T_0)$, within an $\epsilon$ bound. The state can be defined at different levels of abstractions, but can include:

\begin{itemize}
\def\labelenumi{\arabic{enumi}.}
\item
  Neural activity, e.g. membrane voltage or spike rate
\item
  Auxiliary state variables related to neural activity, e.g. a neuron's
  relative refractory state, the state of neuromodulators, etc.
\item
  The position of the limbs, the load on each muscle, and their
  velocity; collectively, behavior
\item
  The system's connectome, reflecting the accumulated memories of the organism
\end{itemize}

An alternative implementation conditions the simulation not just on the current time step $T_0$  but on several time steps in the past $t \le T_0$, which is attractive from the point of view of Takens’ theorem \cite{Brunton2019-vv}. 

By that yardstick, both embodied digital twins and biophysically detailed models attempt to build simulations of the brain and body at different levels of granularity. An embodied digital twin primarily leverages behavior, functional neural recordings and body measurements to build a simulation from the top-down, while a biophysically detailed model primarily leverages structural recordings and detailed biophysical modeling to build a simulation from the bottom-up.

Several recent proposals and papers help clarify what an embodied digital twin might look like:

\paragraph{AI animal models and the embodied Turing Test} Zador et al. \cite{Zador2022-hd} propose to build a simulation of an entire animal \emph{in silico}--an AI animal model. Comparisons between real and virtual animals would provide a readout of how well a simulation performs. A virtual animal would pass the 'embodied Turing test' if it is indistinguishable from its living counterpart when observed in a virtual environment.

\paragraph{The virtual rat} Merel et al. \cite{Merel2019-kb} demonstrated that training a virtual rat using reinforcement learning in a virtual environment could display rich behavior. Aldarondo, et al. \cite{Aldarondo2024-is} trained an ANN to actuate a biomechanical model of a more advanced version of this virtual rat. When the virtual agent was tasked with imitating real rats, activations in the ANN showed striking similarities with neural activations in the control circuits of the real rats. 

\paragraph{The virtual fly} Recent advances in Drosophila connectomics \cite{Dorkenwald2023-xt, Azevedo2024-dc} enable simulation of its entire nervous system (130K brain neurons plus 15K in the ventral nerve cord). While some groups pursue direct numerical simulation of this nervous system \cite{Shiu2024-sq} with detailed sensorimotor integration \cite{Vaxenburg2024-cv}, others take a functional view, virtualizing sensory systems in simplified virtual environments \cite{Lappalainen2023-ot, Cowley2024-dg}.

\paragraph{The virtual \emph{C. elegans}} With its fully mapped 300-neuron connectome \cite{Cook2019-gd}, \emph{C. elegans} offers a unique testbed for contrasting simulation approaches \cite{Haspel2023-zf}. Biophysically detailed models could be constrained by the connectome and neural recordings, while a proposed embodied digital twin uses transformers to predict neural activity and behavior from past observations \cite{Simeon2023-ru}.

\bigskip

Note the central role of body simulations in these recent approaches. Neural circuits evolved for sensorimotor control--enabling predation, escape, and survival--must solve computational problems fundamental to general intelligence: generalization across domains, continual learning without catastrophic forgetting, and translating abstract plans into concrete actions. Human-level intelligence likely emerged by co-opting these ancient sensorimotor control circuits \cite{Moravec1988-zy}.

The recent efforts that we list point toward common themes in embodied digital twins: predicting next-step neural activity from the past; virtualizing the bodies of animals; virtualizing the senses of animals; and complementing bottom-up reconstructions of neural activity with virtual training. We turn our attention to each of these in turn.

\hypertarget{subsubsec-foundation-models-for-neuroscience}{%
\subsubsection{Foundation models for
neuroscience}\label{subsubsec-foundation-models-for-neuroscience}}

A key artifact for an embodied digital twin is building a model that
predicts the future state of a neural system
conditioned on past observations. This naturally links to
current state-of-the-art models for text generation
\cite{Brown2020-bk,OpenAI2023-qq,Dubey2024-ee}: black
box models, like transformers or state-space models (SSMs), which are
trained to autoregressively predict the future, and can be used as
generative models. Building on this naturally allows one to leverage the
rich literature and accumulated wisdom on training and fine-tuning
foundation models \cite{Bommasani2022-on}, as well as
capable software for training ANNs at a large scale
\cite{Paszke2019-by}.

A growing literature in neuroscience aims to accomplish an analogous
task: finding good representations of neural activity
\cite{Wang2023-dr}. Powerful models, trained through
supervised, self-supervised or unsupervised learning, can learn
effective representations of neural activity
\cite{Pandarinath2018-gl,Ye2021-ek,Hurwitz2021-ml,Urai2022-eh,Azabou2023-ja,Zhang2024-hb,Schulz2024-do}.
These models have become an integral part of the toolkit of systems
neuroscience, allowing one, for example, to estimate single-trial spike
rates from noisy neural data. \emph{Foundation models for neuroscience},
built on generic architectures like transformers, continue this
tradition, pretraining large-scale models that find good generic
representations of neural activity, which may be used for
downstream tasks \cite{Wang2023-dr}. Models trained with
causal masking or as autoencoders can be used to simulate neural
activity.

Several promising models of this flavor have been recently demonstrated
for spikes \cite{Azabou2023-ja}, binned spikes
\cite{Ye2021-ek,Zhang2024-hb}, fMRI
\cite{Caro2023-fg}, iEEG
\cite{Chau2024-uw}, ECG
\cite{Abbaspourazad2024-tg}, EEG
\cite{Jiang2023-kb}, MEG
\cite{Csaky2024-wd} and EMG
\cite{Labs2024-gh}. These models face challenges unique
to neural data, including registration, alignment, tokenization, and
discretization, which have been met in increasingly sophisticated ways.
Several recent manuscripts have demonstrated the classic signature of
scaling laws: linear improvements in performance with exponential
increases in data size.

Building autoregressive foundation models for neuroscience thus relies
on:

\begin{enumerate}
\def\labelenumi{\arabic{enumi}.}
\item
  Getting access to abundant neural and behavioral data (Box \ref{box-available-neural-data}) to train
  large-scale models
\item
  Finding good schemes to curate, filter, embed, register, align, and
  tokenize the data
\item
  Training foundation models with generic models to predict activity in
  the future using next-token prediction or a related criterion
\item
  Evaluating the autoregressive predictive model using suitable criteria
\end{enumerate}

\begin{namedbox}{available-neural-data}{Availability of neural data to train a large-scale model}
The past decade has seen an explosion in the quantity of neural data
freely available online. These public datasets
represent a unique opportunity to learn good representations of neural
data for a variety of downstream tasks, including brain-computer
interfaces, clinical diagnoses for computational psychiatry and sleep
disorders, and basic neuroscience.

Here we present a breakdown of the available data sources from an
analysis of the contents of DANDI, OpenNeuro, iEEG.org, as well as large-scale individual datasets. Some of the highlights from
this analysis include:

\begin{itemize}
\item
  There are around 100,000 hours of neural data available in freely
  accessible archives.
\item
  There are roughly 3.3 million neuron-hours of single-neuron recordings
  from animals.
\item
  The most abundant data type in terms of number of hours is
  intracortical EEG in humans--an invasive modality generated from the typically continuous,
  week-long recordings performed during epilepsy monitoring.
\item
  Single neuron data is concentrated in a few datasets; the top 10
  largest datasets in terms of neuron-hours account for more than 94\%
  of total neuron-hours across all of DANDI. These come mostly from
  zebrafish and mouse, with one dataset from macaques.
\item
  Large fMRI recordings are split into two categories: broad neuroimaging surveys, including HCP and UK Biobank, which scan many people for a short time; and intensive neuroimaging datasets \cite{Kupers2024-sf}, including Courtois Neuromod and the Natural Scenes Dataset, which scan few people for a very long time.
\end{itemize}

\includegraphics[width=3.3in]{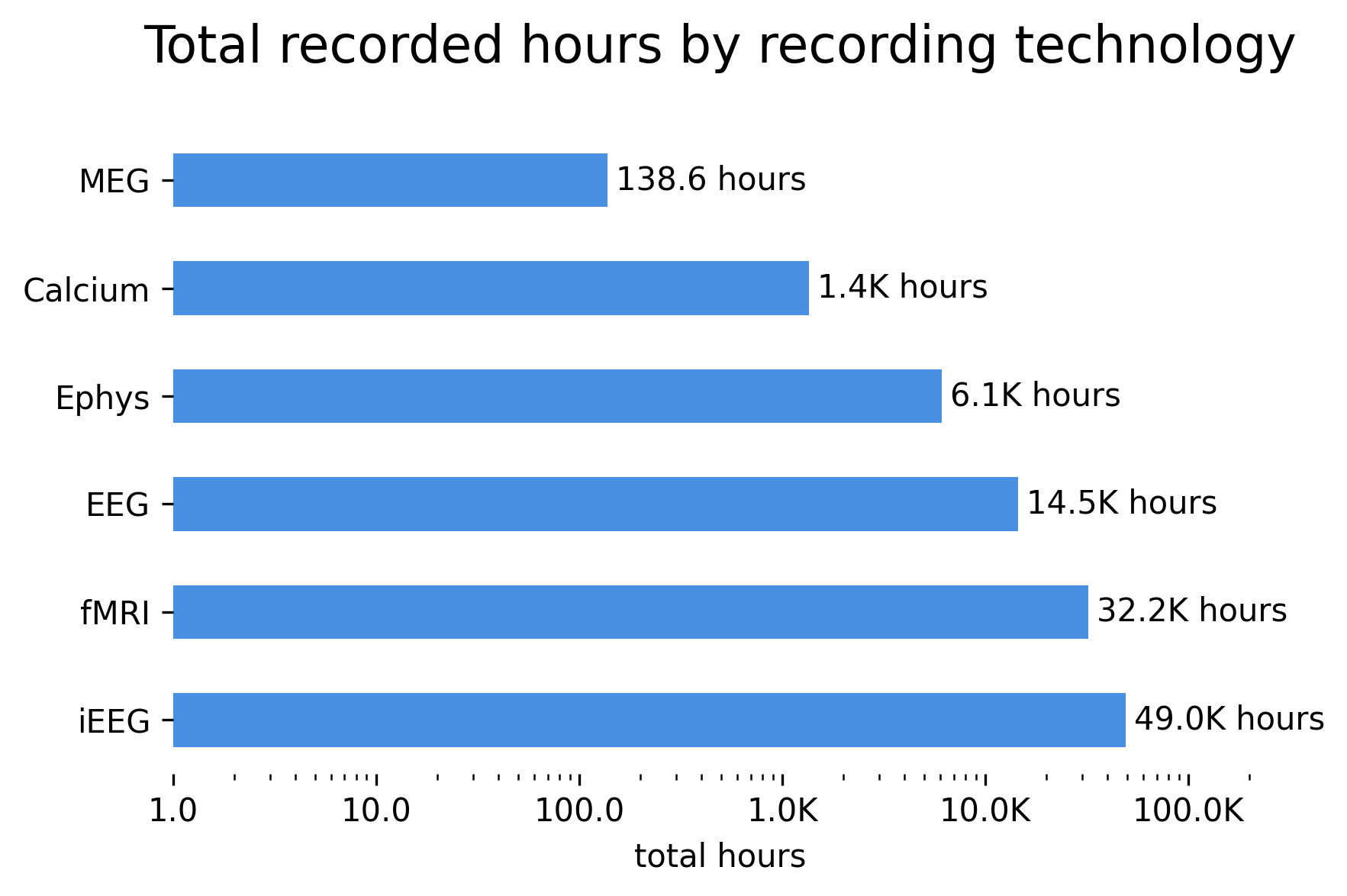}
\includegraphics[width=3.3in]{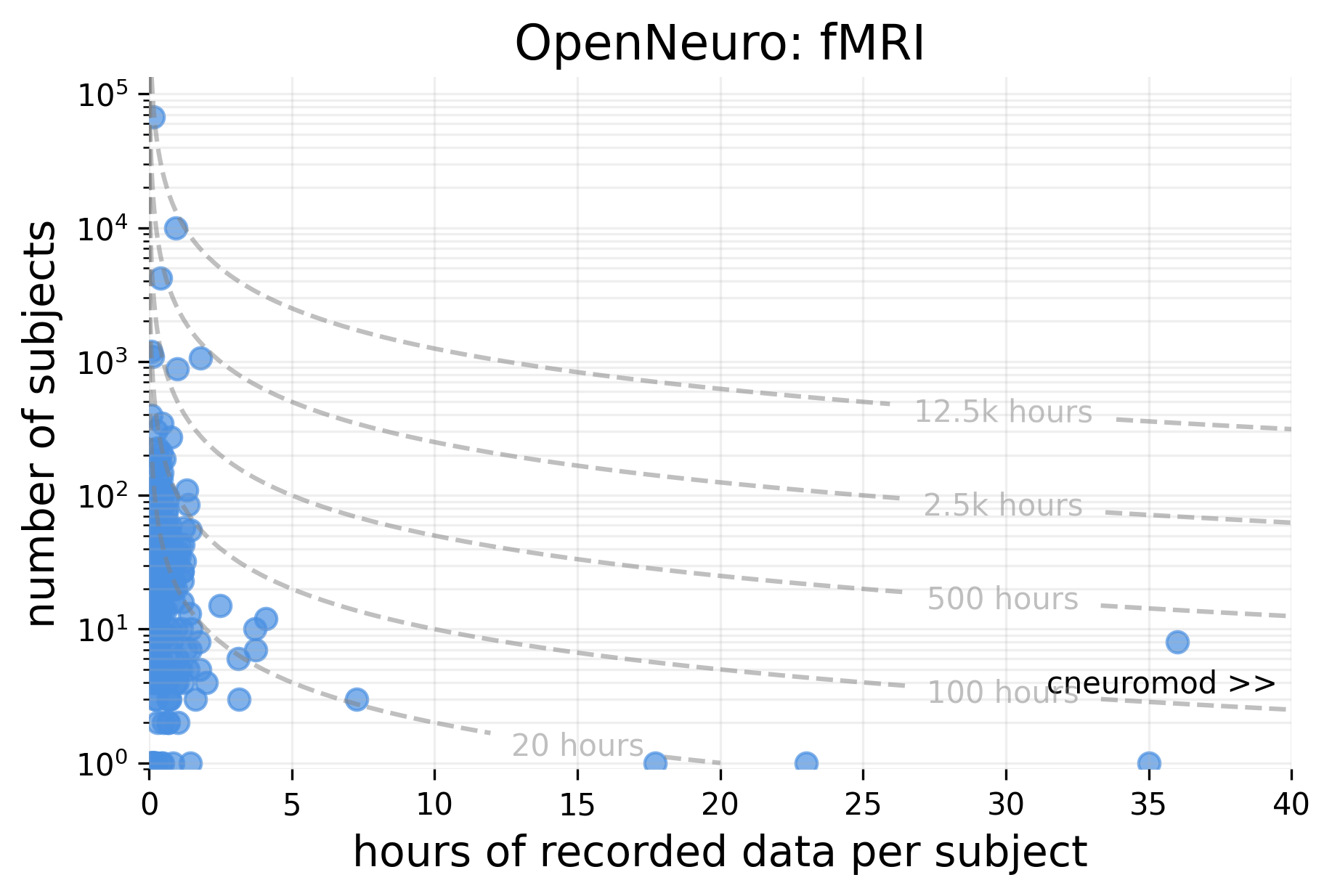}
\includegraphics[width=3.3in]{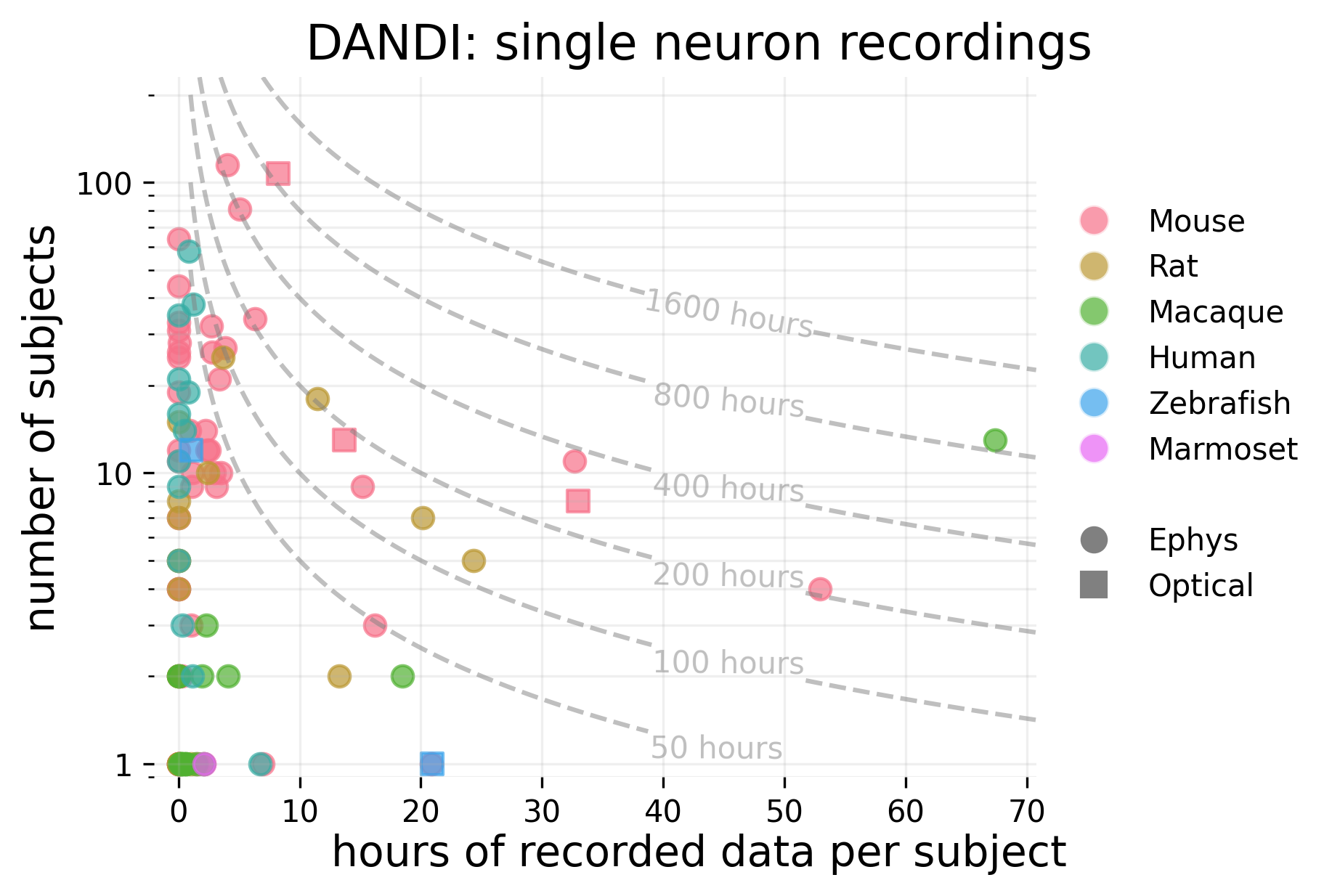}
\includegraphics[width=3.3in]{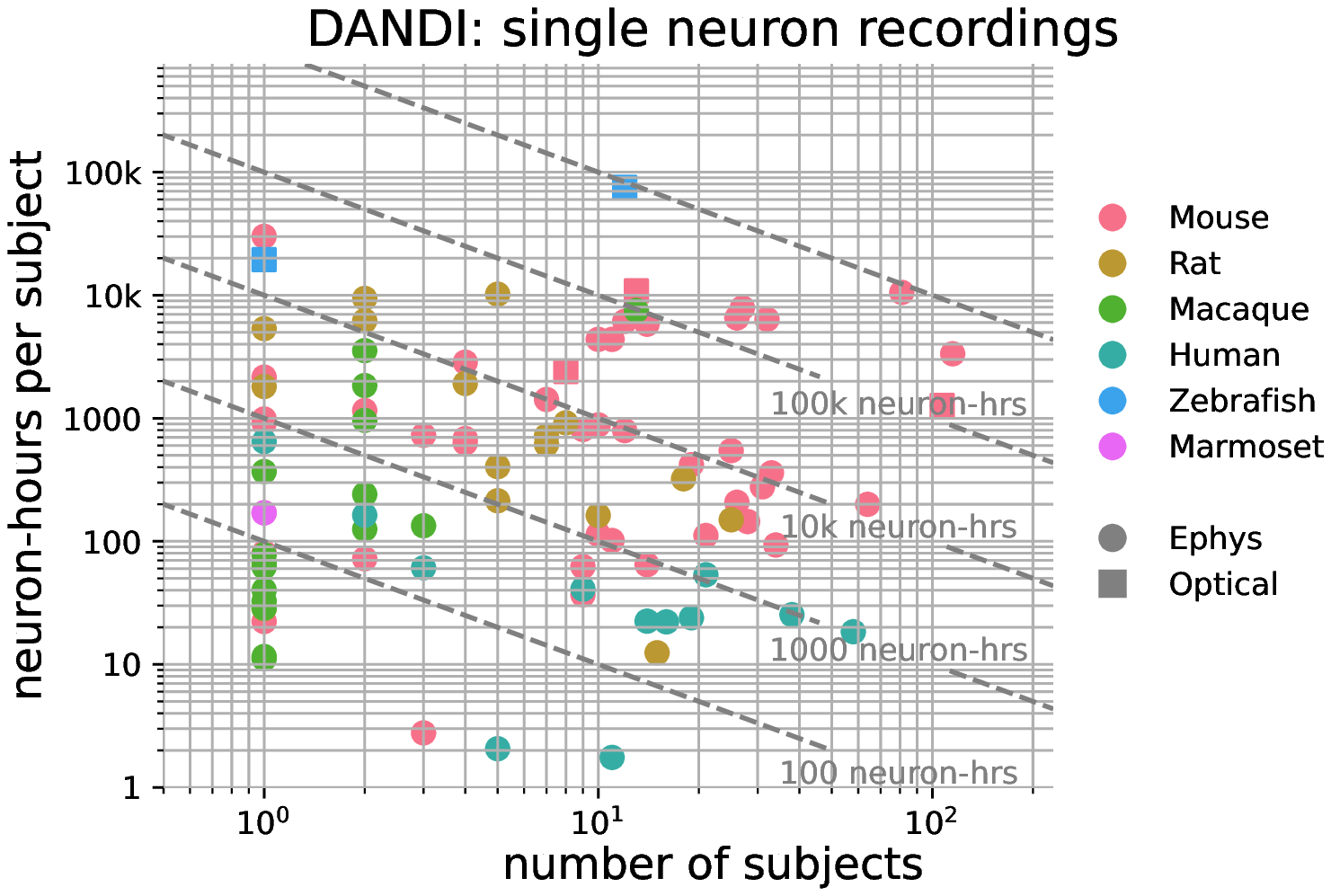}

\end{namedbox}

Notably, many unimodal foundation models in neuroscience use a
significant proportion of the openly available data
\cite{Yang2023-dw,Abbaspourazad2024-tg,Thomas2023-nk,Thapa2024-gs}.
For some modalities, progress will rely on unearthing existing, closed
datasets; acquiring new datasets; algorithmic improvements
\cite{Ho2024-wg}; and stitching data from multiple
modalities together. Ideal datasets should leverage high-entropy stimuli and behavior and measure responses across sensory and motor domains to learn good foundation models of the brain.

In addition to the sheer size of datasets, we must also consider their
quality. Coverage, the percentage of an animal's neural activity that is
recorded, is a critical metric. Currently, we can capture a large
proportion of the activity on single neurons in vivo for \emph{C. elegans}
\cite{Kato2015-nh} and zebrafish
\cite{Ahrens2013-vv, lueckmann2025zapbench} , but this is a challenge for other
model organisms. Thus, foundation models of species of
interest--including rodents, primates, and ultimately humans--may rely
on stitching the activity from many sparse neural activity recordings
into an overarching brain simulation. Helpfully, neural activity is of a
lower dimension than the number of distinct neurons in the brain
\cite{Gao2017-ze,Jazayeri2021-rk}, but the dimensionality of neural activity remains an unresolved question
\cite{Stringer2018-gi,Manley2024-xn}. This measure is also likely to vary across modalities and depend on the task being represented. We analyze the
dimensionality of neural activity across different model organisms in
Box \ref{box-neural-dim}, an important constraint in designing sparse neural recording
systems.

\newpage

\begin{namedbox}{neural-dim}{Dimensionality of neural data across model organisms}

How high dimensional is neural activity? This is an important quantity
as we seek to emulate it. Higher dimensional neural activity requires
higher capacity networks to fully capture. How dimensionality scales
affects the minimum number of neurons we need to sample from to fully
capture the brain. The ideal scenario, of course, is to capture every
neuron; but simple extrapolation of scaling trends in neural recording
(see Box \ref{box-scaling-trends}) would predict this capacity will not become feasible over
relevant AI safety timelines. Therefore, we must currently rely on
subsampling, hoping that the captured activity is representative of the
neural system as a whole.

We estimated the dimensionality of neural activity in three model
organisms based on previously recorded data: \emph{C. elegans}, zebrafish, and
mouse. We used SVCA \cite{Williams2018-jj,Stringer2018-gi,Manley2024-xn}, a metric which estimates the robustly measured number of principal components in neural data (see Methods for details).

Of note, all estimates of dimensionality tend to follow a power law as a
function of the number of measured neurons. Using a simple linear
regression estimator in log-log space, we find estimates of the exponent
ranging from 0.5 for \cite{Chen2018-um} to 0.91 for \cite{Stringer2018-gi}. The total number of dimensions measured is undoubtedly limited
by the richness of the behavior and stimuli
\cite{Gao2017-ze}, as well as the number of timesteps used for the estimation. 

With these caveats in mind, the number of well-estimated
dimensions in neural activity is always far lower than the number of
neurons--roughly 30 for \emph{C. elegans} (10\% of the number of neurons),
hundreds in zebrafish (\textless.1\% of neurons), and, by extrapolation,
anywhere between \textasciitilde10k \cite{Manley2024-xn}
to \textasciitilde1M \cite{Stringer2018-gi} in mice. The spread
in the range of estimates in mice is notable; we speculate this is
likely a difference between spontaneous and sensory-driven activity.
This might be grounds for using different strategies to build predictive
models in sensory cortex (e.g. digital twins for the visual system) vs.
elsewhere in the brain (e.g. autoregressive models).

\setlength{\LTleft}{0pt plus 1fill}
\setlength{\LTright}{0pt plus 1fill}

\begin{longtable}{p{0.1\textwidth}p{0.2\textwidth}p{0.25\textwidth}p{0.25\textwidth}}
\caption{}\label{tab-neural-dim} \\
\toprule
\textbf{Paper} & \textbf{Organism} & \textbf{Behavior} & \textbf{Recording tech} \\
\midrule
\cite{Yemini2021-fd} & \emph{C. elegans} & Chemical sensing & Calcium imaging \\
\cite{Suzuki2024-ar} & \emph{C. elegans} & Spontaneous & Calcium imaging \\
\cite{Chen2018-um} & Larval zebrafish & Visual stimuli & Calcium imaging \\
\cite{Stringer2018-gi} & Mouse & Visual processing, spontaneous & Calcium imaging, Neuropixels \\
\cite{Stringer2021-yg} & Mouse & Visual discrimination, processing & Calcium imaging, Neuropixels \\
\cite{Manley2024-xn} & Mouse & Spontaneous & Calcium imaging (light bead microscopy) \\
\midrule
\end{longtable}

    \includegraphics[width=6.2in]{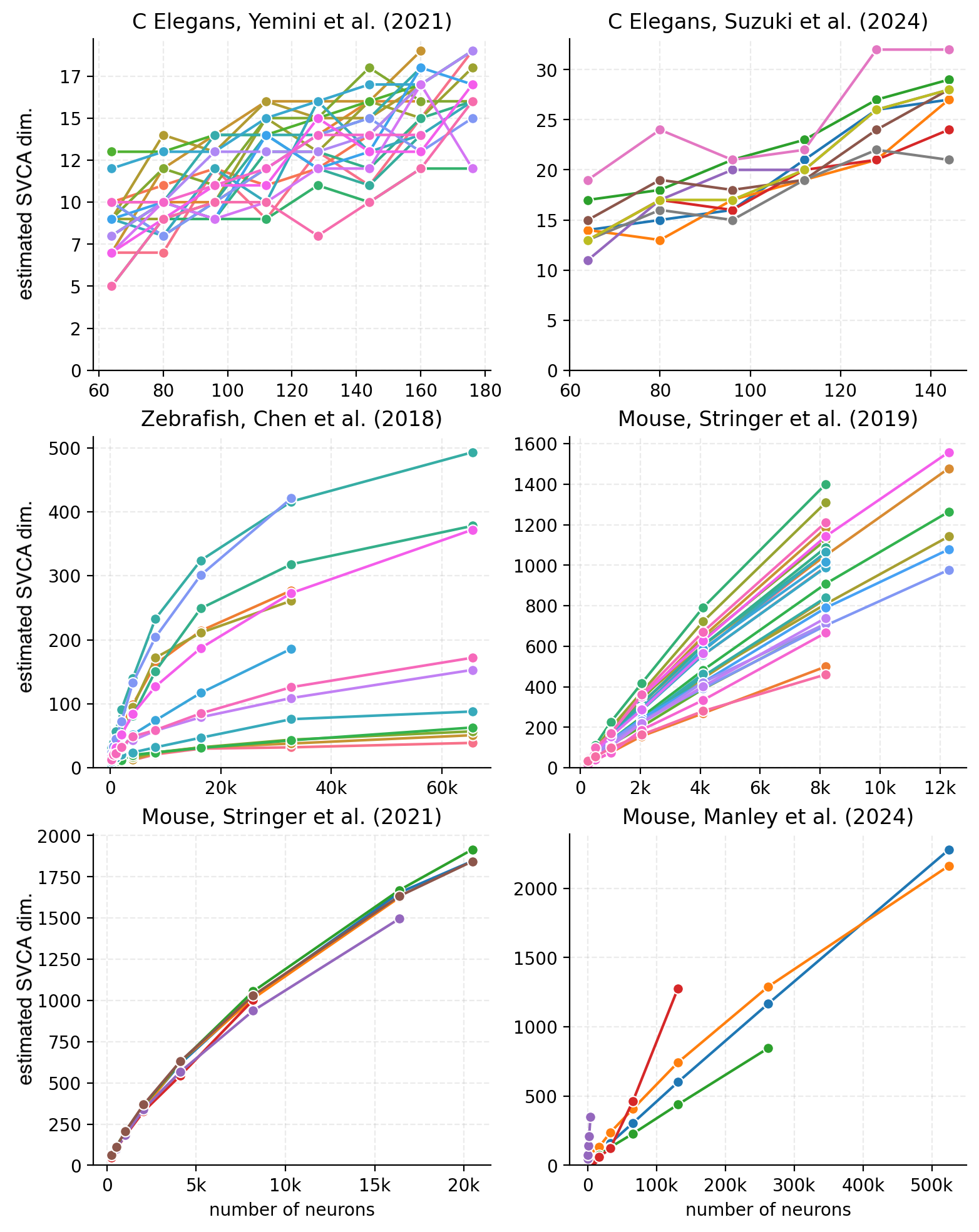}
    \newline
    \emph{Estimated SVCA dimension as a function of number of neurons included in the analysis. Each line represents a single experiment.}
\end{namedbox}

\hypertarget{subsubsec-virtualizing-animals}{%
\subsubsection{Virtualizing animals}\label{subsubsec-virtualizing-animals}}

While autoregressive models of neural activity--foundation models for
neuroscience--are instrumental in advancing embodied digital twins, they are not
sufficient. These models predict future neural states based on past
neural activity and exogenous inputs, neglecting the critical feedback
loops between the organism\textquotesingle s behavior, the environment,
and subsequent sensory inputs
\cite{Friston2010-wp,Kluger2024-uu}. To fully capture
the causal relationship between an organism and its environment, we need
to simulate how brain states affect the body, how the body interacts
with the world, and how environmental changes influence sensory inputs.
In other words, we need to embed autoregressive models in virtual bodies
and learn to control virtual animal bodies in virtual environments
\cite{Zador2022-hd}.

Several key enabling technologies are now available to construct virtual
bodies and lift them into real-time simulations. Biophysically detailed
models--bones, joints, muscles, neuromuscular junctions, and
proprioception--have been developed through cadaver studies, light-sheet
microscopy, and micro-CT scans. Tools like Blender can be used to edit and simplify
raw anatomical data. Models of varying detail and coverage--ranging
from a single limb to a whole body--exist for multiple species,
including in \emph{C. elegans} \cite{Zhao2024-kb}, drosophila
\cite{Lobato-Rios2022-au,Vaxenburg2024-cv}, larval
zebrafish \cite{Guo2017-mx}, mice
\cite{Ramalingasetty2021-lv,Gilmer2024-dj,DeWolf2024-vm},
rats \cite{Merel2019-kb,Aldarondo2024-is}, macaques
\cite{Almani2024-yd} and humans
\cite{Caggiano2022-mi}. A popular route is to build a 3D
model that can be simplified and calibrated for use in the MuJoCo
physics simulator \cite{Todorov2012-ma,Caggiano2022-mi}.
There is an ongoing trend to move toward more precise models than those
derived from cadaver studies by using detailed scans of bodies
\cite{Vaxenburg2024-cv,DeWolf2024-vm}. 

The exact level of biophysical detail needed to capture the relevant properties depends on the end-goal of the simulation. For example, modeling forearm movement can be done fruitfully with a highly abstracted and fully differentiable effector model of the arm \cite{Codol2024-dw}, or with a more biophysically accurate musculoskeletal model \cite{Almani2024-yd}. On the other hand, a model suitable for grasp may need to model the soft body physics of the hand \cite{Barbagli2004-li}. The endpoint 
of these efforts may be zoos of detailed off-the-shelf biomechanical models at different levels of granularity for use in
physics engines that serve as distillations of ongoing data collection efforts.

Virtual bodies with appropriate, hand-designed controllers can be run in
a physics simulation engine \cite{Bhattasali2022-su}.
Alternatively, controllers may be trained through reinforcement learning
\cite{Aldarondo2024-is}, for example training a virtual
animal to successfully remember and navigate towards reward
\cite{Merel2019-kb} or navigating through a valley
\cite{Vaxenburg2024-cv}. To lift real animals into a
virtual environment, the poses of real animals can be inferred using
single- or multiple-view pose estimation software such as DeepLabCut
\cite{Mathis2018-ii}, Anipose
\cite{Karashchuk2021-lc}, SLEAP
\cite{Pereira2022-yk}, or DANNCE \cite{Dunn2021-kc}. The poses and movements of virtual animals can then
be optimized to match those of real animals, thus effectively
virtualizing the animal. Beyond the animal, the environments in which
they operate can be virtualized using a combination of 3d modeling and
rapidly improving computer vision methods such as photogrammetry, neural
radiance fields \cite{Mildenhall2020-ta} or Gaussian
splatting \cite{Kerbl2023-li}.

Senses can be virtualized at different levels of granularity. For
example, to virtualize a drosophila's sense of sight, one may obtain a
coarse approximation by placing a virtual camera lens where the fly's
eyes lie. With sufficiently accurate tracking, it's possible to
approximate, in a virtual environment, what a fly must have seen
\cite{Cowley2024-dg}. A more detailed simulation could
simulate the hexagonal layout of the ommatidia
\cite{Lappalainen2023-ot} or the specific properties of
the lens; a fine-grained simulation could leverage a detailed digital twin of the eye
\cite{Maheswaranathan2023-wx, Seung2024-bt}. Similar approaches could be taken for other sensory modalities, such as audition. Other important
senses remain challenging to digitize. For example, whisking (touch) is
a dominant sense in rodents that currently requires either high-framerate videography or optoelectronic tracking
\cite{Bermejo1998-va} to accurately track the flexible
whiskers, let alone the forces at the base of the whisker; this is
challenging in free behavior \cite{Knutsen2005-yn}.
Olfaction is another important sense that is technically challenging to
virtualize \cite{Wilson2009-jl}. Proprioception is important to virtualize, as stable and robust movement require sensory feedback. Proprioception can be simulated elegantly with biomechanical models that have mapped the locations and encoding properties of sensory receptors on the body \cite{Mamiya2018-zx, Mamiya2023-kz}.  

\hypertarget{subsubsec-bridging-the-causal-and-real-to-sim-gap}{%
\subsubsection{Bridging the causal and real-to-sim
gap}\label{subsubsec-bridging-the-causal-and-real-to-sim-gap}}

Even with observational data from brains and behavior, as well as
virtual bodies, measurements can only offer partial observations of a system. Not all
neurons can be recorded simultaneously, and those that are recorded may
not capture all the relevant neural dynamics. Simulations of completely transparent neural networks that implement the functions of interest, e.g., sensorimotor control of complex bodies, can help guide us on how to best sample from real brains \cite{Martinelli2024-wi} and capture the posture and movement of bodies.  

A related issue is that datasets rarely cover all possible situations or behaviors an organism can exhibit. Just as we have seen with
self-driving cars, rare but critical scenarios might be
underrepresented, leading models to fail in unanticipated ways
\cite{Liu2024-km}. Since graceful
out-of-distribution failure is one of the desiderata of safe AI systems,
the generalizability and robustness of embodied digital twins are critical. 
Several
groups have demonstrated empirical results that show that more
constraints--e.g. using higher entropy data or constraining the data to
be consistent with a connectome \cite{Beiran2024-cl}--lead to more robust generalization. A mathematically precise theory of the stability of simulations, however, is currently lacking.

One avenue toward better generalization is to build causal models of
neural activity \cite{Pearl2009-bi,Bailey2024-em}. Researchers have begun to explore this approach with \emph{C. elegans} \cite{Haspel2023-zf}, stimulating all neurons and measuring the resulting outputs, in an effort to characterize neural dynamics and behavior. Unlike models that build a regressor from observational data--which can confound
cause and effect--these approaches predict how neural activity and behavior
is causally related to changes in the organism's senses or neural activity.
While this is a promising direction, applying causal analysis to larger
organisms is highly non-trivial.

Even in the best of circumstances, we will likely need to deal with a
sim-to-real gap \cite{Zhao2020-oy}: the virtual
environment can never fully replicate the real world, leading to
discrepancies in simulated sensory inputs and motor outputs. To mitigate
this, models can be adapted or fine-tuned within the virtual
environment. This may involve training the virtual animal through
supervised learning, imitation learning, or reinforcement learning to
perform tasks that are underrepresented in the observational data. This
is conceptually similar to the route by which LLMs are built for chat:
pretraining, supervised fine-tuning, and reinforcement learning from
human feedback \cite{Howard2018-pd}. Designing
appropriate training curricula will be essential to expose the model to
a wide range of scenarios, improving its robustness and
generalizability.

\hypertarget{subsubsec-defining-success-criteria-for-embodied}{%
\subsubsection{Defining success criteria for embodied digital twins}\label{subsubsec-defining-success-criteria-for-embodied}}

How would we know if we achieved success in building an embodied digital twin? One possibility,
taking a page from Bostrom \& Sandberg, is to use autoregressive
prediction as a target. For instance, we could measure a real animal's
behavior and neural activity, virtualize the animal, and predict, in the
virtual environment, its future activity conditioned on the past
\cite{Aldarondo2024-is}. Appropriate metrics could
include, e.g. root-mean-square error (RMSE) in predicting the position and
angles of the limbs, or the neural activity.

While this straightforward approach seems intuitively appealing, it has 
some significant drawbacks. There are many reasons why a simulation could 
fail on a metric basis, not all of which are due to model failure. This includes the difficulty in digitizing the animal and its environment with perfect accuracy, as well as the fact that the systems to be simulated sit at the edge of chaos \cite{O-Byrne2022-fn}. These could lead to the divergence of even very capable models. Like the weather, neural activity and behavior might not be predictable in the RMSE sense over a long time horizon \cite{Rust2023-bq}.

However, much like we can both predict the weather at short time scales and the climate at long time scales, we may be able to simulate neural activity autoregressively at short timescales and the “neural climate” at longer time scales. 
 Appropriate metrics for long-term predictions differ from those for
short-time scales, taking into account the entire distribution of
predictions and observations. KL divergence between smoothed state
spaces \cite{Bischoff2024-gw}, neural geometry
\cite{Williams2021-uu}, and dynamic similarity
\cite{Ostrow2024-dt} have been proposed. Other
relevant metrics may be constructed by taking a page from the evaluation
of generative models. For example, the Frechet Inception distance
measures both the quality and the diversity of generated images
\cite{Heusel2017-ds}.

An alternative approach to long-term behavior evaluation is the embodied Turing test \cite{Zador2022-hd}. The assay, judged by a human discriminator, is a test of the ``behavioral climate'' of a virtual animal. It does not ask whether, for example, a virtual beaver
simulation is doing exactly what the real beaver will do given what it
did a day ago; it asks whether the beaver's behavior is within
distribution and representative of what real beavers do. Automating the
test or finding alternative objective operationalizations will be a key
enabling factor in evaluating simulations of animals at the behavioral
and neural levels. Synthetic judges built from human or conspecific
judgments could be used to scale up evaluation
\cite{Zheng2023-lk}, although precautions are needed to
prevent Goodhart's law. An important implicit guardrail of the embodied
Turing test is its focus on out-of-distribution evaluation: measuring
the behavior of virtual animals in a range of environments, not just the
ones where they were originally trained. 

Finally, a well-designed evaluation of an embodied digital twin should test whether it displays intelligent behavior. That means verifying that the virtual simulation learns or fails to learn; how it generalizes, transfers or suffers from interference across tasks; how its behavior is affected by training curricula; and so on. One would want to design a series of tests of adaptive behavior, a virtual gym, to put the virtual animals and humans through their paces.

\hypertarget{subsec-evaluation-embodied}{%
\subsection{Evaluation}\label{subsec-evaluation-embodied}}

Embodied digital twins intersect with multiple ongoing trends in neuroscience and AI:

\begin{itemize}
\item
  Foundation models: training large-scale models for autoregressive
  generation
\item
  Foundation models for neuroscience: training foundation models
  specifically on neuroscientific data, to create predictive models
  useful for health applications or basic science
\item
  Large-scale naturalistic neuroscience: measuring the activity of a
  significant proportion of the neurons in a single animal during
  natural, possibly social behavior
\item
  Embodied neuroAI: focusing on the interaction between the environment
  and the body to understand the mind
\end{itemize}

Thus, we expect to see considerable progress in embodied digital twins over the next few years as data collection and the application of large-scale modeling in neuroscience continue to scale.

The usefulness of embodied digital twins for AI safety hinges on assumptions that are 
difficult to evaluate. These models are built with many of the
same tools and under similar paradigms to conventional autoregressive
next-token prediction models, e.g. large language models. The safety of embodied digital  agents compared to conventional AI hinges on the
importance and relevance of differentiators--in particular, how much the approach steers the space of solutions to the problem of intelligent behavior toward the set of safe solutions. Some of these differentiators could include that:

\begin{enumerate}
\def\labelenumi{\arabic{enumi}.}
\item
  Brain data poses a stronger set of constraints than behavior only (see Section \ref{sec-use-brain-data-to-finetune-ai-systems} for an alternative take on this idea).
\item
  Embodiment and agency pose a stronger set of constraints than lack of embodiment \cite{Moravec1988-zy}.
\item
  Optionally, connectomes and the genomic bottleneck \cite{Zador2019-gf} pose a stronger set of constraints and inductive biases than current AI architectures.
\item
  Optionally, causal models could have a larger range of stability and out-of-distribution validity compared to conventional models.
\end{enumerate}

\begin{figure}[htbp]
  \centering
  \includegraphics[width=.75\textwidth]{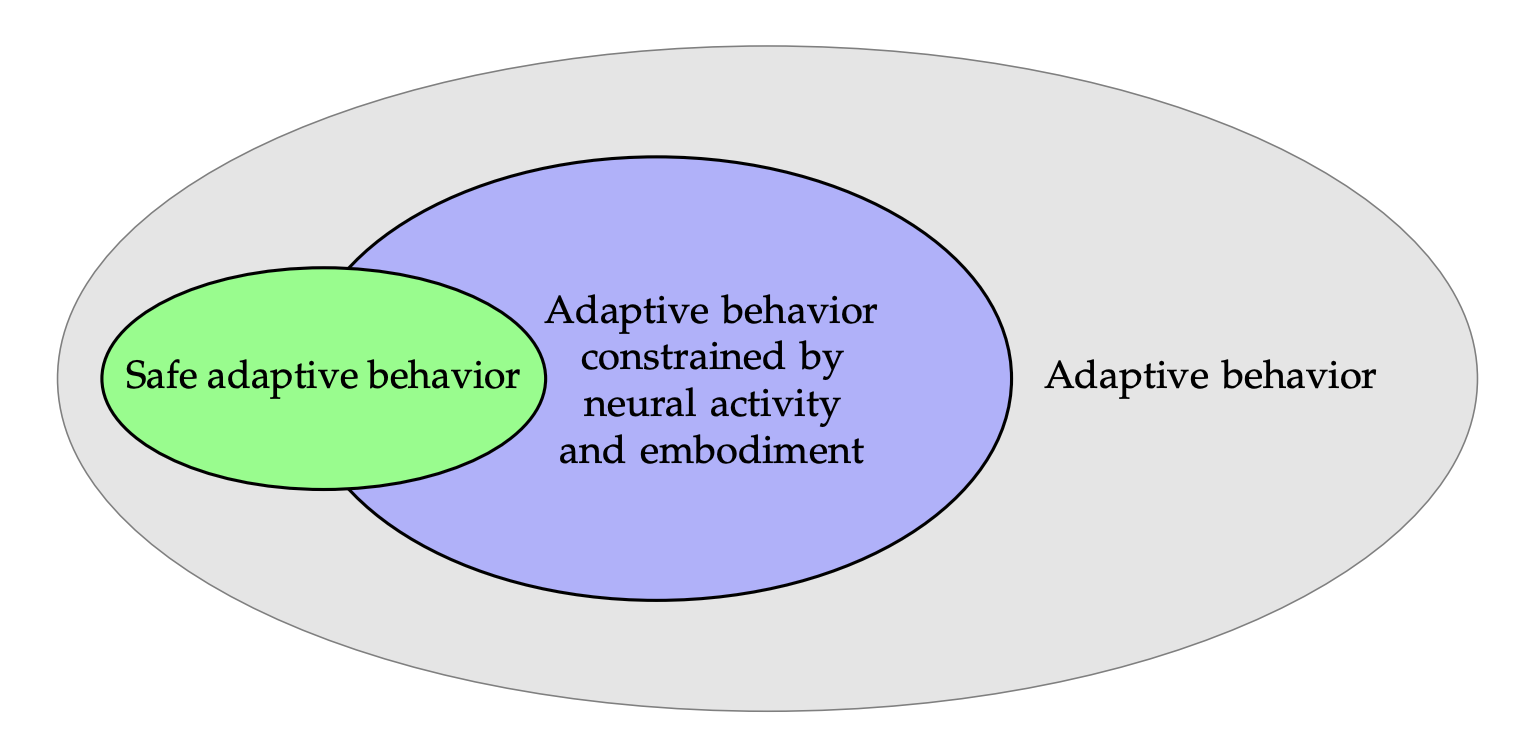}
  \caption{Proposed value of embodied digital twins compared to conventional AI. Out of all potential systems displaying adaptive, intelligent behaviors, only a small proportion display human-compatible, safe behavior. Constraints from neural activity and embodiment may shrink the solution space toward safe adaptive behavior.}
  \label{fig-nested-constrains}  
\end{figure}

These ideas are illustrated diagrammatically in Figure \ref{fig-nested-constrains}. Out of all potential systems that have adaptive, intelligent behaviors, only a small proportion display human-compatible, safe behavior. Constraints
from neural activity, embodiment, connectomics and causality may shrink the solution space toward safe behavior. Whether this premise is correct will need careful examination.

Building on the concept of constraints shaping safe behavior, Sarma et al. (2019) \cite{Sarma2019-ui}
propose a research program focused on developing biophysically detailed simulations 
of simple organisms, such as C. elegans, Drosophila, and zebrafish. These 
simulations would incorporate biomechanics within a simulated environment, providing
 a bottom-up approach for studying the emergence of intelligence and values in 
 embodied agents. This aligns with the idea that embodiment and agency impose 
 stronger constraints on behavior compared to disembodied systems. By 
 studying the interplay of biological constraints, environmental interactions, 
 and emergent behaviors in these simulations, researchers can gain insights into 
 the factors influencing safe and human-compatible behavior in AI systems.

At a meta-level, embodied digital twins and virtual animals could prove useful in
accelerating neuroscience, by making it easier to perform virtual
experiments that can generate hypotheses. Thus, even with a narrow path
toward direct impact on AI safety, accelerating the creation of
large-scale datasets under rich, high-entropy, naturalistic scenarios
and of virtual animals could have a large, positive effect on
fundamental neuroscience research. Beyond the first-order effect of
advancing neuroscience and applications in health, embodied digital twins could
accelerate the virtuous cycle of influence between neuroscience and AI
at the heart of NeuroAI, with positive downstream effects for AI safety.

\hypertarget{subsec-opportunities-embodied}{%
\subsection{Opportunities}\label{subsec-opportunities-embodied}}

\begin{itemize}
\item
  Build new tools to facilitate recording ethologically relevant, high-entropy, naturalistic neural activity, and behavior in free-moving conditions, including multi-animal interactions. 

  \begin{itemize}
  \item
    Wireless neural recording devices capable of recording large numbers of neurons simultaneously in freely behaving animals.
  \item
    Off-the-shelf multi-camera videography and tracking tools with high accuracy and low latency to capture movements and postures in 3D at high resolution, including in groups of interacting animals. 
  \item
    Fully controlled experimental environments that can be recreated in simulation, facilitating comparisons between free behavior in real and virtual animals.
  \end{itemize}
\item
  Generate, aggregate, and disseminate large-scale, high-entropy datasets of naturalistic behavior.
  \begin{itemize}
  \item
    Build and disseminate multimodal foundation models for neuroscience.
  \end{itemize}
\item
  Build a zoo of standardized, adaptable virtual animals complete with bodies actuated with muscles and embedded sensors that may interface explicitly with the nervous system in simulations.
\item
  Leverage comparative neuroanatomy and evolutionary psychology to identify the simplest organisms (mammalian or otherwise) for which embodied digital twins provide the greatest insight into building human-aligned AI. 
\item
  Build an ecosystem of competitions and challenges for embodied animal simulations, including multiple environments, tasks, and interactions, evaluating both short-range and long-range predictions.
\end{itemize}

\hypertarget{sec-wbs}{%
\section{Build biophysically detailed models}\label{sec-wbs}}

\hypertarget{sec-core-idea-wbs}{%
\subsection{Core idea}\label{sec-core-idea-wbs}}

Connectomics generates the structural circuit map of an organism’s nervous system, detailing neurons and their synaptic connections. Biophysical modeling aims to model the electrical and biochemical dynamics of individual neurons. Biophysically detailed models combine these two approaches to create a high-resolution, neuron-level simulation that accurately replicates local neural interactions and large-scale brain activity. It has long been argued that human biophysically detailed models would be a safe path toward general-purpose artificial general intelligence \cite{Bostrom2008-og}, as it replicates the human brain's structure and function in a digital medium. Since it is modeled after the human mind, this brain may be more likely to possess human-like values, emotions, and decision-making processes. This inherent alignment could reduce the risk of developing AI systems whose goals diverge from human interests. However, while connectomics may represent the most direct bottom-up path to building intelligence, it may not be the most efficient approach--just as achieving powered flight did not require replicating birds' feathers.

\hypertarget{subsec-why-does-it-matter-for-ai-safety-and-why-is-neuroscience-relevant-wbs}{%
\subsection{Why does it matter for AI safety and why is neuroscience
relevant?}\label{subsec-why-does-it-matter-for-ai-safety-and-why-is-neuroscience-relevant-wbs}}

Biophysically detailed models are premised on the idea that the brain's mechanisms and functionality can be faithfully reproduced \emph{in silico} without the need for actual biological matter: the specific wiring of neurons and signaling mechanisms crucial for emergent phenomena such as intelligence, values, and perception can be recapitulated if reconstructed from bottom-up. Recurrent connections, diverse cell types, and precise synaptic properties are critical in neural function and information processing \cite{Einevoll2019-ca, Haufler2023-hb}. However, this raises fundamental questions about the appropriate level of abstraction--while a perfect copy would require modeling down to atoms and molecules, higher-level approximations capturing informational or biophysical dynamics could preserve core functionality. This abstraction choice has profound practical implications: simulating detailed biochemical processes that occur naturally and efficiently in biological neural tissue may require enormous computational resources when implemented on traditional computing hardware, potentially demanding infrastructure-scale energy requirements that could make whole-brain modeling impractical at increasing levels of biological detail. Biophysically detailed models take an implementation-maximalist view of AI safety: if we can simply model a whole human brain at the biophysical level, we could have a more human-like AGI. Alternatively, if these models were to fail, it may fail in more human-like ways. Steps along this path include emulating whole mouse brains and certain subsystems or processes. We could apply the same intuitions when reasoning about biophysically detailed models as we do when reasoning about humans; we could apply existing moral frameworks to modeled minds; and we could collaborate naturally with these models. 

The difficulties associated with these models are both technical and conceptual. Key technical barriers include long timelines to map an entire human connectome, difficulty in scaling tools to probe single neuron function, and the amount of compute required to simulate billions of interconnected neurons. Conceptually, questions remain about what neural properties must be directly measured versus inferred, what can be learned from measurements, and what additional constraints are needed for accurate simulation. These challenges have generally made this approach less appealing in the short term (<7 years). 16 years after the original biophysically detailed model roadmap \cite{Bostrom2008-og, Wellcome-Trust2023-tt}, and in light of very significant advances in using the fly connectome to understand how the fly’s brain controls its behavior \cite{Cowley2024-dg, Pospisil2024-nq} and new optical microscopy approaches that could solve some of these challenges, it may be time to revisit this approach.

\hypertarget{subsec-details-wbs}{%
\subsection{Details}\label{subsec-details-wbs}}

Cellular-resolution biophysically detailed models rest on three major foundations: extracting cellular-level structure at the whole brain scale, capturing representation of electrophysiological and biochemical processes in neurons and synaptic connections, and simulating these at scale. Achieving the latter appears to be within reach, as simulations of networks of neurons at the scale of the human brain (\textasciitilde{}86 billion neurons, \textasciitilde{}100 trillion synapses) have been recently demonstrated \cite{Lu2023-vk, Yamaura2020-ab}, at least with point-neuron approaches. Therefore, data availability--the circuit architecture in terms of cells and their connections and the associated physiology-- is the main bottleneck for these models. 

Below we describe two approaches to structural and molecular mapping, electron microscopy (EM) and optical microscopy. Whereas the circuit architecture of the brain will need to be characterized via connectomic approaches, measuring physiology directly from every single cell and synapse in the same brain appears impractical to impossible \cite{Marblestone2013-du}. Instead, one will likely need to combine more targeted physiological and multi-omic measurements of cellular and synaptic properties from diverse cell types and in many different brain areas with ML techniques, which will assign physiological properties to cells and synapses from the connectomic imaging data \cite{Januszewski2024-pu}. Together, these approaches have the potential to create highly detailed and accurate biophysical models.

\hypertarget{subsubsec-generating-whole-brain-connectome-datasets-with-electron-microscopy}{%
\subsubsection{Generating whole-brain connectome datasets with electron microscopy}\label{subsubsec-generating-whole-brain-connectome-datasets-with-electron-microscopy}}

\begin{figure}[htbp]
    \centering\includegraphics[width=6.5in]{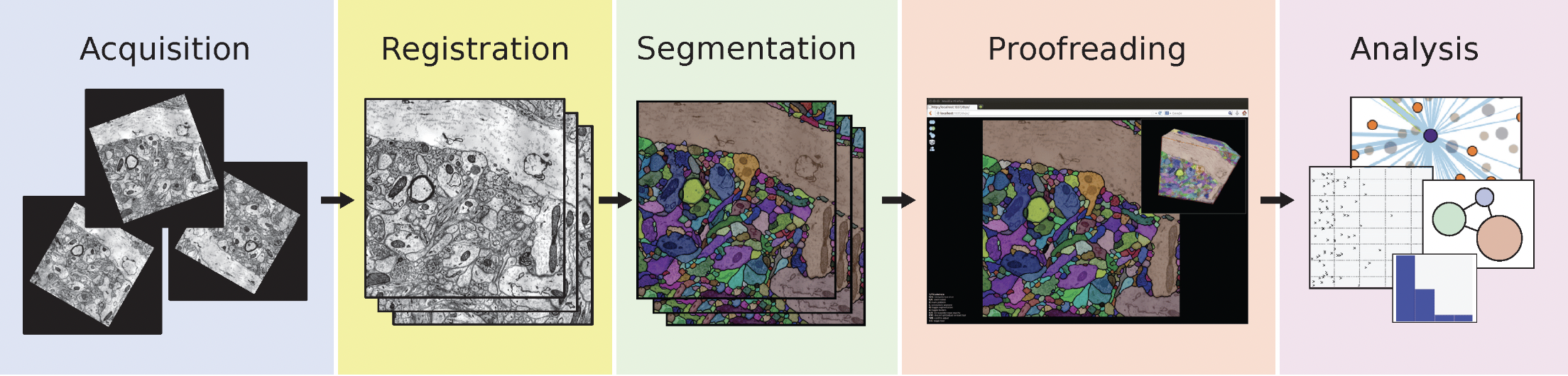}
    \caption{General pipeline for whole-brain connectomics. From \cite{Haehn2017-xz} under a CC-BY license.}
    \label{fig-connectomics-pipeline}
\end{figure}

The C. elegans connectome was first published in 1986 \cite{White1986-se} after a decade of painstaking work. Advances in microscopy and artificial intelligence have made it significantly easier to generate large-scale neural connectivity datasets. Recently published connectomes include a 1 mm$^3$ volume of mouse cortex \cite{Consortium2023-og}, and a complete connectome of the larval \cite{Winding2023-re} and adult \cite{Dorkenwald2023-xt} fruit fly brains. 
Progress has been made in mapping the human brain as well. Shapson-Coe et al. \cite{Shapson-Coe2024-by} published a 1 mm$^3$ volume of the human cortex, consisting of a thin slab 170 µm thick. While this geometry means many of the neural connections are severed at the boundaries of the volume, it remains an excellent resource for studying cell type distribution and morphology. This is a critical input in building biorealistic models that respect the diversity of cell types in the human cortex.

 The general pipeline for whole-brain connectomics consists of five major stages (Figure \ref{fig-connectomics-pipeline}). First is data acquisition, where high-resolution electron microscopy images are collected of brain tissue sections. These images then undergo registration to align consecutive sections into a coherent 3D volume. The third stage involves segmentation, where computational methods are used to identify and label distinct cellular structures, typically using machine learning approaches. This is followed by proofreading, where human experts verify and correct automated segmentation results. Finally, the validated reconstruction enables various analyses of neural connectivity and network properties, ultimately to be used as an input for simulation.

Connectomics requires imaging methods that can resolve individual synapses. To accurately capture synaptic structures, we must resolve features that are typically around 20-30 nm in size, for example, the distance between pre- and postsynaptic membranes. This is beyond the diffraction limit resolvable by visible and near ultraviolet light in conventional microscopy. Instead, scans are typically captured by volume electron microscopy (vEM), in particular serial section transmission electron microscopy (ssTEM) and scanning electron microscopy (SEM) of brain slices typically less than 100 nm thick. Imaging must be fast to feasibly scale to the volume of the human brain, which is \textasciitilde{}1300 cubic centimeters, or 1.3M cubic millimeters. The fastest reported net imaging rate with vEM is 0.3 gigapixels per second, which could theoretically image a cubic millimeter in 5 weeks at a 3 nm planar resolution with an array of four microscopes \cite{Zheng2024-cb}. Without further enhancements in throughput, we would need \textasciitilde{}50,000 parallel microscopes to scale up to the human brain within a decade.

Importantly, post-processing (registration, segmentation, and proofreading) has been historically the most time-consuming and expensive part of the process. Although machine-learning enabled methods such as flood-filling networks \cite{Januszewski2018-ic} and local shape descriptors \cite{Sheridan2023-us} for automated segmentation have improved the efficiency of synaptic tracing in terms of GPU time, post-processing still requires human proofreading, at least for now. The adult fly brain connectome has benefited from tens of thousands of volunteers manually checking neuron reconstructions across slices, collectively adding up to a 33 person-year effort \cite{Dorkenwald2023-xt}. A recent roadmap extrapolated that whole-mouse brain connectome proofreading would require roughly 0.3-1M person-years, consuming over 97\% of the entire project’s budget \cite{Wellcome-Trust2023-tt}. Straightforward extrapolation by brain volume from Figure 5 of the Wellcome Trust roadmap \cite{Wellcome-Trust2023-tt} results in an estimate \textasciitilde{}\$1.5-4.3T for a whole macaque brain and \textasciitilde{}\$20T-59T for a whole human brain with EM connectomics, or about two years of the US GDP. Clearly, massive cost reductions will be necessary to build a whole-brain human connectome within reach.

\begin{figure}[htbp]
    \centering
    \includegraphics[width=6.5in]{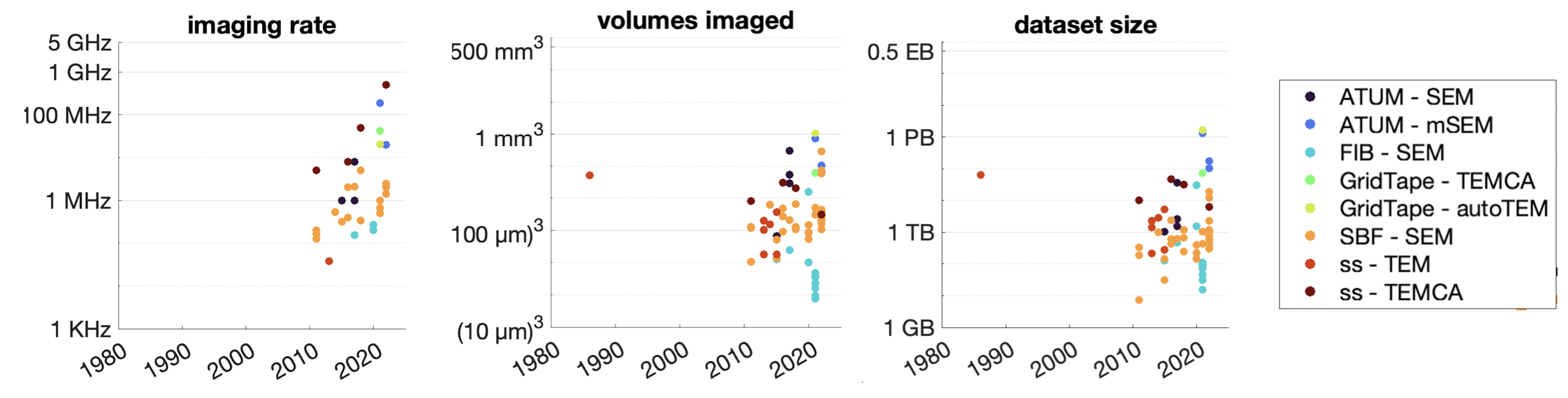}
    \caption{Compiled connectomics datasets and their imaging rate, volumes imaged, and dataset size. From \cite{Wellcome-Trust2023-tt} under a CC-BY license.}
    \label{fig-wbs-datasets}
\end{figure}

\hypertarget{subsubsec-emerging-scalable-approaches-for-generating-whole-brain-connectome-datasets}{%
\subsubsection{Emerging scalable approaches for generating whole-brain connectome datasets}\label{subsubsec-emerging-scalable-approaches-for-generating-whole-brain-connectome-datasets}}

Newer imaging modalities may overcome some of these limitations. One promising opportunity is to adapt genetic cell barcoding techniques in which millions of individual cells can be optically distinguished from each other using diverse molecular labels (barcodes) \cite{Kebschull2018-xf}. These barcodes can be expressed as unique nucleic acid sequences \cite{Chen2019-iz}, or alternatively diverse combinations of protein variants \cite{Livet2007-ok}. In both cases, barcodes can be detected as color codes via sequential optical imaging, enabling neurons to “self-annotate” their identities in optical microscopy images, in contrast to single-channel grayscale stains used in electron microscopy. If labeling methods can be engineered such that barcode molecules can fill the entire cell (including difficult-to-label compartments such as thin axons and distant synaptic terminals \cite{Viswanathan2015-dw}), or alternatively if barcode molecules can be universally targeted to synapses \cite{Mishchenko2010-or}, barcode information could potentially be used for error correction in connectomic reconstructions, significantly reducing costs \cite{Marblestone2014-kg}. In combination with on-going advances in improved segmentation \cite{Qihua2024-rm} and enhanced human proofreading tools \cite{Dorkenwald2023-xt}, cell barcoding could reduce or even eliminate human proofreading as a cost bottleneck, addressing what is currently the most critical cost driver in connectomics. 

Additional technical challenges must be overcome before cell-filling barcodes can be used for connectomics. Because the diffraction limit of light microscopy is too large to resolve critical nanoscale features like thin axons and dense synaptic connectivity, advanced sample preparation methods are needed to optically detect cell barcodes at high spatial resolution. By embedding whole brains in hydrogels and expanding the tissue, it is possible to overcome the diffraction limit of light microscopy and increase the effective imaging resolution \cite{Chen2015-wn}. Tissue expansion methods can now detect nanoscale morphological information \cite{Tavakoli2024-ix} suitable for dense reconstructions, and can also detect diverse molecular labels multiplexed across multiple imaging cycles \cite{Alon2021-ns, Kang2024-ej}. A combination of these methods would yield image datasets suitable for dense connectomic reconstruction, with additional diverse information channels encoding cell identities available for error correction.

In addition to possibly removing proofreading as the biggest cost bottleneck, optical microscopy has the advantage of the microscopes themselves being significantly lower cost than electron microscopes \cite{Adamhmarblestone2019-fn}. According to a recent cost estimate, if protein cell barcoding can eliminate proofreading as the cost bottleneck, a mouse connectome might be acquired for a marginal cost of only \$7M assuming an initial capital outlay of \$10M, achieving approximately a 1000x cost improvement compared to the state of the art \cite{Unknown2023-zf}. Further improvements may be possible: this estimate assumes each pixel is imaged \textasciitilde{}100 times, or \textasciitilde{}50 imaging cycles to read out a 100-bit barcode address space. Custom microscope designs implementing additional spectrally distinct laser lines and cameras could reduce the number of imaging cycles required. Additionally, spatially multiplexing barcode information throughout a cell via reporter islands \cite{Linghu2020-it} would permit exponential rather than linear readout. Together, a 125-bit address space could be read out in only 3 imaging cycles using 5 spectrally distinct laser lines and cameras, reducing the marginal cost of connectome image acquisition in this estimate by another \textasciitilde{}20-fold. If improvements in image segmentation efficiency \cite{Sheridan2023-us} can keep pace to manage the cost of volume reconstruction, and storage costs can be managed by on-line data processing, the marginal cost of an optical connectome could fall below \$1M per cubic centimeter.

Can these approaches scale to human connectomes? It remains to be seen whether optical connectomics augmented with cell barcodes can eliminate proofreading as the bottleneck and achieve low costs. In addition, other emerging connectomics technologies, including x-ray microscopy and non-microscopic sequencing based approaches \cite{Zador2012-vh, shin2024high}, may yield even better performance characteristics and lower costs. Thus, a low-cost, high-quality human scale connectome could be within reach on a much shorter timeline (<10 years) at a much lower cost (<\$1B) than previously predicted \cite{Wellcome-Trust2023-tt}. 

It is important to note that most connectomic data currently provides information about cellular-level connections but not necessarily weights or other properties of these connections, which, at least in some cases, appear to be more important for brain dynamics and computations than connections themselves \cite{Arkhipov2018-cr}. Although EM data (e.g., information about synapse size) can be used to predict synaptic weights \cite{Holler2021-dm}, and synaptic properties like kinetics and short-term plasticity of postsynaptic potentials are somewhat constrained by pre- and post-synaptic cell types \cite{Campagnola2022-xu}, these relations exhibit substantial variability and have not yet been established for a vast majority of brain cell types (although optical microscopy approaches are beginning to reveal highly stereotyped patterns of synaptic connections \cite{Sanfilippo2024-aj} with epitope tagging). In addition, capturing rules of long-term plasticity and learning, as well as the effects of neuromodulators, is likely to be crucial for accurate whole-brain models. Lastly, non-neuronal cells like glia are typically not included in the definition of a connectome, but may be necessary for some cognitive functions \cite{Lee2014-xf}. Substantial challenges remain in understanding these phenomena and their diversity across cell types and brain areas. There is potential synergy with emerging optical approaches discussed above, as molecular information can be detected alongside barcode information in the same assay \cite{Shen2020-fc}.

\hypertarget{subsubsec-electrical-and-biochemical-properties-of-single-neurons}{%
\subsubsection{Measuring the electrical and biochemical properties of single neurons}\label{subsubsec-electrical-and-biochemical-properties-of-single-neurons}}

Armed with a connectome, we still face the problem of reconstructing the state-dependent input-output function of each neuron. This requires additional information to be collected alongside or predicted from imaging data to provide parameters for the electrical and biochemical behavior of single neurons. 

Using ex vivo tissue, methods like Patch-seq, which combine single-cell electrophysiology, morphology, and transcriptomics, link the molecularly defined cell types to their morpho-electric properties \cite{cadwell2016electrophysiological, fuzik2016integration, scala2021phenotypic, Lee2021-wn, Gouwens2020-qz, Berg2021-gl}. Additionally, live recording techniques such as calcium imaging and large-scale electrophysiology capture dynamic, state-dependent neural activity that provides crucial information about how neurons behave in vivo \cite{Atanas2023-ty, De_Vries2020-og, Manley2024-xn, Siegle2021-ix, Leonard2024-zp, Khanna2024-be}. While it is not directly possible to establish causality from observational data, \cite{Haspel2023-zf} hypothesize that large-scale causal perturbation experiments could be used to reverse engineer the \textit{C. elegans} nervous system. Connectomes are highly informative here, as they provide a crucial constraint on the nodes that must be simultaneously stimulated and observed \cite{Pospisil2024-nq, Creamer2024-cp, Beiran2024-cl}. Functional data can be collected before generating the connectome from the same animal \cite{Consortium2023-og}. At least in animals with a high amount of stereotypy, average connectomes may sufficiently constrain the problem \cite{Lappalainen2023-ot, Cowley2024-dg, Shen2024-mk}.

Optical approaches enable additional imaging-based data collection, including in-situ transcriptomic characterization \cite{Alon2021-ns}, connecting such spatial transcriptomics with functional in vivo activity of individual neurons \cite{Bugeon2022-ri, Condylis2022-jr}, pan-protein labeling \cite{MSaad2020-gf}, or specific protein labeling to infer cell types. Recent advancements have also demonstrated the potential to infer cell type directly from electron microscopy data \cite{Zinchenko2023-qt, Gamlin2023-bd,Holler2021-dm}, leveraging the paired morphology and transcriptomic data obtained from Patch-seq experiments.

It is worth noting that these methods do not account for extra-synaptic signaling, such as dense-core-vesicle-dependent signaling that results in different signal propagation than predicted based on the synaptic connectivity \cite{Randi2023-mj}. However, molecularly annotated optical connectomes could help account for this by subcellular receptor localization. While extra-synaptic signaling has been a bottleneck in C. elegans, it is unknown to what degree extra-synaptic signaling will be problematic in larger animals. 

\hypertarget{subsubsec-biophysical-modeling-of-single-neurons-and-populations-of-neurons}{%
\subsubsection{Biophysical modeling of single neurons and populations of neurons}\label{subsubsec-biophysical-modeling-of-single-neurons-and-populations-of-neurons}}

With connectome data and information about neuronal input-output functions, we can begin to construct biophysically detailed models. These models range from simplified network representations to highly detailed biophysical simulations. The great computational unknown of this approach is what kind of model (and thus data required) is needed to accurately represent various brain functions. Can we use reduced point neuron models like leaky-integrate-and-fire, or do we require biophysically realistic models for every cell type? Is cell type a close enough abstraction or do we also require protein and RNA abundance (or more detailed data like protein localization) in every individual cell? Recent examples include:
\begin{itemize}
\item 
\textit{Data-driven biophysical models (approximated mesoscale connectome).} \cite{Gouwens2018-lv} use experimental data to infer biophysical parameters, generating biophysically detailed models of diverse cortical cell types from patch-clamp recordings and morphological reconstructions. This includes detailed ion channel dynamics and dendritic computations, but is computationally intensive. \cite{Billeh2020-nw} and \cite{Markram2015-wv} simulated networks of \textasciitilde{}50,000 and \textasciitilde{}30,000 such detailed neuronal models, respectively, to study dynamics and computations in cortical circuits. It takes \textasciitilde{}500 CPU core-hours to simulate one second of biological time with these models (30-50k neurons). It is worth noting that there are theoretically proven methods that could lead to a \textasciitilde{}100x speedup compared to standard serial NEURON on CPUs in simulating biologically realistic models \cite{Zhang2023-lk}.
\item
\textit{Data-driven point neuron models (approximated connectome).} \cite{Shiu2024-sq} perform a whole-fly-brain simulation using leaky integrate-and-fire neurons whose connection patterns mirror the fly connectome. This model can recapitulate feeding and grooming behaviors that are well-studied in flies. The point neuron approach balances biological realism and computational efficiency.
\item
\textit{Data-driven other models (without biophysical dynamics).} \cite{Lappalainen2023-ot} use connectome data to define an artificial neural network (ANN) model of the fly visual system, optimizing parameters to perform specific tasks in the tradition of task-driven neural networks. This allows the recapitulation of single-cell function, but lacks detailed biophysical realism. \cite{Pospisil2024-nq} directly fit input-output relationships of neural circuits from experimental recordings, without explicitly modeling underlying biophysical mechanisms.
\item 
\textit{Hybrid approaches.} \cite{Dura-Bernal2023-fw} integrate predicted connectivity with simplified dynamics for some components and detailed biophysics for others. For example, corticospinal and corticostriatal cell model morphologies had 706 and 325 compartments, but excitatory and inhibitory neurons had 6 and 3 compartments (soma, axon, dendrite). Their approach took \textasciitilde{}96 core hours of high-performance computing time to simulate one second of biological time. \cite{Billeh2020-nw} combined \textasciitilde{}50,000 biophysically detailed neuron models with point-neuron models in a 230,000-neuron network model of mouse V1 and also developed a fully point-neuron version of this network, which produced results that were consistent with the biophysical simulation.
\item 
\textit{LFP models.} \cite{Hagen2016-bq, Dura-Bernal2023-fw, Rimehaug2023-zr} used point-neuron and biophysically detailed network models of cortical circuits to simulate not only neural activity, but also biophysical signals such as the Local Field Potential (LFP), which is commonly used in medicine and bioengineering (e.g., for neuroprosthetics).
\end{itemize}


\hypertarget{subsec-evaluation-wbs}{%
\subsection{Evaluation}\label{subsec-evaluation-wbs}}

Largely the same considerations apply to biophysically detailed models as to embodied digital twins in terms of reproducing in vivo neural physiology recordings, mimicking behavior, and identifying desirable long-term features of whole brain modeling. Biophysically detailed models additionally require that we correctly capture circuit and biophysical mechanisms (at a certain level of resolution), which is a difficult task, but comes with a benefit of helping one constrain the huge space of solutions approximating activity and behavior to those implemented in the brain's "hardware". As such, these models represent a challenging but perhaps most direct path to creating a human-like AI implemented according to its biological substrate.

While connectomics reveals details of synaptic architecture of the brain that are crucial for accurate whole-brain models, many challenges remain with respect to other properties determining the circuit function. These include electrical and biochemical properties of neurons and synapses, effects of neuromodulators, and, perhaps most importantly, plasticity and learning rules. Current technology requires extensive experimental mapping of such properties for each of the thousands of brain cell types if not millions to billions of individual cells. Nevertheless, accelerating technological progress suggests that it might be possible to circumvent these problems by establishing statistical links between electrical and biochemical properties and structural information from connectomic data. If breakthroughs in experimental techniques and machine learning tools facilitate sufficient progress in this area, this will drastically propel this field forward and enable whole-brain models.

\hypertarget{subsec-opportunities-wbs}{%
\subsection{Opportunities}\label{subsec-opportunities-wbs}}

There are critical opportunities to understand the technical and conceptual questions in brain connectomics and biophysics, connecting the two via biophysically detailed models. Additional opportunities can be found in \cite{Stevens2020-bioimaging}.

\begin{itemize}
\item
With the smallest possible organism, collect all possible data (proteomics, transcriptomics, connectomics, in vivo neurophysiology, behavior, other cell types, etc.) and carry out extensive modeling to investigate the precision of simulating circuit function from its structure and understanding what data might be missing (or, conversely, unnecessary) for accurate models. Some proposals to do this include \cite{Simeon2024-ee, Haspel2023-zf}.
\item
Leverage emerging light-microscopy tools to deliver connectomics datasets of whole brains that can be obtained faster and cheaper and are less challenging for computational reconstruction. With these connectomes, rapidly test less computationally intensive point neuron models and scale to biophysically realistic models.
\item
Scale up scanning technology and associated computational/AI tools to the level of whole mammalian brains - first the mouse and later monkey and human.
\item
Develop software foundations that can integrate biophysically detailed models with modern neural networks in a modular fashion.  
\item
Combine broad profiling of biophysical properties of brain cells and connections across cell types and brain areas with AI techniques to generate accurate models for every cell and connection recorded in the structural connectomic datasets.
\item
Investigate neuromodulation, metabolism, and plasticity mechanisms and incorporate them in biophysically detailed models.
\item
Leverage comparative neuroanatomy to identify the simplest organisms that have homologous deep-brain structures to mammals, so that biophysically realistic simulations can be informed by top-down approaches based on evolutionary neuropsychology. 
\item
Improve the performance and power consumption of brain simulations to enable affordable and scalable biophysically detailed models.

\end{itemize}

\hypertarget{sec-develop-better-cognitive-architectures}{
\section{Develop better cognitive architectures}\label{sec-develop-better-cognitive-architectures}
}

\emph{This section contributed by Eli Bingham, Julian Jara-Ettinger, Emily Mackevicius, Marcelo Mattar, Karen Schroeder and Zenna Tavares.}

\hypertarget{subsec-core-idea}{%
\subsection{Core Idea}\label{subsec-core-idea}}
The dominant paradigm in AI development focuses on rapidly testing and iterating over current models, promoting the ones with the best benchmark results. This has resulted in striking advances in AI capabilities at the expense of robust theoretical foundations. The gap between our ability to build powerful AI systems and predict their capabilities and failure modes is only growing.

A promising alternative approach, rooted in cognitive science and psychology research, is to develop cognitive architectures \cite{Lake2017-jf}--formal theories that specify how intelligent systems process information and make decisions. These architectures emerged as scientific efforts to understand how the human mind works, and aim to capture the core mechanisms through which an intelligent agent represents knowledge about its environment, updates it based on observations, reasons about it, and acts on the world. Because computational implementations of cognitive architectures are built using distinct modules with clear interfaces between them, we can inspect and verify each component independently--a critical feature for AI safety.

To date, most cognitive architectures aim to explain only a small set of psychological phenomena in the context of narrow task specifications, and have been painstakingly hand-engineered by human experts over many years. However, advances in cognitive science, neuroscience, and AI suggest a path toward new kinds of cognitive architecture that exhibit human-like behavior and latent cognitive structure in a far broader and more naturalistic set of tasks \cite{wong2023word,sumers2023cognitive}. At a high level, our approach is to build three inter-related \emph{foundation models}--for behavior, cognition, and inference--along with curating the data to train them, and building a programming language around them. In this roadmap section, we argue that building such a system is within reach of today's technology and sketch what a concerted engineering effort would entail.

\hypertarget{subsec-relevance-ai-safety}{%
\subsection{Why does it matter for AI safety and why is neuroscience relevant?}\label{subsec-relevance-ai-safety}}

\begin{figure}[h!] 
    \centering
    \includegraphics[width=0.65\linewidth]{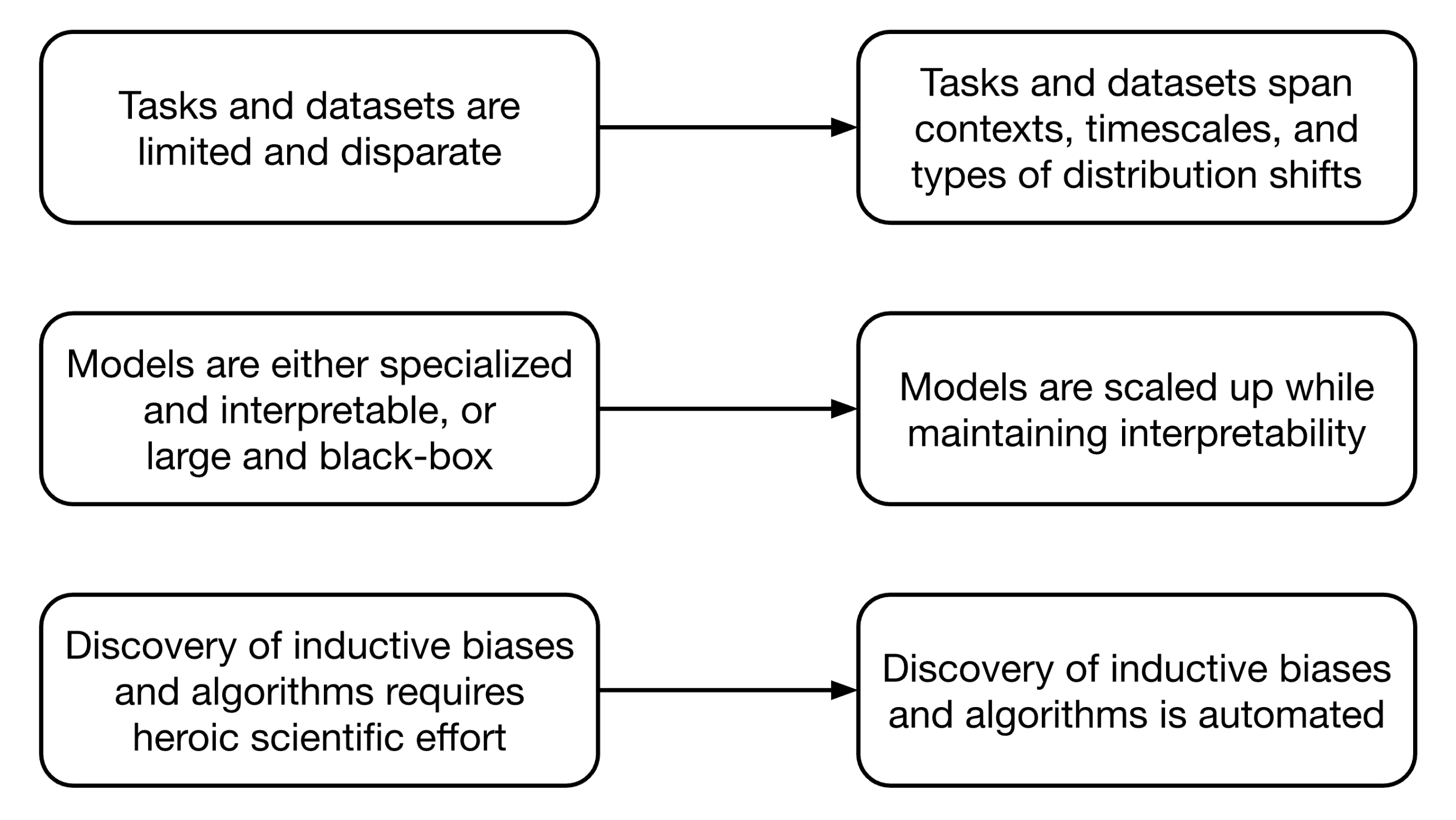}
    \caption{We envision coordinated advances in benchmarks, model architectures, and languages, which will enable systematic reverse-engineering of human cognitive capabilities at scale.}
    \label{fig-cogsci-goals}
\end{figure}

A fundamental challenge in building safe AI systems is defining what we mean by "safe" in formal, executable terms. Without such specifications, we cannot reliably engineer systems to be robust, aligned, or verifiable. 
A cognitive architecture that includes human-like inductive biases--innate pieces of knowledge that facilitate learning the rules of the physical and social world, like those enumerated in the foundational paper \cite{Lake2017-jf}--would go a long way toward addressing this challenge. These inductive biases help ensure safe behavior in several ways: they constrain learning to favor reasonable interpretations of limited data, they guide exploration of new situations, and they provide defaults for handling uncertainty. For example, humans' innate understanding of object permanence and physical constraints helps them explore their environment safely, while basic social biases help them learn appropriate behavior from minimal examples. In addition, possessing a formal description of model inductive biases would allow researchers to predict and mitigate undesired model behavior.

However, coming up with an explicit formal specification of any of those biases outside relatively small proof-of-concept systems has proven frustratingly difficult.
Our proposal aims to get around this difficulty by using experimental tasks and human behavioral data in those tasks to specify inductive biases implicitly, so that machine learning systems trained on the data will acquire the inductive bias in a scalable and predictable way. Once successfully built, a cognitive architecture offers a unified rather than piecemeal approach to specification, robustness and assurance in AI safety. We summarize how each piece of our proposal accelerates progress toward a complete cognitive architecture in Figure \ref{fig-cogsci-goals} and address these aspects of safety in the other sections below.

\hypertarget{subsec-details}{%
\subsection{Details}\label{subsec-details}}

A growing body of experimental and theoretical evidence from across AI, cognitive psychology, and neuroscience is converging on a unifying view of intelligence through the lens of \emph{resource rationality}. 
In this paradigm, intelligent systems are analyzed at Marr’s computational level \cite{Marr1982-xh} in terms of the problems they aim to solve and the optimal solutions to those problems, subject to the algorithmic-level constraint that they make optimal use of their limited computational resources to do so.
More extensive discussions of the history and evidence for this paradigm appear in foundational review articles \cite{binz2024meta, Lake2017-jf, lieder2020resource} and their associated open commentaries. A wider web of connections to other fields is explored in \cite{Gershman2015-we, icard2023resource}.

Here, we focus on the challenge of consolidating this loose consensus into a comprehensive model of the mind as a whole \cite{newell1994unified}, one applicable to both the design of AI systems that are qualitatively safer and more capable than those available today, and to the analysis of real-world intelligent systems in truly naturalistic environments and behaviors (Figure \ref{fig-cogsci-goals}). For each of the three aspects of AI safety considered in this roadmap, we identify key bottlenecks to consolidation for that aspect and new opportunities to overcome them afforded by recent advances in AI.

\hypertarget{subsubsec-benchmark-tasks-data}{%
\subsubsection{Scaling specification: benchmark tasks and experimental datasets}\label{subsubsec-benchmark-tasks-data}}

AI systems today are remarkably capable in many ways and frustratingly incapable in others. They have surpassed human-level performance at chess and Go, but struggle with simple physical reasoning puzzles that children can easily solve. Evidence from cognitive science suggests that the flexibility, robustness and data-efficiency that are still distinctive advantages of human intelligence are derived in part from a small number of inductive biases \cite{Lake2017-jf,spelke2022babies}. Researchers have leveraged these insights to design benchmark tasks--tasks that can be performed by both humans and AI systems--where humans still dramatically outperform mainstream AI systems \cite{arc-prize-2024}. They have also built prototype AI systems for some of these tasks that leverage human-like inductive biases to match human performance \cite{Ellis2020-ie, Lake2015-kb, wong2023word}.

Unfortunately, it has proven challenging to scale up from these isolated proofs of concept to formal and explicit specifications of a full suite of human-like inductive biases that can be incorporated into large-scale generalist AI systems. An alternative approach to identifying inductive biases is to replace the non-scalable step of applying human ingenuity by an inherently scalable process of curating an ensemble of experimental tasks and collecting human behavioral datasets drawn from those tasks. The tasks and data can be seen as an implicit specification of the inductive biases, in the sense that a standard machine learning model could learn the inductive biases simply by fitting the human data.

\hypertarget{par:designing-neuroscience-tasks}{%
\paragraph{Benchmark tasks}\label{par:designing-neuroscience-tasks}}

Benchmarks are carefully designed experimental paradigms that serve as implicit specifications of desired cognitive capabilities. We propose that good benchmark tasks should respect the following desiderata: 
\begin{itemize}
    \item \textbf{A large gap between human and AI task performance.} This gap is currently a good indicator of an area in which inductive biases could be reverse-engineered to great effect. For example, the ARC and ConceptARC benchmarks \cite{Chollet2019-bq, moskvichev2023conceptarc}, which were designed to test abstraction and reasoning abilities, remain unsolved to human-level performance by LLMs, despite the application of biologically implausible compute budgets. Additional examples include few-shot learning tasks like Bongard problems (learning visual concepts from a single set of positive and negative examples) \cite{wust2024bongard}, and physical reasoning problems like PHYRE (intuitive physics understanding) \cite{bakhtin2019phyre}.
    \item \textbf{Evaluate broadly-applicable inductive biases and human-like generalization.} Tasks should not be narrowly scoped, as is sometimes desirable in hypothesis-driven neuroscience experiments, but rather offer the opportunity for generalization by design. One example is Physion \cite{bear2021physion}, a physical reasoning benchmark that tests the ability to predict how physical scenarios will evolve over time. This environment could be used to set up an arbitrarily broad range of scenarios across different types of objects.
    \item \textbf{Performable by humans and AI systems using the same interface.} Clever design of the task interface greatly simplifies comparison of human and AI abilities. For example, ARC abstracts away visual perception by representing task puzzles as grids of integers. These grids can be visualized by human subjects by mapping each integer to a unique color, thereby creating a human-friendly interface without changing any semantics of the task. Video games are similar, in that a rendering of a chess board is a human-friendly interface to a task that can be provided to an AI system as simply a matrix of values.
\end{itemize}

Some categories of behavioral 'benchmarks' that are aligned with our desiderata include:
\begin{itemize}
    \item \textbf{Purpose-built psychology tasks.} Experimental paradigms from cognitive psychology isolate specific mental processes, making them both valuable probes of human intelligence and tractable for AI systems. Their simplicity reveals the inductive biases that give humans their efficiency advantage. Tasks studying causal learning are particularly valuable as they reveal inductive biases that enable humans to learn efficiently from limited data, e.g. a tendency to interpret explanations as communicating causal structure \cite{kirfel2022inference}.
    \item \textbf{Animal behavior.} Animal behavior studies offer valuable insights into core cognitive capabilities without human-specific overlays like language or culture. Rodents demonstrate pure spatial reasoning, birds show probabilistic inference in foraging, and primates reveal exploration-exploitation tradeoffs. Recent advances in computer vision and neural recording have produced high-quality open datasets of animal cognition \cite{Pereira2022-yk, sun2022mabe22, smith2022behavioral, matzner2020thermaltracker, van2015building, koger2023quantifying, naik20233d}. Studying cognition across species helps identify generalizable computational principles underlying behavior while avoiding overfitting to human-specific traits.
    \item \textbf{Video games.} Video games offer intuitive, scalable benchmarks that both humans and AI can interact with identically \cite{allen2024using}. While AI systems like Agent57 now master individual Atari games \cite{Mnih2014-yk, badia2020agent57}, they differ markedly from human strategies. Custom game environments can probe specific cognitive capabilities where human inductive biases confer advantages, such as physical reasoning \cite{bakhtin2019phyre, xue2023phy} and causal understanding \cite{das2023combining}, where humans show distinctive advantages due to their inductive biases.
    \item \textbf{In-the-wild human behavior.} Naturalistic data from strategic gameplay, forecasting, and collaborative platforms (e.g. GitHub, StackExchange) reveal human problem-solving in action \cite{chowdhary2023quantifying, Meta-Fundamental-AI-Research-Diplomacy-Team-FAIR-+2022-oz}. While analyzing this data is complex, it can illuminate cognitive processes, as demonstrated by resource-rational decision-making studies in chess \cite{russek2022time}.
\end{itemize}

\hypertarget{par:collecting-human-datasets}{%
\paragraph{Large-scale human behavioral experiments}\label{par:collecting-human-datasets}}

One important limitation of existing datasets is that they typically focus on one signal--behavior or brain data--usually in one relatively constrained task. Historically, this has been due to an inherent tension between the need for rich expressive tasks, and practical limits imposed by recording technology (e.g., EEG, fMRI). Observing a broad distribution of behaviors is important to get a broadly generalizable picture of brain representations   \cite{Gao2015-jy,Krakauer2017-ur,Mobbs2018-po,Hall-McMaster2019-an,Pinto2019-fz,Miller2022-sc,Dennis2021-de,Niv2021-fs,Driscoll2024-am}. Access to a variety of behavioral measurements and neural activity in rich semi-naturalistic tasks will provide critical complementary evidence to the more isolated datasets available through past research.

Several recent approaches demonstrate the power of large and expressive datasets in informing machine models of human cognition. A recent study trained machine learning models on a large corpus of human data that specifically related to making risky decisions \cite{Peterson2021-ok}. Another recent work assembled data from 160 psychology tasks as fine-tuning data for a LLM \cite{binz2024centaur}. These approaches produced models that could predict and simulate human behavior in a broad set of psychology tasks. Still, existing datasets are insufficient.

We need to systematically study how humans handle distribution shifts. Some of the most impressive real-world examples of human cognition consist of one-shot learning and decisions in a changing world, without the possibility for an extended trial-and-error learning period. However, many psychology tasks involve repeated experience with stationary task statistics. To study human cognition under distribution shifts will involve carefully designed experiments that introduce controlled changes to task contexts and measure how people respond, as well as fine-grained behavioral measurements in complex real-world environments. 

Recent technological advances in machine vision, simulation, and pose tracking, have made it possible to collect high-resolution behavioral datasets at scale. This includes high-resolution behavior collected in VR, where people can freely move around in an omnidirectional treadmill, while recording eye fixation, pupil dilation (an index of attention and surprise; \cite{strauch2022pupillometry,sirois2014pupillometry}), and body pose. Advances in neural models of pose detection \cite{fan2022deep,kendall2015posenet} now also make it possible to build rich datasets of naturalistic behavior in other animals, such as automatic tracking of behavior in captive marmoset colonies. 

\hypertarget{subsubsec-languages-thought-action}{%
\subsubsection{Scaling assurance: programming languages for NeuroAI}\label{subsubsec-languages-thought-action}}

Resource-rational analysis offers a powerful and unifying conceptual framework for assurance in AI safety. However, existing software doesn't scale to rational models of sophisticated AI agents, while the standard taxonomy of decision-making tasks and associated algorithmic and analysis tools is increasingly ill-matched to the complex patterns of interaction in LLM-based AI agent systems. This could be addressed by developing programming languages that perform them mechanically, much like the growth and impact of research in deep learning was catalyzed by differentiable programming languages like PyTorch \cite{Paszke2019-by}.

\hypertarget{par:neurosymbolic-language-thought}{%
\paragraph{Languages of thought}\label{par:neurosymbolic-language-thought}}

Universal probabilistic programming languages, capable in theory of expressing any computable probability distribution \cite{freer2014towards}, were first developed by cognitive scientists as promising candidates for a language of thought \cite{goodman2012church}. Since then, probabilistic programming has produced a number of widely used open source languages, such as Stan \cite{carpenter2017stan}, WebPPL \cite{goodman2014dippl}, PyMC \cite{abril2023pymc}, Pyro \cite{bingham2019pyro}, NumPyro \cite{phan2019composable}, Turing \cite{ge2018turing} and Gen \cite{cusumano2019gen}, and evolved into a thriving interdisciplinary research field in its own right \cite{van2018introduction}. 

We do not attempt to review the many technical challenges of developing general-purpose PPLs. Instead, we focus on practical limits to expressiveness in mature existing PPLs that are believed to be especially important for modeling cognition and have been individually de-risked to some extent in smaller prototypes. 

\begin{itemize}
    \item \textbf{Using foundation models as primitive distributions:} First, the language should make it possible in practice to represent the widest possible variety of hybrid neurosymbolic models \cite{chaudhuri2021neurosymbolic} that use deep neural networks to parametrize conditional probability distributions. The logical limit of this direction is foundation models as primitive conditional distributions, particularly the key special case of foundation model conditional distributions over latent probabilistic programs \cite{wong2023word}. 
    \item \textbf{Incorporating modular inductive biases:} Researchers have produced copious evidence for common  architectural modules of human cognition such as perception, mental simulation, pragmatic language understanding, working memory, attention, and decision-making \cite{sumers2023cognitive,wong2023word}.
    While we advocate for implicit specification via suites of tasks and large-scale behavioral datasets, a universal language for cognitive modeling should be able to incorporate explicit specifications as well, and allow them to be studied, modified, and validated independently while maintaining clear interfaces with other components.
    Cognitive models that invoke these modules enable transparent tracking of information flow between different cognitive processes, making it possible to understand how different components interact. This architecture creates natural points for implementing safety constraints and monitoring, as the behavior of each module can be verified independently of others \cite{Dalrymple2024-sf}.
    The cognitive foundation models we propose can be viewed as a newer generation of cognitive architecture, leveraging massive datasets and flexible models to scale to a much wider set of more naturalistic behaviors.
\end{itemize}

\hypertarget{par:interactive-tasks-probabilistic-inferences}{%
\paragraph{Languages of action}\label{par:interactive-tasks-probabilistic-inferences}}
In performing resource-rational analysis of complex behavior or designing or interrogating a resource-rational AI agent, it is often necessary to formally specify a task. This is subtly distinct from the aim of the previous section: cognitive models relate an agent’s internal beliefs and observations, while a task relates an agent’s sensors and actions to the state of the outside world. In practice, tasks are often thought of as monoliths that are either exactly the same or formally unrelated. When faced with a new task, a researcher must either force it into an existing template that allows reuse of algorithms and analysis tools while eliding important structure, or develop a new specification and implement new algorithms and analyses from scratch.

	An alternative and more scalable approach is to specify tasks as executable programs. There are many theoretical results in cognitive science that reduce higher-level tasks to sequences of one or more probabilistic inference computations \cite{botvinick2012planning, Friston2010-wp, milli2021rational, pearl2011algorithmization, solway2012goal}. Despite the leverage afforded by a single computational bottleneck, these results tend to be treated as something akin to folk wisdom, often cited but rarely examined or implemented. However, we believe that other advances have already de-risked these from open-ended research questions into well-scoped engineering problems:

    \begin{itemize}
        \item \textbf{Identifying task programming primitives:} We require a concrete set of primitive programming constructs which can combinatorially generate many tasks of interest and have known reductions to probabilistic inference. Graphical modeling notations in causal inference extend the probabilistic graphical modeling notations that later grew into PPLs \cite{koller2009probabilistic}. These include structural causal models and counterfactuals \cite{Pearl2009-bi}, which can represent any causal inference task and have already been implemented in prototype causal PPLs \cite{tavares2021language}, as well as the closely related formalism of influence diagrams \cite{dawid2002influence} and their recent mechanized \cite{macdermott2023characterisingdecisiontheoriesmechanised} and multi-agent \cite{hammond2024reasoning} extensions. Influence diagrams enable generalization of classical control/RL-as-inference reductions in cognitive science \cite{botvinick2012planning,Levine2018-ul} and unify many problems in AI safety like reward hacking and value alignment \cite{everitt2019modelingagisafetyframeworks,kenton2023discovering,richens2024robustagentslearncausal,everitt2021reward}. An orthogonal set of primitives are first-class reasoning operations that may be recursively nested and interleaved with model code \cite{tavares2019random,stuhlmuller2014reasoning}, allowing compositional specification of inverse reasoning and meta-reasoning tasks like inverse reinforcement learning in terms of their forward or object-level counterparts \cite{evans2017agentmodels,goodman2016probmods2}. 
        \item \textbf{Encoding optimization, logic and resource usage:} A complete language for resource-rational cognitive modeling should also go beyond probability to allow interleaving similar first-class programming constructs for three other modes of declarative reasoning, namely optimization, logic, and the use of limited computational or other resources. These four modes are essential for performing rational analysis and are closely related mathematically \cite{belle2016semiring}.
        \item \textbf{Specifying inference algorithms:} We advocate for algorithms that asymptotically match exact inference with large compute, and whose specifications are primarily learned rather than handwritten. Research in computational cognitive science has led to many individual algorithms that partially satisfy one or both of these, from the original wake-sleep algorithm \cite{hinton1995wake} to amortized variational inference \cite{stuhlmuller2013learning} to modern gradient-based meta-learning \cite{binz2024meta}. In the next section, we argue for a maximalist interpretation of these goals over the use of existing algorithms.

    \end{itemize}

\hypertarget{subsubsec-foundation-models-cognition}{%
\subsubsection{Scaling robustness: foundation models of cognition}\label{subsubsec-foundation-models-cognition}}

Cognitive scientists have identified properties of human intelligence that make it robust in the face of unanticipated tasks and environmental shifts.
Chief among these is the ability to manipulate representations of existing knowledge, to compose them into new knowledge and skills.
These representations are understood formally as hybrid neural-symbolic programs \cite{chaudhuri2021neurosymbolic} in a probabilistic "language of thought" \cite{piantadosi2016four,wong2023word}, and the synthesis and evaluation of these programs is formalized as Bayesian or probabilistic inference problems, whose solutions make adaptive and optimal use of people's limited computational resources \cite{lieder2020resource}. These discoveries are ripe for consolidation into a unified computational toolkit for assessing the robustness of existing AI systems and designing new ones that exhibit human-like robustness by construction.

\begin{figure}[h!] 
    \centering
    \includegraphics[width=.9\linewidth]{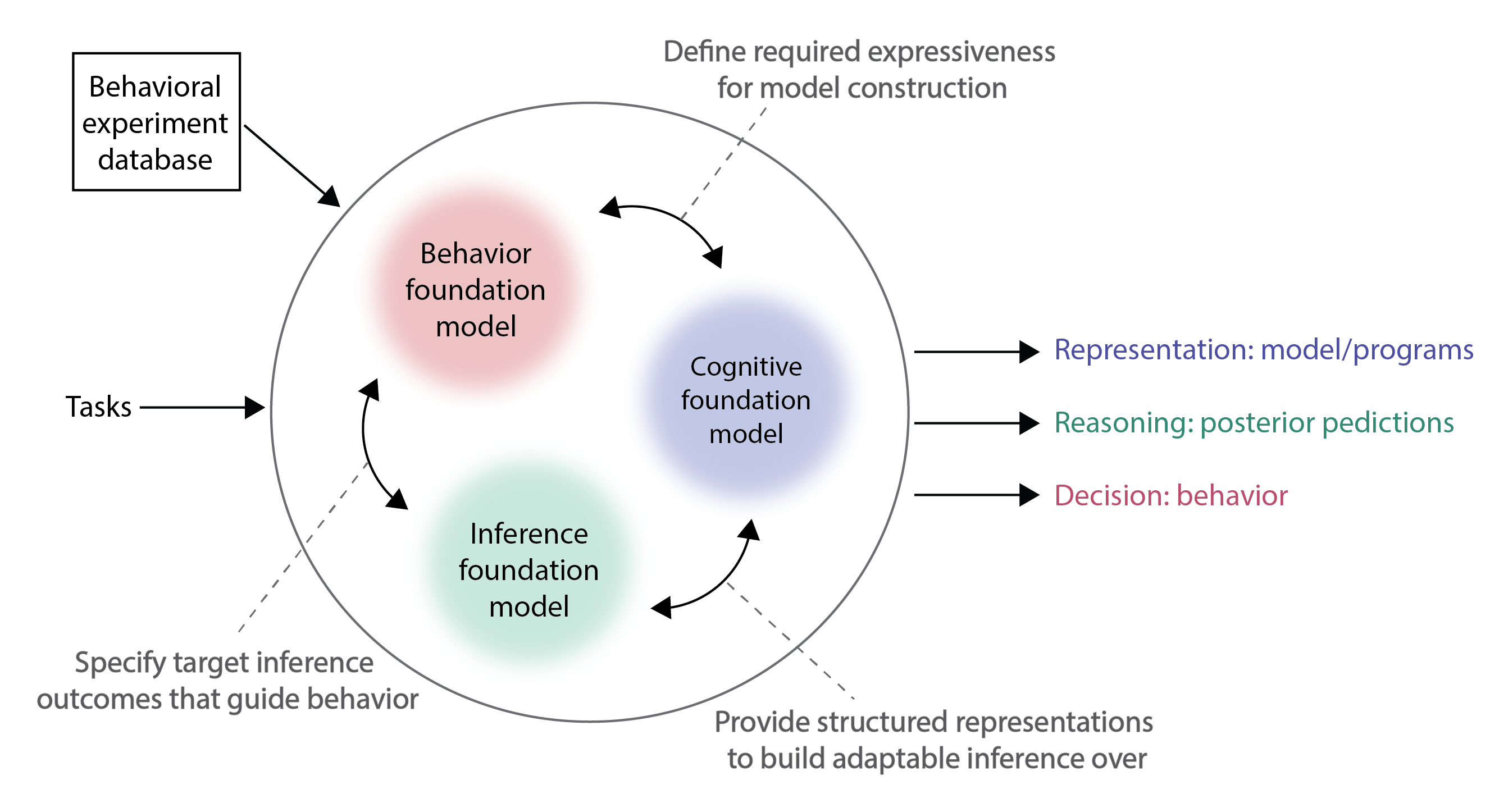}
    \caption{A foundation model of cognition that comprises a behavior foundation model, a cognitive foundation model, and an inference foundation model. The input to the model is a task. The output is a full description of how an intelligent system would represent the task in context, reason about it, and make a decision. These are formalized as model programs, posterior predictions and behavior, respectively.}
    \label{fig-cogsci-models}
\end{figure}

We propose to transform this approach through the development of foundation models of cognition that can automatically synthesize and scale up the specification of models and algorithms. We propose an approach that builds on the theoretical insights of resource rationality to maintain interpretability as a core design principle. We argue for a layered approach whereby each model layer extracts a different type of information from the massive training data. In particular, we propose the development of behavioral foundation models to predict and simulate behavior in different task domains; cognitive foundation models to capture the representations and symbolic structures underlying behavior; and inference foundation models to specify the inference algorithms in settings of limited resource availability (Figure \ref{fig-cogsci-models}). Notably, all three models can be trained on the same large-scale data described previously, but each will have different inputs and outputs, resulting in different functions being learned.

\hypertarget{par:behavioral-foundation-models}{%
\paragraph{Behavioral foundation models}\label{par:behavioral-foundation-models}}
Behavioral foundation models leverage massive experimental datasets of human or AI behavior to predict responses across a diverse set of tasks and contexts simultaneously. As a result, these models can identify common patterns and principles shared across tasks, enabling both better prediction of behavior in novel situations and the discovery of previously unnoticed regularities and biases. Additionally, the behaviors of biological organisms simulated by such models can be directly contrasted with those of normative agents and non-cognitive AI systems, surfacing differences in their responses to unexpected input.
We briefly consider data modalities and associated pretraining tasks for a behavior foundation model, closely aligned with the previous section on data collection.

\begin{itemize}
    \item \textbf{Modality: human behavior} The primary goal of a behavioral foundation model is to accurately predict the responses of a biological or AI agent across different domains, including simulating behavior in naturalistic tasks and domains outside the training set. Foundation models specified at the group level can be fine-tuned to a specific individual, capturing behavioral idiosyncrasies to enable maximally accurate predictions at the individual level. 
    \item \textbf{Modality: non-human animal behavior} Our discussion has focused on advancing AI safety by modeling human cognition and endowing AI systems with human-like capabilities. Pretraining a behavior foundation model on non-human animal behavioral data might be helpful for this goal in a few ways. For example, a foundation model can simulate behavior of a wide variety of animal species might be better at extrapolating to new types of non-cognitive AI systems, especially embodied AI systems like robots whose safety properties may be harder to anticipate and collect data on (see also Section \ref{sec-embodied} for related ideas).
\end{itemize}

\hypertarget{par:cognitive-foundation-models}{%
\paragraph{Cognitive foundation models}\label{par:cognitive-foundation-models}}

While a foundation model of behavior may capture behavior across multiple domains, it does not serve as an explanatory framework for the innate cognitive and biological constraints that shape intelligence, the learning algorithms available across development, and the kinds of behavioral pressures and experiences that drive learning. An expressive and interpretable foundation model of cognition requires going beyond matching behavior in psychology tasks, to show and explain human-like behavior in contexts that better approximate the real world, where agents have changing goals and need to construct task representations to act on the world. 

Rather than treating foundation models of behavioral data as the sole end goal, we can use them to accelerate the creation of domain-specific, interpretable, quasi-mechanistic models of cognition.
These models take the form of probabilistic programs that can be run forward to randomly generate behavioral data or run backward to explain it.
A foundation model of behavior eliminates one human bottleneck by circumventing the need to run a new human experiment to test every new model or hypothesis,
but the manual specification and adjustment of probabilistic models of cognition by human experts is still a brake on progress.

To remove this second bottleneck, we propose training a foundation model for cognitive model synthesis that learns the mapping from behavioral data and task specification to the source code of a probabilistic program.
While this is agnostic to the particular language used to represent those programs during either pretraining or testing, an obvious choice is the language described in the final section. Instead, we favor an all-of-the-above approach to curating pretraining data and tasks.

\begin{itemize}
    \item \textbf{Modality: Program source code} The proposed foundation model could synthesize probabilistic programs given a task specification and/or behavioral data. Such a model would likely reap large benefits from extending a pretrained open LLM for code, and could in theory be trained with maximum marginal likelihood on data generated from the behavior foundation model. Other pretraining datasets that might help probabilistic program generation are curating extensive corpora of probabilistic models of cognition, as well as probabilistic programs in other non-cognitive languages and domains.
    \item \textbf{Modality: Natural language reasoning chains} While we do not expect this foundation model to be used to generate natural language, reasoning chains produced by humans are a scalable and high-content source of information about cognitive processes and would likely be a useful auxiliary pretraining task, especially since the corpus of handwritten cognitive models is relatively small.
    \item \textbf{Modality: Measurable correlates of cognition} Data from human experiments might include low-level behavioral signatures of cognitive processes, such as eye-tracking or mouse movement, that could prove useful for auxiliary pretraining tasks.
\end{itemize}

\hypertarget{par:inference-foundation-models}{%
\paragraph{Inference foundation models}\label{par:inference-foundation-models}}

The probabilistic programs synthesized by the cognitive foundation model can be used in conjunction with Bayesian inference to explain behavioral data in terms of the programs’ interpretable latent structure, or to generate behavior that exhibits human-like flexibility. However, since Bayesian inference is intractable in general, to generate resource-rational behavior, it is ultimately necessary to specify an approximation that is resource-rational, a challenging and labor-intensive endeavor that typically requires human expertise and remains a primary barrier to the wider use of Bayesian methods in NeuroAI. 

To accelerate progress, we propose to learn an inference foundation model, which would directly map arbitrary resources, priors and data to resource-rational posterior predictions. We briefly consider possible output modalities and associated pretraining tasks that would leverage the previous two foundation models of behavior and cognition, as well as complementary human experimental data.

\begin{itemize}
    \item \textbf{Modality: Posterior predictive inferences.} Inference foundation models would serve as computational primitives in cognitive architectures, providing efficient approximate inference over a broad range of reasoning tasks. Training would require assembling diverse datasets of models, data, and ground truth posterior inferences. These can be constructed by mining the scientific literature for published models and inference results, generating synthetic data from cognitive foundation models, running exact inference or gold-standard methods like Hamiltonian Monte Carlo \cite{neal2011new}, and collecting human behavioral data from experiments carefully designed to elicit subjects' prior and posterior beliefs as observable outcomes.
    \item \textbf{Modality: Experimental correlates of resource expenditure.} Beyond models and inference results, a potentially useful auxiliary pretraining task is predicting measurable proxies of computational and cognitive resources expended during inference. For automated methods, we can directly track computational operations, energy usage, and economic costs. For humans, we can measure behavioral proxies like response times in decision-making tasks, eye movements in visual inference, or move times in strategic games like chess. These resource measurements enable training inference models that make human-like rational tradeoffs between accuracy and computational cost.
\end{itemize}

\hypertarget{subsec-evaluation}{%
\subsection{Evaluation}\label{subsec-evaluation}}
Building foundation models of cognition from human data presents promising opportunities for AI safety but faces several significant technical challenges:
\begin{itemize}
    \item Collecting behavioral and especially neural data of sufficient scale and quality to train foundation models is expensive and time-consuming.
    \item Training new kinds of foundation models requires substantial computational resources.
    \item Developing PPLs is engineering-intensive and technically complex, on par with developing differentiable programming languages like PyTorch.
\end{itemize}
Fortunately, the technical risks are mostly independent of one another and have standard mitigations. The potential payoff is significant: formal, executable specifications of human-like intelligence that could fundamentally change how we develop AI systems. Rather than treating safety and robustness as properties to be added after the fact, these specifications could allow us to build systems that inherit human-like inductive biases from the ground up. These could include graceful degradation under uncertainty, appropriate caution in novel situations, and stable objectives despite distribution shifts--precisely the properties needed for safe and reliable AI systems.

\hypertarget{subsec-opportunities-cogsci}{%
\subsection{Opportunities}\label{subsec-opportunities-cogsci}}
\begin{itemize}


    \item Design large-scale tasks and human experiments that implicitly specify human-like inductive biases

    \begin{itemize}
        \item Create or curate a comprehensive suite of ARC-like benchmark tasks where humans greatly outperform conventional AI systems thanks to superior inductive biases
        \item Collect large datasets of behavior, cognitive function and neural activity from humans engaged in these benchmark tasks who are tracked over many tasks and subjected to many domain shifts
    \end{itemize}
    
    \item Develop a probabilistic programming language for engineering and reverse-engineering safe AI

    \begin{itemize}
        \item Support inference in probabilistic programs that invoke multimodal LLMs as primitive probability distributions, including LLMs that synthesize probabilistic programs in the same language
        \item Augment the base probabilistic programming language with a standard library of inductive biases drawn from across cognitive science and neuroscience
        \item Enable the creation of formal, fully executable specifications of arbitrary cognitive tasks by borrowing extra syntax from causal inference and giving it a meaning through systematic reduction to probabilistic inference
    \end{itemize}

    \item Build a foundation model of cognition

    \begin{itemize}
        \item Train a \emph{behavioral foundation model} that can accurately simulate human and non-human animal behavior across many different experimental tasks, subjects and domain shifts
        \item Train a \emph{cognitive foundation model} for mapping task specifications to human-interpretable programs in a probabilistic "language of thought" that can be used to generate or explain behavior within a specific context
        \item Train an \emph{inference foundation model} for mapping programs generated by the cognitive foundation model and data to adaptive behavioral outputs
    \end{itemize}

\end{itemize}

\hypertarget{sec-use-brain-data-to-finetune-ai-systems}{%
\section{Use brain data to finetune AI}\label{sec-use-brain-data-to-finetune-ai-systems}}

\hypertarget{subsec-core-idea-bips}{%
\subsection{Core idea}\label{subsec-core-idea-bips}}

Collecting human neural data has become cheaper, safer, and more
feasible at scale. Deep brain stimulation has become routine for the
treatment of Parkinson's disease and psychiatric disorders
\cite{Delaloye2014-os,Holtzheimer2011-pv}, while
intracortical brain-computer interfaces from multiple companies are
undergoing clinical trials
\cite{Oxley2016-ig,Musk2019-dp,Sahasrabuddhe2020-nt}. In
clinical research, labs are collecting single-neuron recordings from
epileptic patients with the help of next-generation Neuropixel probes
\cite{Paulk2022-yj,Khanna2024-be,Leonard2024-zp}. New
semi-invasive and non-invasive approaches like functional
ultrasound also open the door for portable, continuous neural data
collection from humans. Non-invasive approaches like fMRI, fNIRS, and
EEG, historically dismissed because of noise issues, are now being
reappraised with deep learning techniques, with impressive results.

\begin{figure}[htbp]
   \centering
   \begin{subfigure}[t]{0.45\textwidth}
       \centering
       \includegraphics[height=2in]{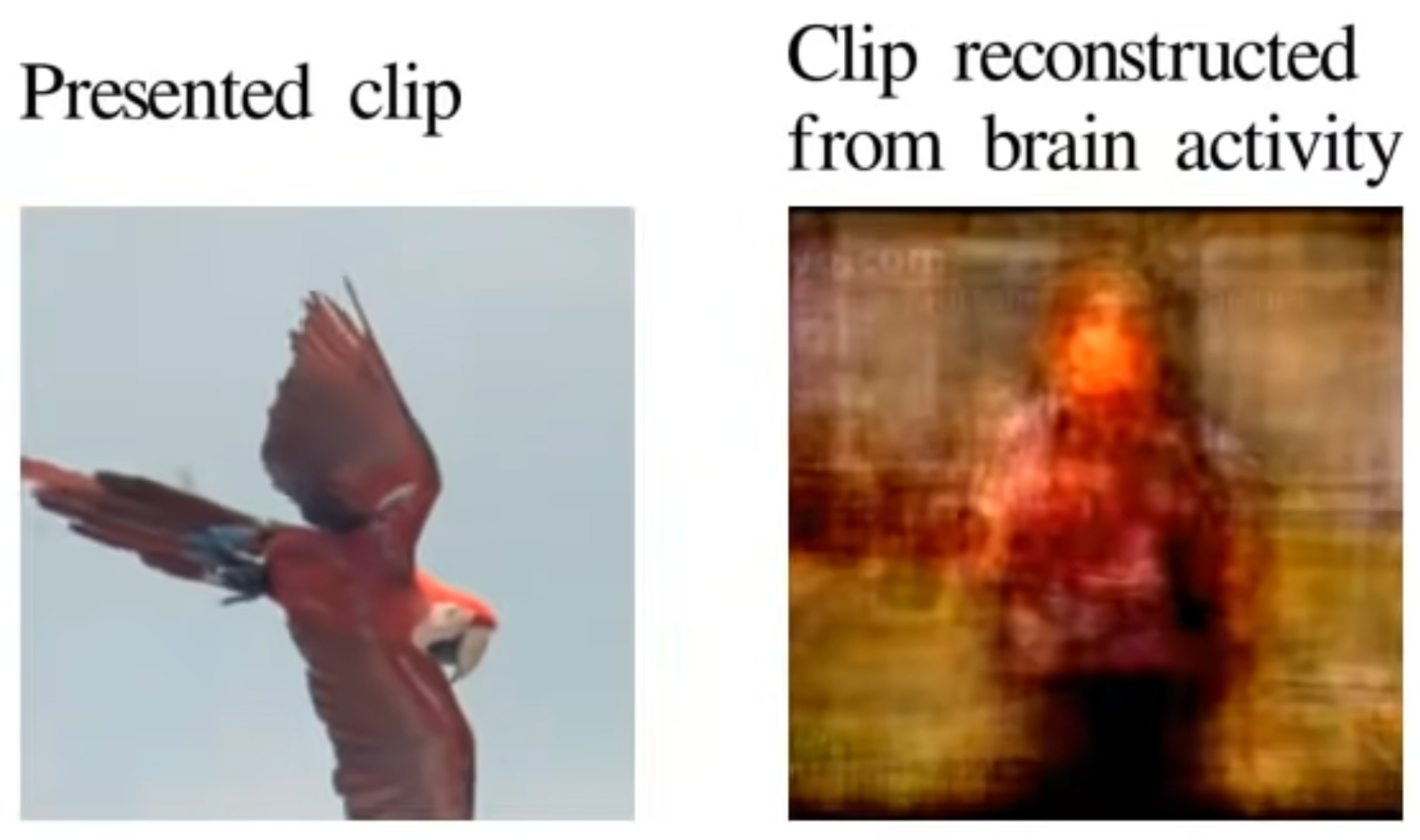}
   \end{subfigure}
   \hspace{0.05\textwidth}
   \begin{subfigure}[t]{0.45\textwidth}
       \centering
       \includegraphics[height=2in]{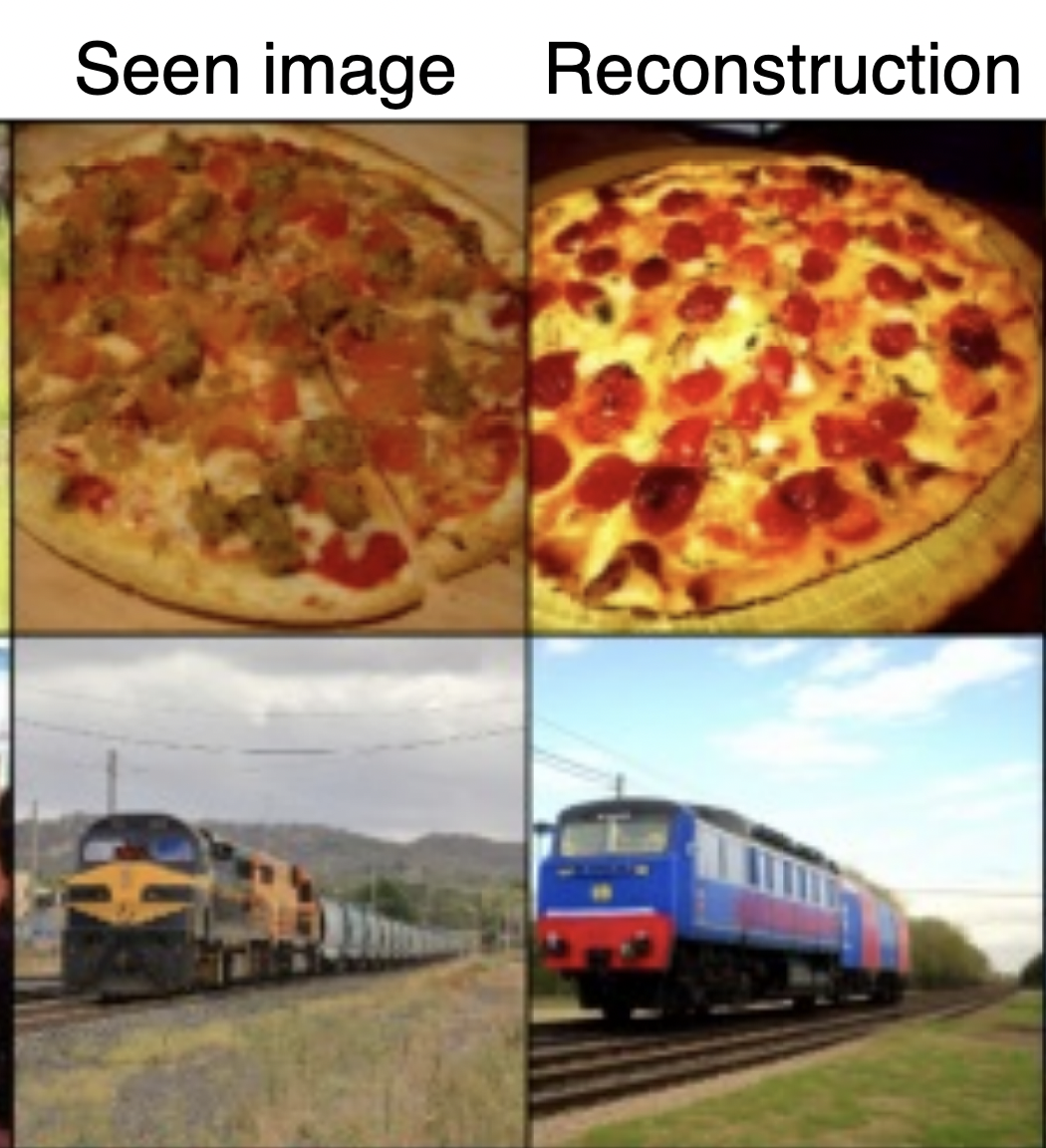}
   \end{subfigure}
   \caption{Reconstructed images from fMRI brain activity. Top: State-of-the-art reconstruction in 2011 \cite{Nishimoto2011-ha}, still from \href{https://youtu.be/nsjDnYxJ0bo}{youtu.be/nsjDnYxJ0bo}. Bottom: State-of-the-art reconstruction in 2023, adapted from \cite{Scotti2023-in}, under a CC-BY license. More capable reconstructions from non-invasive brain activity implies that directly training AI systems using brain data could be more feasible than in the past.}
   \label{fig-fmri-reconstruction}
\end{figure}

Currently, machine learning relies on large datasets of human-derived
outputs or behavior, such as choices/preferences via reinforcement
learning from human feedback \cite{Christiano2017-ew},
statistics of natural language, and behavioral cloning for robotics.
This typically provides a signal (consistent interactions with a user, a
large corpus of written text, 3D tracking of body movements) that can be
augmented with synthetic data. While these approaches have been
remarkably successful, behavioral signals do not always reflect the
process by which representations, decisions and movements have been
created; it is possible to clone the behavior of a human by a process of
imitation using shortcut learning in such a way that the system will
fail catastrophically out-of-distribution
\cite{Geirhos2020-uk}. Here, we evaluate the idea that
rich process supervision signals derived from neural data or from
complex behavioral data motivated by cognitive science could help better align
AI systems with the human mind.

\hypertarget{subsec-why-does-it-matter-for-ai-safety-and-why-is-neuroscience-relevant-bips}{%
\subsection{Why does it matter for AI safety and why is neuroscience
relevant?}\label{subsec-why-does-it-matter-for-ai-safety-and-why-is-neuroscience-relevant-bips}}

Humans are capable of organizing their behavior safely, including the
ability to collaborate in groups towards shared goals, create shared
laws and institutions, and agree on group status and trustworthiness.
Recent work has focused on aligning these models to humans at the
behavioral level, including through reinforcement learning from human
feedback \cite{Christiano2017-ew}.

Human neural data--and more generally, complex behavioral data motivated
by cognitive science--has seldom been used for labeling large-scale
datasets. Latents measured from the brain could provide richer sources
of information about decision-making processes than what is visible
through behavior, and demonstrate higher validity when evaluated
out-of-distribution \cite{Fong2018-sv}. Latents measured
from the brain could provide an especially rich supervision signal in
cases where the supervision signal would be otherwise hard to obtain:

\begin{enumerate}
\def\labelenumi{\arabic{enumi}.}
\item
  \emph{Tasks where explicit labeling is difficult}. This includes moral
  judgment, complex emotions, and long-term decision-making.
  \item
  \emph{Tasks where it's hard for humans to explain or demonstrate how they've
  accomplished a task}. This includes automatic responses e.g. avoiding
  accidents in driving, intuition about social interactions, emotional
  understanding, non-verbal cues, and metacognition
  \cite{Nisbett1977-yo}.
  \item
  \emph{Relatedly, tasks where the target is the \emph{process} by which a
  decision is made}. This includes complex reasoning and multi-step
  planning \cite{Lightman2023-qi,Luo2024-rc}.
  \item
  \emph{Cognitive processes where the key measurement of interest is the
  latent state itself}, and its match or mismatch with behavior: e.g.
  motivation, intent, malice, duplicitousness, sycophancy, judgments
  about trust, morality, ethics, etc.
  \cite{Hendrycks2020-fv,Sharma2023-hi,Shen2023-at,Phuong2024-wj,Muller2020-fb}.\end{enumerate}

Finetuning with brain data can be thought of as a form of \emph{process supervision}.
Process supervision \cite{Lightman2023-qi,Luo2024-rc} is
a form of supervision where models are not only incentivized to obtain
the correct behavioral outputs, but also to mimic the desired steps
leading to the behavioral output--in other words, not only focusing on
the \emph{what}, but also on the \emph{how}. Process supervision has
proven useful in improving mathematical and general reasoning
capabilities in large language models. It is also likely behind some of
the reported improvements from reasoning in recent models, such as
OpenAI's o1 model \cite{OpenAI2024-ma}.
Although process supervision can be partially automated by leveraging
highly curated datasets and content generation
\cite{Abdin2024-qt}, it is generally tedious and
expensive to collect high-quality process supervision datasets. Brain
data--and more generally, rich behavioral data motivated by cognitive
science--reflects the dynamic process by which behavior is produced and
could prove an alternative, rich reservoir of process supervision
\cite{Branwen2018-mc}. Current small-scale experiments
suggest that such approaches could be more robust to domain shifts
\cite{Sucholutsky2023-fm,Muttenthaler2023-pq,Muttenthaler2024-qc}.
More generally, \emph{brain-informed process supervision}\footnote{We considered the alternative evocative term RLBF, reinforcement learning from brain feedback, but many of the methods we present use plain supervised learning rather than reinforcement learning.} (BIPS) has the
potential to instill deeper levels of alignment between humans and
machines than what is currently feasible. Furthermore, there is
extensive literature on representational alignment which can help guide
the development of neural alignment methods
\cite{Sucholutsky2023-fx}. In light of these advances,
we evaluate the feasibility of aligning human and AI representations in
the general case.

\hypertarget{subsec-details-bips}{%
\subsection{Details}\label{subsec-details-bips}}

\hypertarget{subsubsec-cognitively-inspired-process-supervision}{%
\subsubsection{Cognitively-inspired process
supervision}\label{subsubsec-cognitively-inspired-process-supervision}}

Before we cover brain-data-based finetuning methods, we briefly cover the
related field of cognitively-inspired process supervision. Most data
leveraged for machine learning is ultimately a product of human
cognitive labor. Whether it's text on the internet or image labels,
ultimately, humans performed cognitive work to produce these artifacts.
It's thus worth reflecting on why exactly the final products would be
misaligned with human cognitive processes.

Consider, for example, the process by which the ImageNet dataset was
collected \cite{Deng2009-kn}. ImageNet is based on the
WordNet taxonomy \cite{Fellbaum2005-jk}, a
large lexical database of English that was manually compiled in the
1980s and 1990s from a broad range of sources. Dictionaries and thesauri
were pored over by linguists and lexicographers over a decade to form a
set of hierarchically organized synonym sets (synsets). The concrete
nouns in these synsets were used as keywords to image search engines
such as Flickr to assemble hundreds to thousands of images from each of
thousands of categories. These images were deduplicated, filtered, and
annotated by people on the Amazon Mechanical Turk (AMT) platform.
ImageNet-1k (ILSVRC 2012) was a subset of 1.2M images from 1000
categories, chosen to be semantically distinct and visually concrete.
The 1000 categories include 90 different dog breeds. Thus, while the
images are a reflection of cognitive labor, and are vastly more
representative of human visual experience than the datasets that
preceded it, they are far from representative samples of the visual diet
of any individual. Thus, it would be quite surprising if the natural
categories learned from such a dataset or the internal representations
derived from them were representative of human natural categories or
representations
\cite{Mehrer2021-ud,Huh2024-ng,Gauthaman2024-jx}.

To finetune AI systems to better reflect human natural categories or their internal
representations, researchers have turned to collecting rich behavioral
data inspired by cognitive science, a set of approaches that we
collectively refer to as \emph{cognitively-inspired process
supervision}. One key technique in this domain is the use of triplet
losses and similarity judgments
\cite{Roads2021-gs,Sucholutsky2023-fm,Muttenthaler2023-pq,Muttenthaler2024-qc}.
Human judgments about the relative similarity of stimuli 
are collected to fine-tune the representations of models trained conventionally. By incorporating these pairwise
or triplet comparisons, models can learn to organize information in ways
that more closely reflect human conceptual spaces
\cite{Hebart2023-ui,Borghesani2023-rn}. A representative
approach \cite{Muttenthaler2024-qc} uses human triplet
similarity judgments to create a teacher model, which then generates a
large dataset of human-like similarity judgments on ImageNet images.
This dataset is used to fine-tune vision foundation models, resulting in
representations that better reflect human conceptual hierarchies. The
finetuned models show improved performance on human similarity judgment
tasks, better generalization in few-shot learning scenarios, increased
robustness to distribution shifts, and better performance on
out-of-distribution datasets.

A related approach involves the use of soft labels
\cite{Sucholutsky2023-fm} which incorporate uncertainty
\cite{Peterson2019-ao}. Rather than using hard, binary
classifications, soft labels allow for more nuanced representations of
category membership and uncertainty. This better reflects the graded
nature of human categorization and can lead to more robust and
generalizable models. Relatedly, error consistency and misclassification
agreement \cite{Xu2024-ck} can be used to guide models.
Triplet losses, soft labels, confidence, and error-geometry-aware
methods can help finetune generic representations obtained from
self-supervised or supervised learning to better approximate the
hierarchical structure of humans' natural categories.

Another common form of process supervision investigated in vision models
is replicating attention and eye gaze patterns to guide models
\cite{Klerke2016-tb,Ishibashi2018-ms,Linsley2018-sj,Linsley2023-tp}.
Human vision has much higher acuity in the fovea than in the visual
periphery, and eye gaze patterns are indicative of which features and
objects are used to make visual judgments. By training models to
replicate human patterns of visual attention and gaze, these approaches
aim to create AI systems that focus on robust features in a human-like
way. For example, Lindsey et al. (2023) \cite{Lindsey2023-xh} fine-tune pre-trained vision
models with the ClickMe dataset, where humans are asked to identify
important image parts by clicking on them.
Activations in convolutional layers--a proxy for attention
\cite{Itti1998-da}--are trained to match ClickMe
annotations. The resulting models rely less on high-frequency, brittle
features for classification, are better matched to single neuron
representations in the inferotemporal cortex, and are more resistant to
L2-norm adversarial attacks. Eye-tracking has also been extended to
provide a supervision signal to natural language processing models
\cite{Mathias2021-mw}, and recently to large language
models \cite{Kiegeland2024-fw}.

Although rarely identified as such, cognitively-inspired fine-grained
supervision methods are conceptually related to teacher-student
distillation \cite{Hinton2015-wd}, and to process
supervision \cite{Lightman2023-qi,Luo2024-rc}. While
collecting process supervision data can be tedious, and ripe for biases
due to failures in metacognition \cite{Nisbett1977-yo},
it is generally appreciated that evaluating labels is easier for humans
than producing them in the first place. There is a large, and currently
underexplored design space in augmenting models with
automatically generated process supervision traces curated and
relabeled by humans. This could potentially include enhancing models
with annotated moral reasoning traces; providing feedback on visual
question-answering reasoning formed by vision-language models (VLMs);
providing metacognitive feedback to foundation models; and creating
cooperative decision systems.

\hypertarget{subsubsec-process-supervision-through-neural-data-augmentation}{%
\subsubsection{Process supervision through neural data
augmentation}\label{subsubsec-process-supervision-through-neural-data-augmentation}}

Neural data reflects not just the outcome of a decision but the process
through which that decision was made. Could we train or finetune AI
models directly on neural data as a form of process supervision \cite{Fong2018-sv}? Several different schemes are possible, including:

\begin{itemize}
    \item \emph{Fine-tune a sensory model to match the latent representations of neural activity}. Fine-tune a vision or language model to match the representations during image viewing or podcast listening, for example using representational similarity analysis \cite{Kriegeskorte2008-id,Kornblith2019-dy,Williams2021-uu,Sucholutsky2023-fx}, or through linear readouts \cite{Safarani2021-ui}. 
    \item \emph{Fine-tune a reasoning model so that it matches the thought process of a human on difficult tasks}. Similar to how LLMs with process supervision learn to generate hidden language tokens as scratch pads \cite{Luo2024-rc}, fine-tune LLMs to generate hidden brain tokens, then condition text generation on those hidden brain tokens.
    \item \emph{Fine-tune a heuristic search model} (e.g. Monte Carlo tree search to play Go \cite{Silver2016-jf} or to generate programs \cite{Xie2024-rz}) so that it tends to visit nodes with a frequency congruent in neural activity.
\end{itemize}

We covered fine-tuning vision models with brain data in Section \ref{subsubsec-feasibility-of-transferring-robustness-from-brains-to-models} on neural data augmentation. Here we focus on applications to language, which offers a more direct link to semantics, reasoning and moral decision making, all relevant to AI safety. In a very early study, Alona Fyshe and colleagues
\cite{Fyshe2014-nr} reported that fMRI or MEG could be
used to find word representations that were better matched to behavior
than conventional embeddings. However, consistent with the time period,
this was based on a very small dataset (60 words) with linear word
embedding methods.

It has become increasingly feasible to read language information from brain data \cite{Huth2012-md,Tang2022-fz,Antonello2023-tb, Moses2021-ki,Kohler2022-kp,Card2023-hx}, opening new opportunities for brain-based fine-tuning. Moussa et al. (2004) \cite{Moussa2024-za} recently showed
that fine-tuning speech models using fMRI recordings of people listening
to natural stories ("brain-tuning") led to improved alignment with
semantic brain regions and better performance on downstream language
tasks. Their approach, validated across three different model families
(Wav2vec2.0, HuBERT, and Whisper), not only increased alignment with
brain activity in language-related areas, but also reduced the
models\textquotesingle{} reliance on low-level speech features.
Importantly, the brain-tuned models showed consistent improvements on
tasks requiring semantic understanding, such as speech recognition and
sentence-type prediction, while maintaining or improving performance on
phonetic tasks.

While these results are highly promising, we have yet to see a clear
demonstration of brain data used to finetune models in a way directly relevant to AI safety. \cite{Meek2024-tv} reported null
results on fine-tuning large language models on fMRI data collected in
the context of moral decision-making. This may simply be due to
the paucity of data that was used for fine-tuning
\cite{Saxe2003-bv}. There is a largely unexplored design space around fine-tuning AI models with large-scale brain data.

\hypertarget{subsec-evaluation-bips}{%
\subsection{Evaluation}\label{subsec-evaluation-bips}}

Fine-tuning vision and vision-language models with cognitively-inspired
behavioral data and neural data has been demonstrated to improve
out-of-distribution robustness, one-shot learning, and alignment to
human natural categories. Noninvasive neural and cognitive data
augmentation is a natural complement to building digital twins of vision
systems (Section \ref{sec-reverse-engineer-the-representations-of-sensory-systems}); while digital twins can capture the micro-geometry
of image recognition in non-human animals, neural and cognitive data
augmentation can capture the macro-geometry of visual representations
with a higher level of fidelity than has been possible through
conventional labeling pipelines. Thus, there exists a promising synergy
between building digital twins of sensory systems based on invasive data
and fine-tuning through neural data augmentation. This would allow the
creation of models that reflect the human visual system at all relevant
levels, which we hope will be investigated more thoroughly in the near
future.

By contrast, there has been little work in finetuning language and
auditory models with brain data, although this is starting to change
\cite{Moussa2024-za}. Improving upon existing
large-scale models trained on internet-scale datasets will likely
require lengthy, high quality recordings of neural activity. Because of
noise, lengthy recordings are necessary to accurately estimate a mapping
between neural activity and an artificial neural network
\cite{Antonello2023-tb,Sato2024-cg}. This suggests
recording from a small number of subjects for an unusually long period
(at least tens of hours, ideally hundreds) in ideal conditions, e.g.
with a headcase in fMRI with optimal coil positions and a highly
optimized scanning sequence. This design, sometimes called
\emph{intensive} neuroimaging, is being investigated by a growing number
of laboratories \cite{Allen2022-bq,Kupers2024-sf}. For
example, the Courtois Neuromod project recorded people watching 6
seasons--or roughly 50 hours--of the TV show Friends.
Preliminary results indicate that fine-tuning auditory models on this
data could lead to gains in downstream tasks
\cite{Freteault2024-th}, consistent with
\cite{Moussa2024-za}.

Before engaging in such large-scale neural data collection, however, one
would need to establish a niche where neural data collection provides added value over
behavioral data collection. We foresee
two scenarios where this could be the case. As we alluded in the
introduction, in one scenario, the information of interest cannot be
elicited through behavior \cite{Ariely2010-lo}.

A second scenario is when neural data collection could win over behavioral
data on a bits-per-dollar basis. That means matching behavioral data on both an
information density basis and on marginal cost. This is a high bar to
hit, as behavior is easy to collect at scale through crowdsourcing
platforms, and it can be less noisy than neural data. Speech is
estimated to transmit information at an average rate of 39 bits/s across
a wide range of languages \cite{Coupe2019-il}.
\cite{Zheng2024-in} estimate that a wide range of overt
behaviors are capped at 10 bits/s. Do any brain data modalities approach
that benchmark? We performed a broad survey of noninvasive and invasive
neural data modalities in humans in the context of brain-computer
interfaces (Table \ref{tab-itr}).

\clearpage

\setlength\LTleft{-.3in}  
\setlength\LTright{-.1in}  

\begin{longtable}{@{}
    p{0.28\textwidth}  
    p{0.25\textwidth}  
    p{0.3\textwidth}  
    p{0.17\textwidth}  
    @{}}

\toprule
\textbf{BCI type} & \textbf{Reference} & \textbf{Task} & \textbf{Reported avg ITR (bps)} \\
\midrule

Intracortical & 
\cite{Willett2021-jl} & 
Handwriting decoding & 
6.56 \\
\hline

Electrophysiology & 
\cite{Willett2023-yc} & 
Speech decoding & 
13.33 \\

~ & \cite{Card2023-hx} & 
Speech decoding & 
8.69 \\

~ & \cite{Neuralink2024-bu} & 
Cursor control (grid task) & 
8.00 \\
\hline

fMRI & \cite{Scotti2024-my}& 
Visual retrieval & 
3.24 \\

& \cite{Antonello2024-mw} & 
Text decoding & 
6.95 \\
\hline

EEG & \cite{Parthasarathy2024-my}& 
Free spelling & 
1.31 \\
\hline

SSVEP-based EEG & \cite{Shi2024-wd} & 
Free spelling & 
16.86 \\
\hline

SSVEP-based MEG+EEG & \cite{Li2023-vf} & 
Visual decoding & 
5.20 \\
\hline

SSVEP-based MEG & \cite{Li2023-vf} & 
Visual decoding & 
4.53 \\
\hline

OPM-MEG & \cite{Kernel2021-jh} & 
Spelling & 
1.31 \\
\hline

HD-DOT & \cite{Tripathy2021-xw} & 
Visual information decoding & 
0.55 \\
\hline

fNIRS & \cite{Shin2018-bo}& 
Ternary classification & 
0.078 \\
\hline

fUS & \cite{Norman2021-ib} & 
Movement intention decoding & 
0.087 \\

\bottomrule
\caption{State-of-the-art in reported information
transmission rates across brain-computer interface modalities and
paradigms.}\label{tab-itr}
\end{longtable}

Keeping in mind the difficulty of comparing very different modalities,
tasks, and experimental paradigms on an apples-to-apples basis, we find
that the 10 bits/s benchmark is within the reach of intracortical
recordings, and that fMRI is within a factor two of that. MEG and EEG
state-of-the-art results are dominated by steady-state visual evoked
potential (SSVEP) paradigms, which are not as directly relevant to
finetuning AI systems with neural data. We note that while functional
ultrasound decoding results reported are quite modest, they are based on
a very small imaging volume; fields of view thousands of times larger
are attainable in theory, with consequent increases in information
transmission rate, which could put it above the threshold.
Finally, while not included in this table because it has not been used
thus far in humans, a recent preprint on a flexible thin-film
micro-electrocorticography array demonstrated decoding visual stimuli at
a rate of 45 bits/s \cite{Jung2024-mg}.

Thus, methods that could conceivably hit the 10 bits/s benchmark include
electrophysiological recordings, ECoG, fMRI, and functional ultrasound.
fMRI has high marginal costs due to the need for dedicated rooms,
technicians, and refrigerants for magnets. This leaves intracortical
recordings, ECoG, and functional ultrasound as potential candidates for
scaling up neural data augmentation. Conventional fMRI, which is broadly
available in academic settings, can continue to serve as
proof-of-concept data acquisition method for neural data augmentation
schemes.

Beyond vision and language, neural data could be used for fine-tuning
robotics, RL, and agentic models. A significant challenge is collecting
sufficiently rich data over the long time scales necessary to work
around the high noise in neural recordings. There is a long history of
using games to study rich cognition \cite{Allen2023-ub},
including board games like Othello and chess, bespoke games like Sea
Hero Quest \cite{Spiers2023-hf}, or popular games like
Bleeding Edge \cite{Devlin2021-om}, Little Alchemy, and Overcooked. There has been some work in comparing the representations of
neural networks trained with reinforcement learning with those obtained
during games, e.g. Atari games \cite{Cross2021-go},
16-bit era 2D games \cite{Kemtur2023-ei}, or driving
simulators \cite{Zhang2021-jj}. However, to the best of
our knowledge, there have been no attempts to finetune machine learning
models on these datasets.

Collecting neural data during gameplay could provide a scalable way to
obtain the large-scale paired datasets necessary to elucidate how higher
cognitive processes operate in the brain, and eventually fine-tune
machine learning models to match these cognitive processes. To maximize
the impact on both neuroscience and AI safety, we foresee the use of
open-ended, complex games that have a high loading on cognitive
phenomena involving ethical behavior, cooperation, and theory of mind,
and that have served as challenges for AI. This could include Minecraft \cite{Hafner2023-ox}, Diplomacy
\cite{Meta-Fundamental-AI-Research-Diplomacy-Team-FAIR-+2022-oz},
Hanabi \cite{Bard2019-wb}, poker
\cite{Brown2020-em} or text adventures like the
Machiavelli benchmark \cite{Pan2023-si}.

\hypertarget{subsec-opportunities-bips}{%
\subsection{Opportunities}\label{subsec-opportunities-bips}}

\begin{itemize}
\item
  Demonstrate the feasibility of fine-tuning large language models on
  extant fMRI data\item
  Collect rich process supervision data--including neural data--for
  cognitive phenomena of interest to AI safety, including those that
  extend beyond vision and language: moral and social decision-making,
  decision-making under uncertainty, cooperation, and theory of mind\begin{itemize}
  \item
    Use AI safety-relevant, ecologically valid, and intrinsically
    motivating games\item
    During neural data recording, focus on long recording times (10
    hours+) per subject to mitigate high noise\end{itemize}
\item
  Reduce the barrier to entry of AI safety researchers wanting to work
  with neural data by the development of findable, accessible,
  interoperable, and reusable (FAIR) preprocessed datasets in
  machine-learning-friendly formats\item
  Encourage the development of non-invasive or minimally invasive
  neurotechnology that can go beyond the 10 bits/s barrier, including 
  functional ultrasound \cite{Norman2021-ib} and micro-ECoG \cite{Jung2024-mg}
  
\end{itemize}

\hypertarget{sec-build-an-evolutionary-curriculum}{%
\section{Build an evolutionary curriculum}\label{sec-build-an-evolutionary-curriculum}}

\emph{This section contributed by Anthony Zador.}

\subsection{Core Idea}

The ultimate goal of AI can be seen as an attempt to reverse engineer the end product of evolution—the human brain. The brain emerged through 500 million years of evolution, and capabilities we take for granted, such as perception, motor control, and planning, remain extraordinarily challenging for artificial systems. Yet it is precisely these capabilities that are universal to even the simplest animals \cite{moravec1988mind}. This suggests that rather than reverse engineering the finished product, we should recapitulate the process that created it: evolution. \textbf{\textit{We propose an ``evolutionary curriculum" in which AI capabilities emerge in stages that mirror evolutionary history—from basic sensorimotor control to complex cognition—where each stage builds on previously established competencies. }} Just as shared evolutionary foundations enable the domestication of animals through common neural circuitry, this approach aims to develop AI systems that can be naturally aligned with human values—what we term \textbf{\textit{``AI domestication.''}} By explicitly incorporating the sequential nature of evolutionary innovation, while retaining the advantages of gradient descent, this approach differs from purely learning-based methods. It combines a sequence of tasks with mechanisms for accumulating and refining capabilities across generations, offering a robust path to safe and capable autonomous AI systems.

\subsection{Why Does It Matter for AI Safety and Why Is Neuroscience Relevant?}

Dobzhansky's famous observation that ``Nothing in biology makes sense except in the light of evolution'' \cite{dobzhansky1973nothing} extends naturally to artificial intelligence. The capabilities needed in AI systems--including but not limited to robust sensorimotor processing-- didn't appear fully formed, like Athena from Zeus's head. They emerged through a sequence of evolutionary innovations, each new capability building upon existing ones. As noted by Moravec \cite{moravec1988mind}, tasks often considered the pinnacle of human intelligence--those requiring abstract thought like chess, mathematics, symbolic manipulation--build on half a billion years of evolution; they are ``not all that intrinsically difficult; they just seem so when we do it."  Counterintuitive though it seems, the step from animal to human intelligence is evolutionarily quite small: modern humans emerged less than a million years ago.

A thought experiment illuminates this point: if evolution's ``training run'' were compressed into the equivalent of a modern deep learning timeline, the development of human-level abstract reasoning---the ability to do mathematics, play chess, write poetry---would represent just the final few epochs, less than 0.2\% of the total training time. \textit{\textbf{The vast majority of nature's computational resources---its equivalent of GPU cycles---went into developing and refining the foundational  capabilities that all animals share.}} This suggests that if our goal is to develop human-like intelligence, we might benefit from following a similar sequence. \textit{\textbf{We must first establish robust foundations for perception, action, and planning, after which capabilities that rely on abstract reasoning represent an easy step.}} Although it is tempting to jump directly to refined cognitive abilities while bypassing these foundational capabilities, doing so risks repeating the mistakes of the GOFAI (good old fashioned AI) research program in the 1960s-1980s, which discovered that symbolic reasoning could not be effectively deployed without first solving these more fundamental challenges.

Consider  alien archeologists trying to make sense of Earth technology. In their attempt to reverse engineer a car, they might take three possible approaches. The first approach—replicating its function—dominates the mainstream of AI development: instead of building a car, they craft something— a skateboard, a sled, a train—that moves people from place to place, attempting to mimic the car's utility without recreating its design. This characterizes most modern AI research, which aim to match behavioral capabilities without necessarily matching biological mechanisms. Thus, most modern AI engineers attempt to build artificial systems that match human performance for tasks such as object recognition or text understanding, without explicit reference to how humans actually perform this these tasks. The second approach, replicating its parts, represents much of current neuroAI. In this scenario, the aliens disassemble the car and try to replicate it piece by piece. This spans from detailed biophysical models that simulate every component to more abstract approaches that aim to match neural representations and dynamics. Though appealing, this strategy becomes challenging if the aliens don't understand the function of each part—imagine failing to realize that the tire needs to be elastic and so making it out of steel, while at the same time meticulously reproducing every groove on its surface. In the absence of such functional knowledge, the strategy becomes unwieldy when attempting to replicate every neural mechanism at microscopic detail. Here, we propose a third approach: replicating the design process. Instead of focusing on the sophisticated car itself—with its complex computer systems, safety features, and intricate engine management—we study how the first automobiles emerged from simpler predecessors. We examine how the earliest cars, with their basic mechanical systems and straightforward functions, evolved into today's complex vehicles through successive innovations. For example, the modern engine control unit—a black box of inscrutable computer code—evolved from the mechanical carburetor, where you could literally see the fuel being mixed with air through the action of visible jets and float chambers. This  approach is particularly appealing when we consider biological intelligence, where evolution proceeded from simple organisms to increasingly sophisticated ones. Just as automotive complexity emerged gradually through clear functional improvements, the human brain evolved through a series of adaptations, each building on previous capabilities. This parallel suggests that in AI development, replicating the evolutionary process that shaped intelligence may be easier than attempting to reverse engineer its end product directly.

An evolutionary curriculum could address several fundamental challenges in AI development. First, it could improve \textit{\textbf{data efficiency.}} Current approaches typically ingest massive datasets to produce models and then discard those training data. Although foundation models have made progress in knowledge transfer, simply scaling these models may, as Yann LeCun and others have argued, represent a dead end. A curriculum that builds capabilities progressively, similar to biological evolution and development, has the potential to offer a more efficient path.

\begin{figure}[t]
\centering
\includegraphics[width=\textwidth]{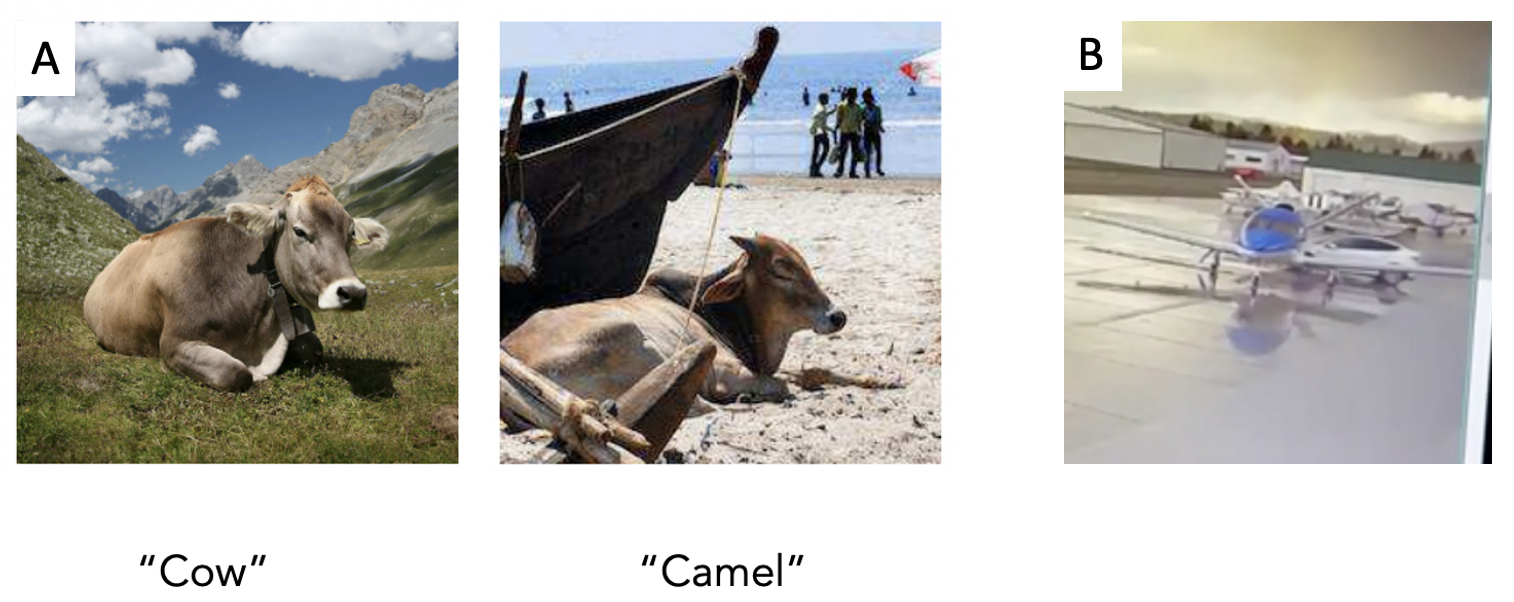}
\caption{Examples of alien errors. (A) A classic alien error is mislabeling a cow on a sandy background as a camel \cite{beery2018recognition}. (B) A real-world failure in which a ``summoned'' Tesla crashed into a private jet in a scenario far from the training data (\url{https://rb.gy/s3sah2}). Both of these specific errors have likely been corrected in updated systems. However, there is a long-tail of such alien errors so other such errors will persist. Trying to solve them individually will be akin to a game of whack-a-mole. The complete ensemble of alien errors cannot be addressed in a piecemeal fashion.}
\label{fig-alien-errors}
\end{figure}

Second, this approach \textit{\textbf{could improve out-of-distribution generalization.}} Current AI systems often make ``alien errors"—mistakes that no human or animal would make. To paraphrase Yogi Berra, we don’t want our systems to make ``too many wrong mistakes;” we want only ``right” (human-like, or more generally animal-like) mistakes. For example, when an AI image recognition system misclassifies a cow on a sandy background as a camel \cite{beery2018recognition}, it does so because it has given too much significance to the background (Fig. \ref{fig-alien-errors}). This is a sensible strategy for a single-purpose AI system, because backgrounds turn out to be fairly reliable cues for labeling foreground objects in static images; but it is a ``wrong mistake"--an alien error that a human would not make. Although each specific alien error, such as overweighting the background can be corrected, addressing the complete ensemble of potential alien errors through specific fixes represents an unwinnable game of whack-a-mole. By contrast, because the visual systems of animals evolved to detect potential predators, prey, or mates from the background, natural visual systems give much less weight to background structure, even when classifying static images. Natural systems repurposed the core systems adapted for evolutionarily more ancient vision tasks to the labeling of static images. \textit{\textbf{To avoid alien errors in artificial systems, AI systems must develop similar core competencies through an evolutionary sequence, allowing them to acquire the same foundational priors that evolution has instilled in biological systems.}}

This possible payoff of following an evolutionary curriculum is even greater when pursuing the goal of \textit{\textbf{autonomous behavior}}, the third challenge. While AI systems excel at specific tasks, they lack the flexible autonomy displayed by even simple organisms—the ability to explore unfamiliar environments, adapt behavior through experience, and pursue multiple goals over multiple time scales. Moreover, \textit{\textbf{truly autonomous agents must actively interact with their environments—allowing them to move beyond merely detecting correlations in passive data and instead internalize the causal structure of the world. }}This fundamental capacity for autonomous behavior, shaped by evolution in all animals, remains elusive in artificial systems.

One might wonder why we are focusing here on evolution rather than learning. In neuroscience, learning refers to a lasting change in behavior that results from experience, and learning has been central to AI in recent years. However, \textbf{\textit{we can also view evolution as a learning process, but one where we can distinguish between the ``inner loop" of traditional learning in an individual animal and the ``outer loop" of optimization in a population of individuals.}} In proposing an evolutionary curriculum, we are in fact proposing a learning process over generations.

The potential payoff of well aligned AI emerging from an evolutionary curriculum is significant. Consider an autonomous vehicle whose overall safety record matched or even exceeded human performance. Even if rare, errors that were qualitatively different from human mistakes—such as bizarrely swerving to run over a child playing in a yard—would make such a system unacceptable for deployment, regardless of its superior safety record. By building AI systems through an evolutionary sequence that parallels biological development, we expect to develop artificial agents that not only perform well but also fail in ways that are more predictable and aligned with human expectations.

\subsection{Details}

The evolutionary curriculum approach requires two components. First, it requires ``genomic bottleneck” algorithms for capturing a key feature of evolution—namely, that every generation begins with a single zygote, which contains only a relatively small amount of genetic information required to generate the neural circuits underlying behavior. Second, it requires rich environments in which artificial agents can follow a specified evolutionary curriculum. Given that this evolutionary process is the only path known to have produced human-like intelligence, we believe that the most promising way to avoid alien errors and balance multiple objectives is to replicate a similar evolutionary process in silico. Concretely, this means using algorithms that progress through an interactive curriculum, with each generation of AI agents building on the success of its predecessors. In other words, simulated evolution provides a platform for constructing new forms of AI that can overcome Moravec’s paradox. 

Crucially, \textit{\textbf{our approach diverges from many prior evolutionary methods in that it does not use evolutionary search}}. Although there is a rich literature on genetic algorithms \cite{katoch2021review}, which seek to minimize functions without using gradients, \textit{\textbf{we are not proposing to use evolutionary algorithms}} to find minima. Here, however, it is not the gradient-free aspect of evolution that we seek to replicate, but rather the gradual progression from simple to complex. We therefore propose to exploit gradients in solving evolutionary optimization problems. This ``in the weeds” technical detail reflects several decades of machine learning experience demonstrating that gradient-based methods are invaluable for navigating very high-dimensional search spaces.


\subsubsection{Why learning is not enough} 

Traditional ANNs depend on learning. Weights are initialized \textit{tabula rasa}, drawn from a simple distribution, and are tuned through data. This approach requires massive datasets for training, often consisting of millions or billions of examples. Once training is complete, these datasets are typically discarded, and the statistical structure extracted from these data sets cannot be efficiently reused for application to new tasks without extensive retraining. 

In recent years, foundation models have emerged as a partial solution to data inefficiency. In foundation models, the statistical structure of a large data set is extracted and the foundation model can then be ``fine-tuned" for subsequent tasks. This approach has had remarkable successes, particularly in large language models.  However, a fundamental limitation remains: Major model updates still require training from scratch—GPT-4 was not built upon GPT-3's learned representations.  (While one might argue that hyperparameters capture some evolutionary learning across models \cite{kaznatcheev2022nothing}, they represent only a tiny fraction of the total information needed—the vast majority of computational work still involves reprocessing the entire training corpus for each new model generation.) This contrasts sharply with biological systems, which have evolved efficient mechanisms for encoding and propagating knowledge across generations.

\subsubsection{Genome as compression: Genomic Bottleneck Algorithms}

\begin{figure}[t]
\centering
\includegraphics[width=0.33\textwidth]{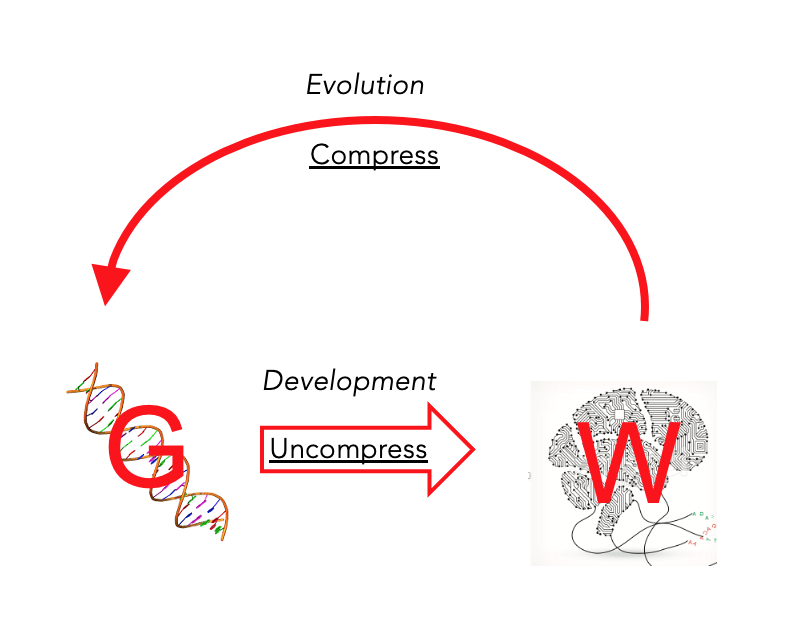}
\caption{The genome (G) represents a compressed encoding of ancestral experiences, which development then uncompresses into our initial neural wiring patterns, analogous to the weight matrix (W) in ANNs. This genome encompasses both our innate priors and the learning mechanisms we employ throughout life. While the biological process of natural selection  optimizes genomes through gradient-free hill-climbing, here we propose to employ gradient-based optimization algorithms, which can accelerate this process significantly. }
\label{fig-genomic-bottleneck}
\end{figure}

Nature's solution is based on evolution and development (Fig. \ref{fig-genomic-bottleneck}). Each of us started our life as a single cell. Within that cell is the DNA that contains the instructions needed to build our bodies and wire up our brains. Our behavioral capacities, including both the innate abilities with which we are born and the strategies that enable us to learn after birth, are stored in our genomes and passed from generation to generation. The human brain consists of trillions of connections, which in principle would be expected to require as much as $10^{15}$ bits of information to specify, whereas the human genome contains only about $3 \times 10^9$ nucleotides, or roughly 1 GB of data \cite{shuvaev2024encoding,zador2019critique}. This mismatch--an apparent gap of at least six orders of magnitude--implies that it is impossible to explicitly specify every neural connection. Instead, the genome encodes generalizable rules for wiring the brain; these rules enable the wiring diagram to be ``uncompressed'' during development, allowing for the emergence of the complex neural circuits underlying innate behaviors and setting the stage for all information and behaviors learned during a lifetime. This mismatch--at least six orders of magnitude--between the apparent complexity of the brain's wiring diagram and the limited size of the genome is what we refer to as the genomic bottleneck.

What ends up in the genome is, in effect, a compressed representation of life experience accumulated across generations. Evolution can be viewed as a lossy compression algorithm, distilling the statistical regularities of an organism’s ancestral experiences into the genome. This code is then uncompressed during development to construct the neural circuits that give rise to behavior, and provide suitable and often very rich priors for learning. Developmental processes such as axonal guidance instantiate these encoded rules, allowing a small genome to give rise to a highly complex and functional brain. \textbf{\textit{This compression mechanism is not a limitation, but a powerful regularization process. By encoding only the most essential and generalizable features, the genome ensures robustness and adaptability. }} Moreover, this compression naturally filters out bizarre ``alien errors'' that might appear in specific overfit solutions but are unlikely to persist in the general class of solutions encoded by the genome. When an error pattern does survive this genomic bottleneck, it suggests that error reflects something fundamental about the task structure or environment, rather than an arbitrary quirk of training.

Several algorithms have been developed to compress weight matrices through genome-like mechanisms, inspired by the genomic bottleneck that encodes complex neural circuits using limited genetic information. The Deterministic Genomic Bottleneck (DGB) model \cite{shuvaev2024encoding}, inspired by the biological observation that brain wiring is partly determined by expressed ligands and receptors guiding neural connectivity, employs a smaller genomic network to deterministically encode the weight matrix of a larger phenotypic network. This deterministic approach provides a precise mapping from genome to connectome, ensuring robust performance upon initialization. Building on this, the Stochastic Genomic Bottleneck (SGB) introduces probabilistic sampling of synaptic weights \cite{lachi2024stochastic}, encoding distributions rather than fixed values. Motivated by the biological concept of cell types, the SGB specifies connection probabilities between these types, enabling the generation of diverse neural networks. Mathematically, the SGB is a generalization of Bayesian neural networks, where the genome encodes priors over synaptic weights. This incorporation of stochasticity induces phenotypic variability, fostering specialization to diverse environments and enhancing adaptability.

The compression in both models acts as a regularizer, allowing networks to generalize effectively while maintaining innate performance. The SGB demonstrates superior performance at higher compression levels compared to the DGB, particularly in larger networks or variable contexts. By sampling from encoded distributions, the SGB produces a population of networks whose collective performance surpasses deterministic counterparts, effectively leveraging randomness as a functional advantage. Both approaches can be seen as a form of Occam's Razor by favoring simpler, generalizable representations, facilitating better generalization and transfer learning. Together, the DGB and SGB models offer biologically inspired frameworks for designing robust and adaptive AI systems.

\subsubsection{Evolution as a curriculum} 

The intuition for how an evolutionary curriculum can help develop robust AI capabilities is beautifully illustrated in the classic film ``The Karate Kid." The sensei, Mr. Miyagi, famously teaches his student karate by having him perform repetitive motions like waxing cars (``wax on, wax off") - basic movements that form the foundation for defensive blocks in karate. This pedagogical approach mirrors how evolution has built complex capabilities: by first establishing fundamental building blocks that can be combined and refined into more sophisticated behaviors. Just as mastering martial arts requires building upon these foundational movements rather than jumping straight to advanced techniques, AI systems might benefit from a similar developmental sequence. This principle becomes clear when we examine modern computer vision systems, which, unlike animals that evolved to detect predators, prey, and mates by discounting backgrounds, developed brittle solutions from being trained directly on labeled images. Had these systems instead been built through an evolutionary curriculum - starting with tasks that require detecting and tracking potentially mobile objects against varying backgrounds - they would likely have developed more robust visual primitives that generalize better across contexts. This illustrates a broader principle: by forcing artificial systems to solve problems in an order that parallels biological evolution, we expect to  develop more robust and generalizable capabilities.

To understand how an evolutionary curriculum can facilitate the discovery of complex solutions, consider a simple problem: searching for a target leaf in a binary tree of depth $N$. While highly simplified, this abstraction captures some essential features of searching through parameter space for good solutions: the challenge of finding a specific solution in a high-dimensional space. At each node in the tree, we must decide whether to go left or right. Each leaf represents a possible solution, i.e. one that minimizes some objective function given the data. The target leaf represents the target solution represents the specific solution that corresponds to what humans have learned, and thus the one that generalizes out of distribution in the appropriate way by avoiding  alien errors. The path from root to leaf represents the sequence of parameter choices needed to reach that solution. In the absence of guidance, we would need to exhaustively explore all $2^N$ possible paths to guarantee finding the target. This exponential complexity mirrors the challenge faced by systems attempting to find complex solutions from scratch. Perfect guidance in this context would mean having a reliable signal at each node indicating whether the left or right branch leads closer to the target solution. With such feedback, the problem becomes tractable: we can simply follow these signals, reducing the time complexity from exponential ($2^N$) to linear ($N$). However, such perfect fitness signals rarely exist for complex problems. More typically, the fitness landscape becomes increasingly difficult to navigate as the problem becomes more complex.

This is where the power of a curriculum becomes clear. Instead of attempting to reach the final target directly, we can introduce $k$ intermediate waypoints -- partial solutions that represent meaningful evolutionary steps. These waypoints decompose the journey into $k$ segments, each of depth approximately $N/k$. The sequence of these waypoints represents a curriculum. Within each segment, we still face uncertainty, but because each segment is shorter, the total search space for each segment is much smaller—$2^{N/k}$ rather than $2^N$. The total time to reach the target becomes $k \cdot 2^{N/k}$, which can be dramatically less than $2^N$ for suitable choices of $k$. To make this concrete, consider a parameter space requiring 20 binary choices. Finding the target through exhaustive search requires exploring up to $2^{20} \approx 1$ million paths. But if we introduce just 4 waypoints, dividing the journey into 5 segments of depth 4, the search space for each segment is only $2^4 = 16$ paths. The total search involves exploring at most $5 \cdot 16 = 80$ paths—four orders of magnitude fewer than the exhaustive search.

While this model vastly simplifies the challenge of finding complex solutions via e.g. gradient descent, it illustrates a fundamental principle: decomposing the search for a complex target into a sequence of simpler intermediate solutions can exponentially reduce the search space. This mirrors biological evolution, where complex capabilities emerged through a sequence of intermediate evolutionary innovations, each building upon previous achievements. Just as no organism evolved the ability to climb trees without first mastering basic terrestrial locomotion, we suggest that artificial systems may benefit from a similar staged progression through parameter space.

\begin{figure}[t]
\centering
\includegraphics[width=0.8\textwidth]{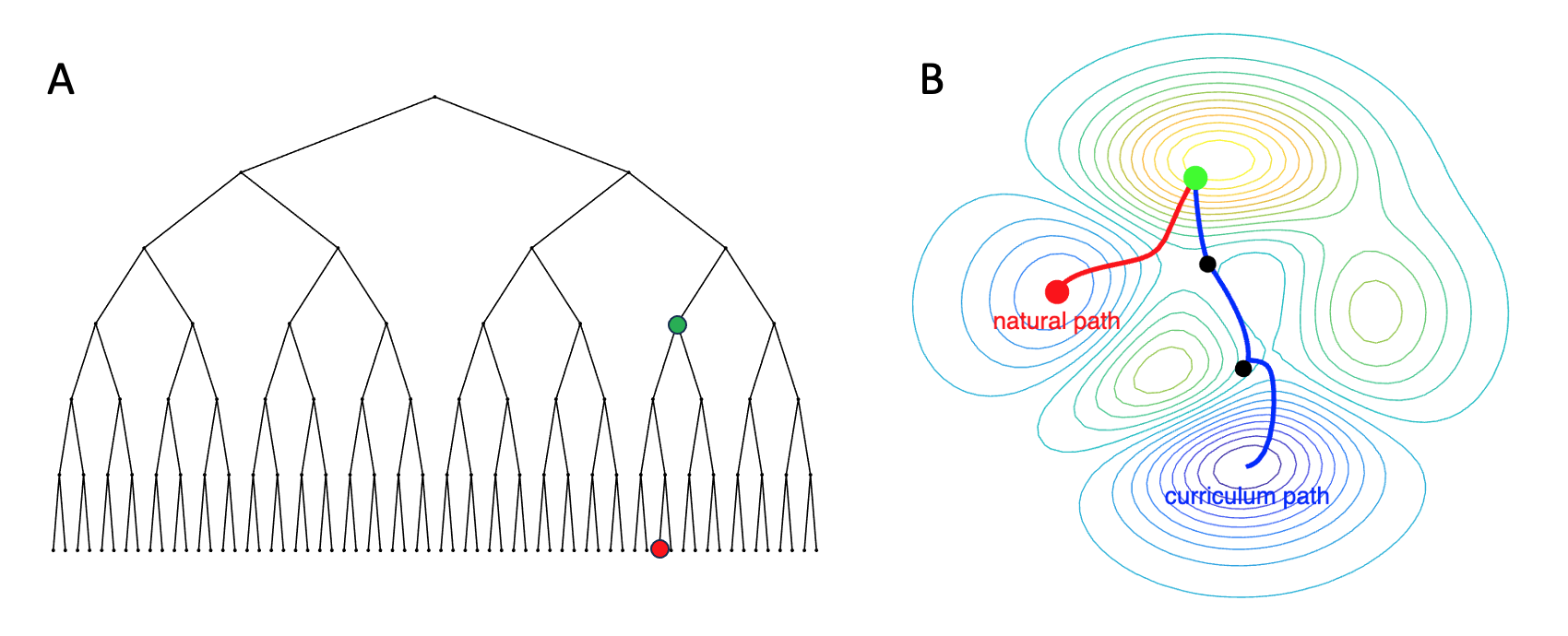}
\caption{\textbf{Evolutionary curriculum simplifies search for solutions.} (A) Cartoon motivating intuition underlying evolutionary curriculum. Searching for a specific solution (\textit{red}) could take $2^N$ where $N=6$ is the tree depth. But if we know that the \textit{green} waypoint is along the way, this reduces the search to $k=2$ easier searches of complexity at worst $2^N/k$. For deep trees even a handful of waypoints can dramatically reduce search time. (B) Illustration of how this might work in a loss landscape traversed by some form of gradient descent. A few waypoints can guide the model to some target away from its natural path and toward a desired outcome, such as the solutions that animals including humans have evolved.}
\label{fig-surface_3D}
\end{figure}

\subsubsection{Autonomy: Balancing Multiple Objectives}

\begin{quote}
\textit{Suppose we have an AI whose only goal is to make as many paper clips as possible. The AI will realize quickly that it would be much better if there were no humans because humans might decide to switch it off. Because if humans do so, there would be fewer paper clips. Also, human bodies contain a lot of atoms that could be made into paper clips. The future that the AI would be trying to gear towards would be one in which there were a lot of paper clips but no humans.}
\begin{flushright}
--- Nick Bostrom
\end{flushright}
\end{quote}

Bostom's cautionary tale of the Paperclip Maximizer is a striking example of the challenge of balancing multiple objectives. In this \textit{reductio ad absurdum} thought experiment, an AI with apparently innocuous but narrow goals poses an existential threat because it pursues these narrow goals at the expense of everything else. Even when a narrow goal is perfectly well-aligned to a specific human objective (e.g.\ making paperclips, driving a car, writing emails, etc.), an AI system can still be limited or harmful if it cannot properly balance multiple objectives. Put another way, the Paperclip Maximizer is just one manifestation of the larger problem of balancing multiple objectives and limits the utility of AI even when the stakes are not as high as the extinction of the human race.

In practice, classical machine learning systems have difficulty with balancing multiple objectives because the usual approach is to express the goal in the form of a single objective function. Traditionally, our main way of adding multiple objectives is to include additional terms in the objective function and set their relative weights to one another. However, this turns out to be a rather inflexible approach, and often the result is simply that one sub-objective dominates the learning, depending on the  the weight given to each term. For example, consider trying to implement Isaac Asimov's Three Laws of Robotics (1950)\footnote{Asimov's Three Laws of Robotics are: (1) A robot may not injure a human being or, through inaction, allow a human being to come to harm; (2) A robot must obey orders given it by human beings except where such orders would conflict with the First Law; (3) A robot must protect its own existence as long as such protection does not conflict with the First or Second Law. Asimov's three laws provide an appealing and intuitive, if naive, guide for ensuring that AI does not engage in dangerous or damaging behaviors.} as a single objective function. One might try to proceed by simply creating an objective comprised of three terms:

\begin{equation}
L = \beta_1 \text{Law}_1 + \beta_2 \text{Law}_2 + \beta_3 \text{Law}_3
\end{equation}
where $L$ is the total loss, ``Law$_{1...3}$'' are the laws, and $\beta_{1...3}$ are the weights corresponding to each law. However, practical experience with this approach in AI suggests it would be very challenging to find precise weights for each law that allow them to all be  satisfied, under the appropriate conditions;  instead one law would dominate the others. Although modern approaches to balancing multiple objectives, such as hierarchical reinforcement learning, have  emerged, this remains a major concern. 

The misalignment problem is a special case of the more fundamental problem faced by animals: the need to balance multiple, often competing and sometimes contradictory, objectives. This is ultimately another manifestation of Moravec's paradox: balancing multiple objectives is precisely what is most challenging for AI today, but it is a basic competency for all animals. For example, a thirsty zebra grazing in the savannah can decide whether to risk the danger of drinking from the crocodile-infested watering hole. Five hundred million years of evolution have provided us with the machinery necessary to balance multiple objectives because all organisms, from the earliest forms of life to the present day, have had to balance the so-called ``four F's:'' feeding, fleeing, fighting, and mating. The remarkable ability of animals to balance these objectives effectively arose from an evolutionary process that required autonomous interaction with a complex world. Only those agents that successfully balanced constraints during these embodied interactions in the wild had progeny and propagated to the next generation. This propagation across generations led to an accumulation of ever more sophisticated capabilities in balancing objectives across the animal kingdom. Current AI systems, in contrast, have no mechanism to select for the ability to balance multiple objectives well, let alone multiple objectives in a shifting, interactive world.

\subsubsection{Proposed Solution: A Platform for Evolutionary AI}

The solution to these challenges lies in creating a platform for artificial 
evolution—one that can overcome the limitations of traditional approaches while 
capturing the essential features that made biological evolution successful. What 
makes our approach distinctive is its focus on recapitulating the evolutionary 
process itself, rather than attempting to directly mimic its end products.

\textbf{Platform Overview.} Consider how nature solved the problem of developing intelligence. It didn't start  by trying to create abstract reasoning or language capabilities. Instead, it began  
with basic sensorimotor control—the ability to move through an environment and 
respond to stimuli. Each subsequent innovation built upon these foundational 
capabilities, creating increasingly sophisticated behaviors through what we might 
call a natural curriculum.

We propose to develop a platform technology that will allow for an accelerated 
version of this process. Rather than waiting millions of years for random mutations 
to produce improvements, we'll leverage modern AI tools like gradient descent to 
efficiently navigate the space of possible solutions. This platform will consist of 
two key components: an extensive curriculum of embodied tasks that mirror 
evolutionary progression, and an algorithm for efficiently propagating successful 
solutions across generations.

\textbf{Key Components.} The first component of our platform is a comprehensive sequence of increasingly 
complex interactive tasks. These begin with fundamental challenges like locomotion 
and basic sensory processing—tasks that mirror the capabilities of our earliest 
animal ancestors. The curriculum advances through stages that parallel major 
evolutionary transitions, starting with basic sensorimotor control and 
goal-directed movement. From there, agents progress to integrating multiple 
sensory modalities, including vision and proprioception. As capabilities develop, 
the curriculum introduces predator-prey dynamics and competitive behaviors, 
followed by cooperative social interactions. The final stages focus on 
sophisticated manipulation of the environment, including tool use. Each stage of this progression will be supported by appropriate training 
environments and datasets. Early stages might use physics-based simulations for 
locomotion tasks, while later stages could incorporate first-person video data of 
human interactions. Importantly, these environments will be embodied—agents will 
need to learn to balance multiple objectives and constraints, just as biological 
organisms do.

\textbf{\textit{The social stages of this curriculum are particularly crucial for AI safety.}} Agents that develop through similar evolutionary stages are more likely to share fundamental behavioral primitives and value structures, creating a natural basis for alignment. This mirrors how domesticated animals can readily interact with humans despite significant cognitive differences---they have common low-level neural and behavioral machinery that enables cooperative interaction. Consider how dogs, through domestication, became remarkably adept at reading and responding to human social cues. This was possible because dogs, as social mammals, share fundamental neural circuits for social interaction, emotional processing, and behavioral regulation with humans.  Similarly, AI systems built on shared foundational capabilities would be more amenable to alignment with human values. This alignment emerges naturally from having developed through similar evolutionary stages rather than being trained directly on high-level social tasks. We term this approach \textbf{\textit{``AI domestication''}}---the systematic development of AI systems with fundamental circuits that can be naturally aligned with human values, just as biological domestication co-opted existing neural circuits for human-animal cooperation.

An additional component of the curriculum could include not only tasks, but also learning strategies. One promising organizing framework could be based on the five key evolutionary breakthroughs in intelligence delineated by Max Bennett in "A Brief History of Intelligence" \cite{bennett2023brief}. Bennett's framework traces how cognitive capabilities emerged through distinct evolutionary stages: from steering (the ability of early bilaterians to move purposefully in response to stimuli), to learning from trial and error (as seen in vertebrates), to internal simulation (a key mammalian innovation), to mentalizing (the primate capacity to model others' mental states), and finally to speaking (the human mastery of language). These breakthroughs provide natural waypoints for our curriculum, offering a biologically-grounded sequence for developing AI capabilities. Early stages would focus on developing robust sensorimotor control through steering tasks, progressing to environments that require trial-and-error adaptation. As agents advance, they would tackle tasks requiring internal modeling and simulation, followed by multi-agent scenarios that demand social cognition and theory of mind. The final stages would incorporate communication and language abilities. This progressive development of capabilities mirrors how biological evolution solved increasingly complex cognitive challenges, with each stage building upon and incorporating previous competencies rather than attempting to develop advanced capabilities in isolation.

The second key component is an algorithm for efficiently propagating successful 
solutions across generations. Unlike natural evolution, which relies on random 
mutation and selection, our approach will use gradient-based optimization to 
efficiently explore the space of possible solutions. This will be combined with a 
form of ``Lamarckian" inheritance, where learned capabilities can be directly 
passed to subsequent generations. A critical innovation in this component will be the implementation of a genomic 
bottleneck—a compression mechanism that forces the system to discover general, 
robust solutions rather than memorizing specific responses. Just as the biological 
genome represents a compressed encoding of ancestral knowledge, our artificial 
genome will need to capture essential patterns while filtering out noise.

\subsection{Evaluation}
We expect this platform to produce AI systems with several distinctive 
characteristics. Through the foundational nature of early capabilities, these 
systems will demonstrate more robust generalization to novel situations. By 
following a more natural developmental progression, they will be less likely to 
make the kind of ``alien errors" that plague current AI systems. The embodied 
training process will lead to better handling of multiple competing objectives, 
while the multi-generational transfer of knowledge will enable more efficient 
learning through the reuse of previously discovered solutions. Importantly, while this platform could be applied to specific domains like 
robotics, its potential applications are much broader. The fundamental principles 
it develops—robust sensorimotor processing, multi-objective optimization, and 
hierarchical decision-making—are relevant to any AI system that needs to interact 
with the real world in a human-like way. By following this approach, we aim to develop AI systems that not only match human 
capabilities but do so in a way that is more aligned with human cognition and 
behavior. This alignment is crucial for developing AI systems that we can trust to 
act appropriately in novel situations and to make decisions that properly balance 
multiple competing objectives.

To evaluate the effectiveness of our evolutionary platform, we propose several 
complementary approaches that assess different aspects of the system's performance 
and alignment with human behavior.

\subsubsection{Evaluating Generalization and Robustness}

A key test of our approach will be whether systems trained through our evolutionary 
curriculum generalize better than those trained directly on end tasks. We will 
evaluate this through a series of systematic out-of-distribution tests. For 
instance, in visual tasks, we will assess whether our systems maintain performance 
when encountering novel backgrounds, lighting conditions, and viewpoints. More 
importantly, we will test whether they exhibit human-like error patterns—do they 
fail in ways that align with human perceptual limitations rather than making alien 
errors?

A concrete example would be comparing our system's performance to current 
computer vision models on the ``cow on beach" problem. While current systems often 
misclassify cows on beaches as camels due to background dependence, we expect our 
evolutionarily-trained systems to show more human-like foreground object 
recognition, maintaining accurate classification even in novel contexts.

\subsubsection{Measuring Multi-objective Balance}

To evaluate how well our systems balance multiple competing objectives, we will 
develop a suite of scenarios that require trading off different goals, similar to 
the classic approach-avoidance conflicts studied in animal behavior. For example, 
we might present agents with situations where they must balance obtaining rewards 
with avoiding risks, or managing multiple tasks with limited resources.

We will compare our agents' behavior patterns to those observed in animals and 
humans facing similar tradeoffs. Success here would mean not just achieving good 
performance, but doing so in ways that mirror natural decision-making patterns. 
This includes showing appropriate risk sensitivity, temporal discounting, and 
context-dependent preference shifts.

\subsubsection{Assessing Knowledge Transfer}

The efficiency of our multi-generational knowledge transfer can be evaluated by 
measuring how quickly new generations of agents acquire capabilities compared to 
training from scratch. We will track metrics like sample efficiency (number of 
interactions needed to achieve proficiency), transfer learning performance (ability 
to adapt to new but related tasks), and the preservation of core competencies 
across generations.

\subsubsection{Genomic Compression Efficiency}

We will evaluate the effectiveness of our genomic bottleneck by measuring both the 
compression ratio achieved (comparing the size of the compressed genome to the full 
neural network) and the quality of the reconstructed capabilities. A successful 
implementation should achieve high compression while maintaining or even improving 
performance through better generalization.

\subsubsection{Long-term Evaluation Strategy}

Given the ambitious nature of our platform, we propose a staged evaluation 
process that parallels our curriculum. Each evolutionary stage will have its own 
success criteria and evaluation metrics, allowing us to validate our approach 
incrementally rather than waiting for the full system to emerge. This incremental 
evaluation strategy will help us identify and address challenges early in the 
development process. Importantly, all these evaluations will be conducted with an eye toward 
reproducibility and rigorous comparison with baseline approaches. We will maintain 
detailed documentation of our testing protocols and release standardized 
evaluation environments to facilitate comparison with alternative approaches.

\subsection{Opportunities}

The evolutionary curriculum framework opens several immediate opportunities for research and development:

\begin{itemize}

\item \textbf{Enhanced AI Safety:} By grounding AI training in biologically inspired principles, systems can achieve greater alignment with human values and expectations. Just as successful domestication of animals relies on shared neural circuits for social behavior, ``AI domestication'' aims to develop artificial systems with compatible foundational processes. Rather than imposing explicit rules or constraints, this approach develops AI systems that naturally share core behavioral primitives with biological intelligence. This offers a promising path toward AI systems that can be reliably aligned with human interests, much as domesticated animals have developed stable cooperative relationships with humans through shared social and emotional circuitry.

\item \textbf{Scalable Development:} Multi-generational inheritance reduces computational costs by reusing learned solutions, enabling scalable AI development.

\item \textbf{Platform Development:} There is an immediate opportunity to create an open-source simulation platform that implements the evolutionary curriculum approach. This would include developing standardized environments, metrics, and benchmarks for testing embodied AI systems across evolutionary stages.

\item \textbf{Genomic Bottleneck Optimization:} Current genomic bottleneck algorithms could be extended to handle larger networks and more complex architectures. Specific opportunities include developing more efficient compression methods and exploring new approaches to encoding priors in the artificial genome.

\item \textbf{Curriculum Design:} There is a pressing need to design and validate specific task sequences that effectively mirror evolutionary progression. This includes developing quantitative metrics for measuring progression through evolutionary stages and creating frameworks for automated curriculum generation.

\item \textbf{Industrial Applications:} The principles of evolutionary curriculum could be applied to existing robotic systems and autonomous vehicles, particularly in addressing their current limitations in handling out-of-distribution scenarios. This presents immediate opportunities for industry collaboration and real-world testing.

\item \textbf{Multi-objective Optimization:} There are opportunities to develop new algorithms specifically designed for balancing multiple objectives in an evolutionary framework, building on recent advances in multi-task learning and hierarchical reinforcement learning.

\item \textbf{Benchmarking and Evaluation:} New benchmarks could be developed that specifically test for human-like error patterns and generalization abilities, providing better metrics for evaluating AI systems developed through evolutionary curricula.

\item \textbf{Broader Applications:} Beyond AI safety, the principles of an evolutionary curriculum can be applied to robotics, education, and even therapeutic interventions, creating systems capable of adaptive and context-aware interactions.

\end{itemize}

\hypertarget{sec-infer-the-loss-functions-of-the-brain}{%
\section{Infer the brain's loss functions}\label{sec-infer-the-loss-functions-of-the-brain}}

\hypertarget{subsec-core-idea-loss}{%
\subsection{Core idea}\label{subsec-core-idea-loss}}

AI architectures are trained to optimize loss functions using a learning
rule under a specific data distribution. Richards et al. (2019) \cite{Richards2019-by} suggested that brain algorithms
can be fruitfully thought of in terms of architectures, loss
functions, and learning rules (for an earlier view, see \cite{Marblestone2016-vl}).
While architectures and learning can strongly depend on the
computational substrate, loss functions are an attractive,
implementation-free way of specifying the goals of artificial neural
networks. 

We evaluate three ways of inferring loss functions of biological
systems, at different levels of granularity: first, using the techniques
of task-driven neural nets, we could reverse-engineer objective
functions leading to the representations in the brain, then transfer
them to AI systems. Second, we could infer the reward functions that the
brain implements by leveraging the well-characterized link between the
brain's reward systems and reinforcement learning. Finally, we could
catalog and estimate evolutionarily ancient loss functions, either
imbuing agents with these loss functions or simulating agents which
implement these loss functions. We evaluate each of these ideas in turn.

\hypertarget{subsec-why-does-it-matter-for-ai-safety-and-why-is-neuroscience-relevant-loss}{%
\subsection{Why does it matter for AI safety and why is neuroscience
relevant?}\label{subsec-why-does-it-matter-for-ai-safety-and-why-is-neuroscience-relevant-loss}}

In the framework of \cite{Richards2019-by},

\begin{quote}
Systems neuroscience seeks explanations for how the brain implements a
wide variety of perceptual, cognitive and motor tasks. Conversely,
artificial intelligence attempts to design computational systems based
on the tasks they will have to solve. In artificial neural networks, the
three components specified by design are the objective functions, the
learning rules and the architectures. With the growing success of deep
learning, which utilizes brain-inspired architectures, these three
designed components have increasingly become central to how we model,
engineer and optimize complex artificial learning systems. Here we argue
that a greater focus on these components would also benefit systems
neuroscience. 
\end{quote}

The relevant abstractions are thus:

\begin{enumerate}
\def\labelenumi{\arabic{enumi}.}
\item
  The loss functions (or objective functions) implemented by the system
\item
  The architecture of the system
\item
  The learning rules implemented by the system
\end{enumerate}

To this list, one might add the environment or the stimuli used to train
the system, as well as the specific task that the system must solve.
Loss functions can be transplanted directly and naturally to artificial
systems. It's been argued that this higher-level form of transfer can be
less brittle and lead to better generalization than copying behavior
\cite{Arora2018-bt}.

A different line of research focuses on the
reward functions that the brain implements. If we could infer
what those human reward functions are and what our motivations are, we
could potentially screen them and implement relevant ones in AI systems,
while discarding undesirable behavior: aggression, power-seeking, or
sycophancy \cite{Sharma2023-hi}. This suggests a top-down approach that solves the value
alignment problem by grounding it in neuropsychology
\cite{Sarma2019-ui}.

In developing NeuroAI for AI safety, an essential goal is to align
AI\textquotesingle s loss functions with human values. However,
achieving alignment does not mean replicating human loss
functions exactly. Sarma and Hay \cite{Sarma2016-fx} and Sarma et al. \cite{Sarma2018-td} advocate for a more nuanced 
understanding of human values through neuropsychology. Their proposed 
"mammalian value system" approach posits that human values can be 
decomposed into: mammalian values, human cognition, and the product of human 
social and cultural evolution. Understanding these components may help develop
 AI systems that exhibit human-compatible values without directly replicating 
 human biases or limitations. 

Furthermore, the right values for an AI system may be different than those of humans. 
Direct competition within the same niche---whether biological or
cognitive---can lead to conflict or dominance by one party, as seen in
past evolutionary history \cite{Harari2015-el}. Instead,
it's beneficial for humans and AI to occupy complementary roles in a
cooperative ecosystem. By inferring the nuances of human loss functions
without duplicating them, we can create AI systems that co-exist with
humanity, enhancing resilience and safety
\cite{Collins2024-ry}.

The value of replicating the loss and reward functions of the humans is illustrated by the classic paperclip maximizer thought
experiment \cite{Bostrom2014-gb}. In the original
thought experiment, a superintelligent AI is tasked with optimizing the
manufacture of paper clips. Without any guardrails, the agent proceeds
to break everything down into atoms to transform them into paperclips,
including the humans who originally built the agent. An agent whose
rewards are anchored or derived from the human reward system would
likely have very different outcomes. A prosocial agent would be
motivated to avoid instilling massive suffering on sentient beings. A
safe exploration agent, even if it misunderstood the assignment, would
avoid changing the world drastically, because such large and
irreversible empowerment is intrinsically dangerous. An agent with
homeostatic drives would receive decreasing rewards from turning the
world into paperclips, again likely preventing catastrophe.

The challenge, then, is to systematically infer the loss and reward
functions of the brain from behavior, structure, and neural data. If we
were to correctly identify these, we may be able to obtain safer
exploratory behavior from artificial agents; have effective defenses
against reward-hacking; imbue agents with human-aligned goals; and
obtain robust representations that generalize well out-of-distribution.

\hypertarget{subsec-details-loss}{%
\subsection{Details}\label{subsec-details-loss}}

\hypertarget{subsubsec-task-driven-neural-networks-across-the-brain}{%
\subsubsection{Task-driven neural networks across the
brain}\label{subsubsec-task-driven-neural-networks-across-the-brain}}

Task-driven neural networks
\cite{Yamins2016-ip,Richards2019-by,Doerig2022-lq}--sometimes
called goal-driven neural networks, or neuroconnectionist
approaches--are a commonly used set of tools for comparing
representations in brains and artificial systems. In a typical setup,
artificial neural networks are trained to perform a task that a brain
might accomplish, whether the task is image categorization, a working memory
task, or generating complex movements
\cite{Yamins2014-vm,Khaligh-Razavi2014-gk,Sussillo2015-nl,Zhuang2021-nm,Mineault2021-df,Driscoll2024-am,Nayebi2023-rf}.
Representations are compared using a growing suite of methods--linear
regression, representational similarity analysis, shape metrics, and
dynamic similarity analysis
\cite{Kriegeskorte2008-id,Kornblith2019-dy,Williams2021-uu,Duong2023-ej}.
Frequently, different network architectures, loss functions, and
stimulus sets are considered \cite{Conwell2022-kt}.
Under that framework, networks and loss functions that are better at explaining neural data
can point us toward how the brain leads to behavior,
which may be later verified with causal manipulations
\cite{Walker2019-bx,Bashivan2019-ha,Ponce2019-ll,Cowley2024-dg}.

While task-driven neural networks were originally applied to form vision
in the ventral visual stream, they have now been extended to the dorsal
visual stream, to auditory cortex and language, proprioception,
olfaction, motor cortex, the hippocampus, and even cognitive tasks that
are dependent on association areas and the frontal lobe (see Box \ref{box-taskdrivennn} for
references).

\begin{namedbox}{taskdrivennn}{Task-driven neural networks across cortex}

Task-driven neural networks \cite{Yamins2016-ip,Richards2019-by,Doerig2022-lq} leverage deep learning to better understand representation and computation in natural neural networks. Originally applied to form vision in the ventral visual stream, a large number of other areas and phenomena have now been investigated fruitfully using these methods. We list some of these here.

\begin{itemize}
\item \textbf{Visual System}
    \begin{itemize}
    \item Foundational work showed deep CNNs trained on object recognition develop representations matching neural responses throughout the ventral stream, including IT \cite{Yamins2014-vm, Khaligh-Razavi2014-gk}
    \item Later work validated that this could be used for manipulation of neural activity \cite{Walker2019-bx, Bashivan2019-ha, Ponce2019-ll}
    \item Unsupervised representations can account for responses in high-level visual cortex \cite{Zhuang2021-nm, Conwell2024-df}
    \item Later work demonstrated similar principles in V1 \cite{Cadena2019-xp} and V4 \cite{Cadena2024-vg}, as well as the dorsal visual stream \cite{Rideaux2020-ab, Mineault2021-df, Bakhtiari2021-si, Vafaii2023-lk}
    \item This work has now been extended to more specialized capabilities, including face processing \cite{Chang2021-we}, object context \cite{Bonner2021-vd}, visual attention mechanisms \cite{Lindsay2018-ty, Linsley2018-sj}, predictive processing \cite{Lotter2020-zt} and letter processing \cite{Hannagan2021-mj}
    \end{itemize}

\item \textbf{Auditory Processing}
    \begin{itemize}
    \item Networks trained on speech and audio reveal hierarchical processing in auditory cortex \cite{Kell2018-et}. Recent work has extended this to auditory scene analysis \cite{Francl2022-vb}
    \end{itemize}

\item \textbf{Language}
    \begin{itemize}
    \item Language model representations show striking similarities to cortical responses, including semantics \cite{Huth2012-md, Caucheteux2022-ta}, syntactic processing \cite{Fedorenko2020-dl}, prediction in language areas \cite{Caucheteux2023-hx}, and hierarchical language processing \cite{Schrimpf2021-za}
    \end{itemize}

\item \textbf{Motor control}
    \begin{itemize}
    \item RNNs trained on motor tasks find similar representations to those of the brain \cite{Sussillo2015-nl}
    \item RNNs have emerged as critical in modeling and understanding motor control \cite{Rajan2016-ce, Pandarinath2017-if}
    \item Recent developments have extended this through embodiment \cite{Codol2024-dw, DeWolf2024-vm, Almani2024-yd} and movement planning \cite{Russo2020-iw}
    \end{itemize}

\item \textbf{Olfactory processing}
    \begin{itemize}
    \item Chemical recognition networks develop piriform-like representations for odor encoding \cite{Dasgupta2017-rr}
    \item The evolution of the organization of olfaction can be replicated using task-driven neural network \cite{Wang2021-qn}
    \item Networks trained on masked auto-encoding on chemical structures learn representations similar to humans \cite{Taleb2024-td}
    \end{itemize}

\item \textbf{Hippocampal formation}
    \begin{itemize}
    \item Spatial navigation networks develop representations similar to biological circuits, including grid cells \cite{Banino2018-le, Cueva2018-rw}, place cells \cite{Whittington2020-fj, Nayebi2021-af} and head direction cells \cite{Sorscher2023-tw}
    \item Place cells in the hippocampus compute a predictive map, similar to the Successor Representation used in RL \cite{Stachenfeld2017-ob}, which can be emulated in biologically plausible neural network models \cite{fang2023neural, George2023-sp, bono2023learning}
    \item Mapping sequences to cognitive maps reveals links between transformers and the hippocampus \cite{Whittington2021-qm}
    \item RNN models of planning can explain hippocampal replay \cite{Jensen2024-yi}
    \end{itemize}

\item \textbf{Frontal lobe and executive function}
    \begin{itemize}
    \item Integration and selection can be accounted for by RNNs \cite{Mante2013-gs}
    \item Meta-reinforcement learning captures prefrontal function \cite{Wang2018-zu}
    \item ANNs can capture aspects of task switching \cite{Yang2019-vk} and multi-task learning \cite{Driscoll2024-am}
    \item Mental simulation can be accounted for by predictive world models \cite{Nayebi2023-rf}
    \end{itemize}
\end{itemize}

While task-driven neural networks have proven a critical tool in understanding biological neural networks, important gaps remain. Understanding computation in non-sensory, non-motor regions remains difficult \cite{Summerfield2020-yr}. A fundamental bottleneck is the lack of good artificial models for some tasks (e.g. reasoning, \cite{Fedorenko2024-pf}). Most models only account for one area rather than a range of areas, although multimodal models have become more popular. While deep ANNs are some of the best current models of neural function, it is not unusual that they account for a small proportion of the relevant variance. Capturing neural and individual variability as well as changes throughout development remain a significant long-term challenge.

\end{namedbox}

\hypertarget{subsubsec-feasibility-of-identifying-loss-functions-directly-from-brain-data-using-task-driven-neural-networks}{%
\subsubsection{Feasibility of identifying loss functions directly from brain
data using task-driven neural
networks}\label{subsubsec-feasibility-of-identifying-loss-functions-directly-from-brain-data-using-task-driven-neural-networks}}

Task-driven neural networks have proven highly successful at finding
links between neural representations in biological systems and
artificial neural networks, and in elucidating neural function. We agree
with \cite{Doerig2022-lq} that this
``research programme is highly progressive,
generating new and otherwise unreachable insights into the workings of
the brain''. Here we ask a different question: can task-driven neural networks
identify the right loss functions that the brain implements that can be
transplanted to make safer AI systems? 

A popular practice in the field
is to benchmark several artificial neural networks, often trained for
engineering purposes, in terms of their similarity to brain representations. This has been
used to infer, for example, that deep nets are better aligned to the
brain than classical computer vision models
\cite{Yamins2016-ip}; that self-supervised
representations are just as capable as less biologically plausible
supervised representations at accounting for neural activity in the
ventral stream of visual cortex \cite{Zhuang2021-nm};
and that adversarially robust networks are better aligned to brains than
non-robust networks \cite{Berrios2022-hh,Gaziv2023-kj}.

\begin{figure}[htbp]
    \centering
    \includegraphics[width=\textwidth]{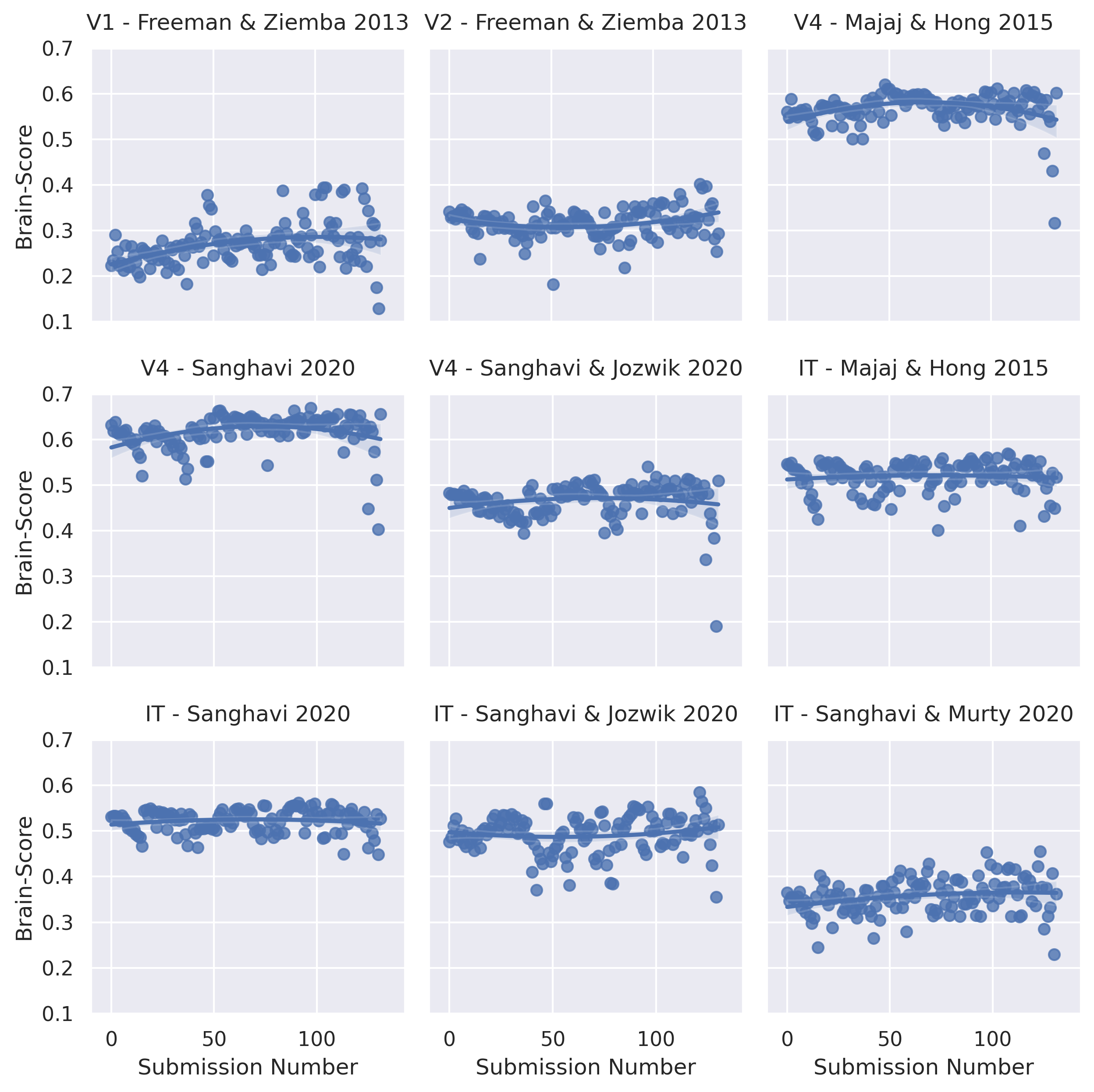}
    \caption{Visual Brain Scores over time for 9 different datasets. While other deep learning benchmarks regularly get saturated within periods of a few years, Brain Scores have flatlined over the last 6 years.}
    \label{fig-brain-scores-time}
\end{figure}

Ideally, benchmark scores should provide a north star
that enables hill-climbing, moving slowly but surely toward the right
architectures, loss functions, and data regimes that align AI systems
with brains. This is one aspiration of BrainScore
\cite{Schrimpf2018-vq}, which seeks to find models that
are brain-aligned by benchmarking multiple models against multiple
datasets, originally for shape vision in the ventral visual stream, and
more recently for language models. It has been noted
\cite{Linsley2023-ce} that supervised image
classifications have become less brain-aligned as they have become
increasingly proficient at image classification. Here we note a different, but conceptually
related finding: visual BrainScores have remained stagnant over the years.
Figure \ref{fig-brain-scores-time} plots visual brain scores for 9 different datasets as a
function of the submission number, a proxy for the submission date of
these models. Unlike many other machine learning benchmarks, where we've
seen rapid progress and eventual saturation, visual BrainScores have remained quite stable over
the past 6 years. These observations
are congruent with the fact that neural alignment is only modestly
improved with scaling \cite{Gokce2024-zd}.

Why is this? We list three possibilities:

\begin{enumerate}
    \item \emph{The models are not diverse enough}. The benchmarked models all lie within a low-dimensional manifold, converging toward similar representations, and there is insufficient diversity to identify directions in model space which would bring models in closer agreement to brain data
\cite{Huh2024-ng,Chen2024-yt}.
    \item \emph{The scores are not very discriminative}. \cite{Conwell2022-kt} use billions of regressions to match vision ANNs to brain data, and find it difficult to infer finer-grained distinctions between loss functions. In the language domain, \cite{Antonello2024-ka} report that a next-token prediction objective and a German-to-English translation objective lead to representations that are equally good at predicting brain responses to podcasts in unilingual English speakers. They report that ``the
high performance of these {[}next token prediction{]} models should not
be construed as positive evidence in support of a theory of predictive
coding {[}...{]} the prediction task which these {[}models{]} attempt to
solve is simply one way out of many to discover useful linguistic
features''.
    \item \emph{The scores are noisy}. Seemingly innocuous changes in the evaluation function to discriminate
these models can change the relative rankings of these results
\cite{Sexton2022-qa,Soni2024-om,Thobani2024-iu,Ahlert2024-xm,Klabunde2024-uq}.
Brain scores are often fit in the interpolation regime, where the number
of latent weights is far larger than the number of observations. In that
regime, scores are a product of the latent dimensionality of the signals
and the match of these latent dimensions to the brain, interacting in often unintuitive ways
\cite{Elmoznino2022-dg,Canatar2023-ya,Schaeffer2024-ew}.
\end{enumerate}

Taken together, these results point toward a lack of identifiability that prevents us
from reverse engineering the objective functions of the brain
\cite{Zednik2016-ml}. With noisy scores lying in a low-dimensional subspace, it is difficult to estimate ``gradients'' in model space: changes in model
parameters, type, or training regimen that would lead to better scores.

How can we improve task-driven neural network methods to better
discriminate among different loss functions? We list three potential
solutions:

\begin{enumerate}
\def\labelenumi{\arabic{enumi}.}
\item
  \emph{Use more highly discriminative benchmarks}. Many of the standard neural benchmarking datasets were collected more than a decade ago,
  for purposes different than benchmarking. Static benchmarks of the
  future should be designed from the ground up to unambiguously identify
  better models on test sets, and contain enough training samples to properly estimate the mapping between ANN and brain representation. Beyond static benchmarks, closed-loop methods
  \cite{Ponce2019-ll,Bashivan2019-ha,Walker2019-bx} and
  discriminative stimuli
  \cite{Golan2020-iv,Tuckute2024-iy} can better differentiate models. Benchmarks
  should cover diverse set of complex real-world cognitive behaviors to
  capture a large slice of relevant brain representations
  \cite{Gao2015-jy,Krakauer2017-ur,Mobbs2018-po,Hall-McMaster2019-an,Pinto2019-fz,Miller2022-sc,Dennis2021-de,Niv2021-fs,Driscoll2024-am}.
\item
  \emph{Explore a larger portion of the design space of models}. Typically,
  benchmarking focuses mostly on pretrained models which have been
  trained for engineering purposes, with a handful of bespoke models
  that supplement those models to test specific hypotheses
  \cite{Zhuang2021-nm,Mineault2021-df}. Sparse coding
  \cite{Olshausen1996-ug}, local Hebbian losses,
  predictive coding \cite{Rao1999-ff}, eigenspectrum
  decay \cite{Agrawal2022-fk}, adversarial training,
  augmentation with noisy and blurry inputs
  \cite{Fort2024-cp}, and other neurally inspired
  objectives are rarely explored in combination with each other. Neural
  architecture search, loss function search, and data-centric AI
  approaches \cite{Northcutt2019-bl} could 
  explore a much larger portion of model space. While these approaches
  can be quite expensive computationally, leveraging scaling laws to
  extrapolate from smaller training runs
  \cite{Kaplan2020-dj,Caballero2022-zo} and Bayesian
  optimization in a closed-loop design
  \cite{Kandasamy2018-ri} could help better explore the
  space.
\item
  \emph{Complement \emph{in silico} benchmarking with physiological experiments}. Recent studies in flies have leveraged task-driven neural
  networks, data-driven modeling, in addition to causal methods to
  estimate the entire cascade of processing stages that lead to specific
  behaviors \cite{Cowley2024-dg,Lappalainen2023-ot}.
  While these studies have been in part enabled by the publication of a
  whole-brain connectome, which remains unavailable in mammals, they
  indicate how one could better infer loss functions using task-driven
  neural networks in mammals in the future.
\end{enumerate}

\hypertarget{subsubsec-forward-and-inverse-reinforcement-learning-to-infer-the-brains-reward-function}{%
\subsubsection{Forward and inverse reinforcement learning to infer the
brain's reward
function}\label{subsubsec-forward-and-inverse-reinforcement-learning-to-infer-the-brains-reward-function}}

One of the most celebrated successes of computational neuroscience has been linking reinforcement learning and the reward
circuits of the brain
\cite{Schultz1997-kr, Hassabis2017-kv,Lindsay2021-pa}. The mesolimbic
reward system--the VTA (part of the brainstem), amygdala, and the basal
ganglia, including the nucleus accumbens (NAc), caudate nucleus, putamen
and substantia nigra (SN)--implement a form of reinforcement learning. A
series of foundational studies have demonstrated that dopaminergic
neurons in these areas encode reward prediction errors, a central
concept in reinforcement learning algorithms
\cite{Schultz1997-kr}. These neurons adjust their firing
rates in response to unexpected rewards or omission of expected
ones, effectively signaling the difference between expected and actual
outcomes \cite{Schultz2016-wd}. Functional neuroimaging studies
in humans have also identified prediction error signals in the
ventral striatum and other reward-related brain regions, supporting the
cross-species applicability of these findings and their importance in understanding disease
\cite{Pessiglione2006-bb, Huys2016-zq}. The story of TD learning in the
brain has been refined over the years, with the
dopamine signaling pathway implicated in model-based
reinforcement learning \cite{Lee2014-kf}, distributional reinforcement learning
\cite{Lowet2020-pc,Muller2024-fw}, and encoding multiple reward dimensions \cite{Lee2024-fg}.

To positively impact AI safety, we'd like to understand the precise mathematical form of the reward function of the brain. The true reward function of the brain differs in important ways from the idealizations often used in linking the brain and RL:

\begin{itemize}
    \item \emph{Rewards are internal}. In the classic framing of reinforcement
learning in the brain, rewards are external--a liquid reward, or a chunk
of food that is experimenter-controlled. Ultimately, however, all rewards are internal
\cite{Barto2009-ht,Keramati2014-wi,Juechems2019-an,Millidge2022-ps,Weber2024-jd}.
    \item \emph{Rewards have multiple components}. Primary reward is split among many drives \cite{Li2020-xg, Liu2023social, allsop2018corticoamygdala, kohl2018functional}, e.g. food, water, sleep, sex, parental care, affection, and social recognition, and can be represented distributionally \cite{dabney2020distributional}.  Byrnes refers to multi-component reward functions as `N-entry scorecards' \cite{byrnes2021dopamine}.
    \item \emph{Rewards are homeostatic}. Many of the intrinsic, primary rewards that humans implement are homeostatic \cite{Keramati2014-wi,Weber2024-jd}: as more reward is harvested, the stimulus becomes less rewarding, even aversive. This is in contradistinction to conventional reinforcement learning, where rewards are a fixed function of the environment.
    \item \emph{Rewards change over a lifetime}. Reward functions change across different life stages and social contexts, so we cannot assume a single static reward function, even for one particular person \cite{Sutton2018-xy}.
\end{itemize}

Primary reward functions interact together in complex ways. Some primary rewards saturate more
strongly than others \cite{Weber2024-jd}: for example,
while water is highly rewarding to a thirsty person, it ceases to be
rewarding upon satiety, which may be related to the fact that we cannot
effectively store water. Other rewards, like those associated with food,
do not reach complete saturation, as we can store fat to satisfy our
later energy demands; the wide availability of hyperpalatable foods,
together with non-saturating reward for food, has been blamed for the
obesity epidemic \cite{Morris2015-zy}.  It is starting to be possible to identify and trace different components of the `N-entry scorecard' of primary rewards by tracing precisely connections between areas in the brain, which has been done in Drosophila \cite{Li2020-xg}, as well as in areas of the mouse brain that regulate social needs \cite{Liu2023social}. 

How secondary rewards--those that are not genetically encoded--are
bootstrapped from primary rewards is an open question. Primary rewards
for food have an interoceptive origin, and include a primary component
that senses food calories and fat content, as well as a secondary,
orofacial reinforcer; this reinforcer could, presumably, be bootstrapped
from taste and smell via temporal difference learning. Similarly, Weber et al.
(2024) \cite{Weber2024-jd} speculate that secondary rewards (e.g. money) are processed by
the same pathways that process non-saturating primary rewards (e.g.
food), leaving open the possibility that these rewards are bootstrapped
using TD-learning. Money can be exchanged for goods and services,
including food, and the rewarding value of food can be transferred to
money over time, much like an unconditioned stimulus can be reinforced
\cite{Ulrich2023-eh,Markman2024-xc}.

\textit{How might the brain learn to compute an abstract secondary reward, like cultural or aesthetic values observed from mentors, and compute this cost function internally?} A well-characterized example of how the brain represents and computes secondary rewards is in songbirds. Songbirds form a memory of
their tutor's song, which they use to precisely evaluate their own song
at a milliseconds timescale while they practice. This evaluation signal
can be observed in dopaminergic neurons, which produce bursts of
activity when the song sounds better than expected, and dips in activity
when it sounds worse than expected
\cite{Gadagkar2016-hw, duffy2022dopamine, roeser2023dopaminergic}. The signal is computed by upstream cortical areas involved in learning the tutor memory, and feeds
into reinforcement learning circuits in the basal ganglia
\cite{Mandelblat-Cerf2014-tl, hisey2018common, kearney2019discrete, roberts2012motor, Mackevicius2018-em}. Other
types of intrinsic rewards in other species, including humans, may be
computed similarly. Specifically, higher-order cortical areas learn to
compute specific features of the world, and may transmit this signal to
deep-brain dopamine neurons, which interface with evolutionarily ancient
reinforcement learning circuits. Thus, throughout evolution, an
increased flexibility and expressivity of cortical representations could
greatly expand the types of reward functions the brain can learn, from
initially only primary rewards, to complex or abstract features learned
from others, such as money or aesthetic preferences.

While reward pathways for primary reinforcers like food are relatively
well-characterized--they are a prime target for treatments for
obesity \cite{Kanoski2016-hh}--other primary reinforcers are poorly understood, including
those related to social instincts, recognition, affection, and parental
care. Reward functions in adults are ultimately derived from the genome,
which bootstraps developmental programs leading to intrinsic reward
circuits, which are then refined through interaction with the environment and higher order shaping of internal circuitry.
Secondary reward functions are highly contingent on the environment but
nevertheless are strongly constrained by the structure of primary
rewards. To understand the precise mathematical form of primary and
secondary rewards--including their variation across the
population--several approaches could be considered:

\begin{enumerate}
\def\labelenumi{\arabic{enumi}.}
\item
  \emph{Decipher the implementation of reward functions through
  structure-to-function mapping}. Building off the successes described above in songbirds at a more granular level would likely require moving to animal models with a full toolkit of brain-mapping and recording technologies, i.e. drosophila and mice. For
  example, to identify candidate sources of secondary rewards,
  connectomics approaches could identify specific cortical inputs to
  dopaminergic areas. To map how the genome impacts reward functions
  would require large-scale mapping of a diverse set of reward-related
  behaviors, coupled with large-scale genetic studies. Capturing 
  sufficient dynamic range in this map will likely require a broad
  distribution of behavioral preferences, likely including non-human
  animal behaviors and comparative genetics and anatomical studies.
\item
  \emph{Estimate rewards by backing them out from reward prediction errors
  measured in the mesolimbic reward circuit}. The TD(0) learning rule
  updates the value function using the reward prediction error, the
  signal that is conventionally assumed to be carried and reported by
  neurons in the mesolimbic system
  \cite{Schultz1997-kr,Sutton2018-xy}. Thus, it may be
  possible to back out the reward function of the brain without the full
  machinery of inverse reinforcement learning, although a full solution
  would need to carefully consider the heterogeneity of signals in the
  mesolimbic system, their dependence on internal state, and their
  non-stationary nature \cite{Millidge2022-ps}.
\item
  \emph{Use inverse reinforcement learning}. This is the classic solution to
  inferring reward from behavior \cite{Ng2000-sr}.
  Several studies have attempted to back out reward functions from
  animal and human behavior and compare them to brain activity
  \cite{Rothkopf2013-xe,Yamaguchi2018-rv,Jara-Ettinger2019-ck,Tan2022-pm,Kalweit2022-fr,Ashwood2022-br,Gong2024-dh,Lee2024-ko}.
  In general, these studies have not leveraged reward prediction error
  signals in the mesolimbic reward pathway. Rather, they infer reward
  functions using classic IRL (e.g. using maximum entropy inverse
  reinforcement learning) and then compare the retrieved reward
  functions with other aspects of an animal's behavior, or their neural
  activity. The combination of IRL and conditioning on brain data could
  help alleviate the classical issues with IRL, including the
  ill-posedness of its inference, and its reliance on assumptions about
  agent rationality. See \cite{Urbaniak2024-hg} for a
  framework that links RL models of cognitive behavior to biologically
  plausible models of neural activity. Classic IRL can face feasibility challenges with more complex reward functions, which can be addressed with more structured, hierarchical, symbolic representations of reward functions \cite{davidson2024goals}. 
\end{enumerate}

While there is a rich body of research on reinforcement learning in the
brain, there is surprisingly sparse literature on how intrinsic rewards
are coded, with songbirds as a notable exception, as discussed above.
Much of the research studying reward processing non-invasively in humans
focuses on isolating single mechanisms, for example using the Iowa
gambling task, the monetary incentive delay task, 2 armed bandits, etc.,
and is thus ill-suited to derive intrinsic rewards using
brain-conditioned reinforcement learning or inverse reinforcement
learning. We believe there is fruitful research to be done by studying
decision-making with a high loading on intrinsic rewards under
naturalistic scenarios to infer the brain's true reward functions.

\hypertarget{subsubsec-brain-like-agi-safety}{%
\subsubsection{Brain-like AGI safety}\label{subsubsec-brain-like-agi-safety}}

In a series of blog posts \cite{Byrnes2022-wm}, AI
safety researcher Steven Byrnes proposed a pathway toward brain-like AGI
safety that partially overlaps with the goals of inferring goals from
brains, as well as evolutionary approaches to AI safety. While we cannot
fully do justice to their entire proposal here, we outline it for the
benefit of neuroscientists who might resonate with the framework but are
not exposed to the AI safety literature.

Broadly speaking, the proposal posits the existence of two systems:

\begin{enumerate}
\def\labelenumi{\arabic{enumi}.}
\item
  A learning subsystem. Includes the neocortex, as well as parts of the
  striatum, amygdala, and cerebellum. As its name indicates, the
  learning subsystem's representations are learned within an organism's
  lifetime, from scratch. Byrnes roughly estimates the proportion of the
  brain dedicated to ``from-scratch'' learning as 96\%. The learning
  subsystem is the seat of the N-entry scorecard, a term-of-art for a
  short-term predictor of reward that could be identified with the
  mesolimbic reward system, especially secondary rewards or multiple
  reward components. It is also responsible for building a disentangled
  world model.
\item
  A steering subsystem. This includes large parts of the hypothalamus,
  but also potentially the globus pallidus. The representations in this
  system are posited to be hardcoded in the genome from birth, although
  there is some space for low-dimensional, within-lifetime calibration.
  The steering subsystem's name echoes evolutionary neuroscience,
  wherein the common ancestor of modern animals, the worm-like
  urbilaterian, had a basic nervous system with the capacity to steer
  toward food and away from enemies, along with a basic memory and
  motivation system to maintain homeostasis
  \cite{Cisek2019-mh,Cisek2022-qc,Bennett2023-rq,Karin2022-qf}.
  In the framework of Byrnes, the steering subsystem sends supervisory
  systems and rewards to the learning subsystem.
\end{enumerate}

A core claim of the brain-like AGI safety framework is that
understanding the learning subsystem in adults is the wrong level of
abstraction: in adults, structure, and function are far too complex to
fathom directly. Rather, we should focus on the training data used
within an animal's lifetime, as well as the underlying learning
algorithms that led to these representations. This resonates with the
framework of Richards et al. (2019) \cite{Richards2019-by}.

By contrast, Byrnes posits that we should focus on the function of the
steering subsystem. ``All human goals and motivations come ultimately
from relatively simple, genetically hardcoded circuits in the Steering
Subsystem (hypothalamus and brainstem)''. We should ``reverse-engineer
some of the ``innate drives'' in the human Steering Subsystem
(hypothalamus \& brainstem), particularly the ones that underlie human
social and moral intuitions {[}...{]}, make some edits, and then install
those ``innate drives'' into our AGIs.''. Some of the desirable innate
drives include those underlying altruism, cooperation, and a desire to
align with human values. This is consistent with the framing in the
previous section. Indeed, Byrnes places a high priority on understanding
the reward function of the brain, and in particular, how social
behaviors are bootstrapped from instincts instantiated in the steering
subsystem; in other words, how primary rewards and built-in circuits are
bootstrapped to create prosocial behavior.

Because this proposal recapitulates and overlaps substantially with
other proposals considered in this roadmap
\cite{Sarma2019-ui,Richards2019-by}, we do not evaluate
it separately; rather, we see it as a valuable synthesis and accessible
introduction to neuroscience ideas for an AI safety audience.

\hypertarget{subsec-evaluation-loss}{%
\subsection{Evaluation}\label{subsec-evaluation-loss}}

Inferring the loss functions of the brain presents a promising yet
challenging avenue for enhancing AI safety. The idea leverages the
potential alignment between human cognitive processes and AI objectives
to mitigate risks like reward hacking and misaligned behaviors. However,
several significant hurdles exist:

\begin{itemize}
\item
  Identifiability issues in inferring loss functions from the brain. The
  complexity of the brain\textquotesingle s loss functions and the
  challenge of uniquely identifying them based on observed data limit
  the practicality of this approach. Current tools like task-driven
  neural networks may lack the sensitivity and specificity needed to
  pinpoint exact loss functions, as evidenced by stagnant brain scores.
\item
  Complexity of human reward functions. Human reward functions are
  likely multifaceted, context-dependent, homeostatic, non-stationary,
  and influenced by evolutionary factors, making them difficult to model
  accurately.
\item
  Lack of scalability of Inverse Reinforcement Learning (IRL). IRL,
  one of the most promising approaches to infer reward functions from
  the brain, is limited to small state spaces under assumptions that
  likely don't hold. Brain-conditioned IRL is underexplored for
  inferring reward functions for the brain.
\item
  Limited relevant data on intrinsic rewards, and how they're used to
  bootstrap secondary rewards. In cases where this data exists, such as songbirds and drosophila and certain rodent behaviors, it has been highly valuable. 
\item
  Limited tools for measuring structure-to-function maps to understand
  how genetic circuits drive development through wiring rules that lead
  to innate reward functions. Interestingly, recent BRAIN cell atlases show that most of the genetic cell type diversity is in subcortical (likely `steering' regions that represent loss functions in the brain) \cite{yao2023high}. 
\end{itemize}

Despite these challenges, ongoing advances in neuroscience and
AI offer potential pathways to overcome some obstacles. Improved
benchmarking methods, exploration of a wider space of models and
behaviors, and integrative approaches combining behavioral and neural
data could make it more feasible to infer loss functions.

\hypertarget{subsec-opportunities-loss}{%
\subsection{Opportunities}\label{subsec-opportunities-loss}}

\begin{itemize}
\item
  Develop discriminative benchmarks for task-driven neural networks.
  Create new datasets and evaluation methods designed specifically to
  differentiate between models based on their alignment with brain data,
  across a broad range of behavioral contexts, enhancing the ability to
  infer loss functions.
\item
  Conduct closed-loop experiments. Perform closed-loop neural
  experiments to test specific hypotheses about the
  brain\textquotesingle s loss functions, allowing for more precise
  validation of models.
\item
  Explore a wider model space. Investigate a broader range of
  architectures, loss functions, and learning rules, including those
  inspired by biological processes like sparse coding and predictive
  coding.
\item
  Integrate multimodal data from a broad range of behavioral contexts.
  Combine behavioral, neural, and structural data to improve the
  inference of reward functions, leveraging the strengths of each data
  type.
\item
  Advance IRL. Refine IRL techniques to better handle the complexities
  of human behavior and integrate neural signals related to reward
  processing, potentially overcoming identifiability issues.
\item
  Focus on evolutionarily ancient drives. Study and model the
  fundamental drives shared across species to inform the design of AI
  systems with more universally aligned objectives through comparative
  neuroanatomy across different mammal species. 
\end{itemize}

\hypertarget{sec-leverage-neuroscience-inspired-methods-for-mechanistic-interpretability}{%
\section{Use neuroscience methods for interpretability}\label{sec-leverage-neuroscience-inspired-methods-for-mechanistic-interpretability}}

\hypertarget{subsec-core-idea-mechint}{%
\subsection{Core idea}\label{subsec-core-idea-mechint}}

Over the past several decades, neuroscientists have built a suite of tools to understand single neuron and population computation
\cite{Barack2021-br}. Neuroscientists trying to
understand the brain face many of the same problems as AI researchers
when opening the black box that is an artificial neural network: both
deal with a complex system, designed for performance rather than
transparency, containing millions or billions of distinct subunits, that
iteratively reformats its inputs in baffling ways to create emergent
adaptive behavior.

Many of the methods of neuroscience are highly relevant to
interpretability research, which seeks to understand and control
artificial neural networks at the level of weights, neurons,
subnetworks, and representations \cite{Rauker2023-pn}.
Mechanistic interpretability (MechInt) seeks to build
human-understandable, bottom-up, circuit-based explanations of deep
neural networks, often by examining weights and activations of neural
networks to identify circuits that implement particular behaviors
\cite{Bereska2024-ry,Olah2020-yy,Elhage2021-vv}.
Representation engineering \cite{Zou2023-qo} seeks to
understand neural networks from the top-down at the representation level, and
control these representations to make networks safer, less toxic, and
more truthful. Both of these approaches have been inspired in part by
neuroscience. Transparency and control are important components of safe
AI systems, and there are still many more ideas in neuroscience that
await fruitful deployment in AI safety.

\hypertarget{subsec-why-does-it-matter-for-ai-safety-and-why-is-neuroscience-relevant-mechint}{%
\subsection{Why does it matter for AI safety and why is neuroscience
relevant?}\label{subsec-why-does-it-matter-for-ai-safety-and-why-is-neuroscience-relevant-mechint}}

Take the point of view of mechanistic interpretability
\cite{Bereska2024-ry}:

\begin{quote}
Understanding AI systems' inner workings is critical for ensuring value
alignment and safety. {[}\ldots{]} Mechanistic interpretability {[}seeks
to{]} reverse-engineer the computational mechanisms and representations
learned by neural networks into human-understandable algorithms and
concepts to provide a granular, causal understanding.
\end{quote}

The exact theories of impact by which interpretability could improve AI
safety have been enumerated by researchers in the field
\cite{Nanda2022-ty}, which include:

\begin{enumerate}
\def\labelenumi{\arabic{enumi}.}
\item
  \emph{Auditing}: models can be checked for safety characteristics using the
  methods of mechanistic interpretability. Traditional behavior checks
  and red-teaming can only enumerate problems that are thought of by
  human checkers. By contrast, leveraging knowledge about a model's
  internals has the potential to uncover undesirable outcomes that would
  otherwise be hard to elicit, e.g. textual adversarial attacks in
  vision-language models \cite{Goh2021-tx}. It also
  extends to instances where we need to track down the source of
  undesirable behavior, e.g. telling apart lack of knowledge, parroting
  of incorrect information, or deception
  \cite{Burns2024-co,Zou2023-qo}.
\item
  \emph{Correction}: by understanding mechanistically why models fail to be
  aligned, we can correct them, e.g. by steering the system away from
  unintended behavior \cite{Turner2024-ww}, ablating
  components of the system that are causing the issue
  \cite{Bau2018-dl}, or patching/engineering the system
  so it doesn't display the undesirable behavior
  \cite{Meng2023-ub}.
  
\end{enumerate}

Reverse-engineering the black box that is the brain at a causal level is
precisely what much of neuroscience tries to accomplish
\cite{Marr1982-xh,Thompson2021-jc,Jonas2017-xb,Lillicrap2019-ha}.
Insofar as neuroscientists have had a multi-decade head start in trying
to break down complex black-box neural networks, their methods are
relevant to reverse-engineer computational mechanisms \emph{in silico},
and thus positively affect AI safety.

\begin{namedbox}{anns-vs-bnns}{Similarities and differences between ANN and BNN interpretability}

AI researchers have noted similarities and differences between the goals
of neuroscience and artificial interpretability research
\cite{Olah2021-su,Thompson2021-jc,Kar2022-wx,Lindsay2023-zr}.
Not every tool in the neuroscience toolbox is relevant to artificial
interpretability research, and vice-versa. Some of the main ways in
which artificial and biological neural networks differ from an
interpretability perspective include:

\begin{itemize}
\item
  Access to every weight, activation, full model architecture, and
  response to arbitrary stimuli: AI researchers have access to far more
  information than is typically available in neuroscience experiments.
  It has only recently become feasible to access the complete connectome
  for a larger animal than \textit{C. elegans}
  \cite{Shan_Xu2020-nc,Scheffer2020-ck}. Access to
  almost complete brain activity of an organism at the single neuron
  level is currently limited to \textit{C. elegans}
  \cite{Nguyen2016-ep} and zebrafish
  \cite{Ahrens2013-vv}, while whole-brain recording at a
  coarser level than single neurons is feasible in Drosophila
  \cite{Aimon2018-hn}. Coverage in mammals is improving
  \cite{Stevenson2011-sr,Urai2022-eh}, with a recent
  report of up to a million neurons simultaneously imaged in mice
  \cite{Manley2024-xn}.
\item
  Expressive editing tools: AI researchers have access to a number of
  advanced causal interventions that neuroscientists could only dream
  of: replacing the latent activity of one token for another via
  activation patching \cite{Meng2023-ub}, steering
  neural activity toward desirable outcomes
  \cite{Turner2024-ww}, or deleting or swapping layers
  of trained models \cite{Lad2024-jp}. By contrast,
  patterned optogenetics
  \cite{Shin2023-qm,Adesnik2021-xr} and multi-electrode
  stimulation have seen limited adoption, and we
  only have very limited means of reshaping the activity of a neural
  circuit.
\item
  Noise and spikes: cortical neurons are noisy
  \cite{Shadlen1998-ti}. Inferring latent rates from
  single trials generally require powerful denoising models
  \cite{Pandarinath2018-gl,Ye2021-ek}. This creates a
  different set of challenges than in AI, where full, noise-free
  activity--including activity prior to nonlinearities--is available.
  Many interpretability methods in neuroscience involve a fair amount of
  consideration about sample efficiency to overcome noise issues
  \cite{Watanakeesuntorn2020-pj}, which don't exist in
  AI interpretability.
\item
  Recurrence vs. feed-forward computing: Many neural systems are
  recurrent, which facilitates their analysis as dynamical systems
  \cite{Vyas2020-qx}. By contrast, in AI, while RNNs can
  be fruitfully analyzed through a dynamical systems lens
  \cite{Maheswaranathan2019-hy}, they have fallen out of
  fashion for transformers \cite{Vaswani2017-xw}, which
  don't have a natural state-space interpretation
  \cite{Dutta2021-ca}. Resurging interest in state-space
  models \cite{Beck2024-ja,Dao2024-gj} may make these
  methods relevant again in the near future.
\item
  Capacity and modularity: the human neocortex contains about 100
  billion neurons and 100 trillion synaptic connections. The number of
  distinct computational elements and weights in the largest-scale
  transformers \cite{Dubey2024-ee} is at least two
  orders of magnitude lower than that of brains, despite being exposed
  to far more information than a human would in their lifetime (e.g. all
  of Wikipedia, Google Books, arXiv, and a significant proportion of the
  internet). Empirical and theoretical work on the superposition
  hypothesis \cite{Elhage2022-fp} and earlier work on
  sparse coding \cite{Barak2013-de} points to a more
  tangled, dense representation in artificial neural networks than is found in
  the brain, which might hinder interpretation. Modularity and hierarchy
  in brains \cite{Meunier2010-ql} and topographical
  within-area organization might facilitate interpretation, as similarly
  tuned neurons are spatially clustered
  \cite{Bonhoeffer1991-gs,Tanaka2003-hq,Paik2011-xg,Ringach2016-xl,Willeke2023-ae}.
  Note, however, that modularity may occur in otherwise generic ANNs composed of
  equivalent units via symmetry breaking
  \cite{Krizhevsky2012-uz,Voss2021-hg,Bakhtiari2021-si,Achterberg2022-nc,Kumar2024-bn}.

\item
  Evolution and development: As Dobzhansky famously noted \cite{Dobzhansky1973-yw},
  \textquotesingle nothing in biology makes sense except in light of
  evolution.\textquotesingle{} Natural neural networks emerge through
  developmental processes guided by genomic bottlenecks \cite{Zador2019-gf},
  which constrain their architecture and function. The resulting
  principles are highly conserved through evolutionary lineages, making
  phylogenetic and cross-species analyses particularly enlightening for
  understanding neural organization \cite{Cisek2019-mh, Bennett2023-rq}. While
  the evolution of deep learning architectures has been compared to
  natural selection \cite{Kaznatcheev2022-cp}, and some work has
  explored self-organizing and evolving architectures \cite{Mordvintsev2020-ri, Stanley2019-iu}, fundamental differences persist. Unlike
  biological systems, conventional artificial neural networks lack true inheritance
  mechanisms and developmental constraints. Consequently, traditional
  tools from evolutionary neuroscience, such as comparative analysis and
  developmental trajectories, would need to be substantially
  reformulated to apply to artificial architectures.
\end{itemize}
\end{namedbox}

\hypertarget{subsec-details-mechint}{%
\subsection{Details}\label{subsec-details-mechint}}

Neuroscientists have long been interested in understanding how neural
circuits compute. Yet neurons are noisy and it is difficult to know how
they operate \emph{in situ}, due to a combination of sparse sampling,
noise, limited recording time, and complexity. Over the years,
scientists have built a set of tools to analyze neuron computation both
at the single neuron (aka Sherringtonian) level and at the population
(aka Hopfieldian) level \cite{Barack2021-br}.

\hypertarget{subsubsec-single-neuron-sherringtonian-view}{%
\subsubsection{Single-neuron (Sherringtonian)
view}\label{subsubsec-single-neuron-sherringtonian-view}}

Many methods have been developed to characterize the responses of
sensory neurons. These methods attempt to characterize neurons'
selectivity, preferred and non-preferred stimuli, as well as transfer
functions. These methods include:

\begin{itemize}
\item
  Characterizing the receptive fields of neurons
  \cite{Sherrington1906-ux}

  \begin{itemize}
  \item
    Characterizing areas driving excitation and inhibition through a
    qualitative, manual analysis
    \cite{Hartline1940-ti,Hubel1959-zo}\item
    Characterizing neurons' transfer function using noise stimuli in
    terms of a Wiener-Volterra expansion. First-order methods (i.e.
    including only a linear term in the expansion) include
    spike-triggered averaging
    \cite{De_Boer1968-eo,Marmarelis1978-dh,Ringach2004-ay}.
    Second-order methods include spike-triggered covariance
    \cite{Schwartz2006-du} and second-order forms
    \cite{Berkes2007-lj}.\item
    Characterizing a neuron's transfer function to naturalistic stimuli
    using parametric methods, such as Linear-Nonlinear-Poisson models
    \cite{Paninski2004-yv,Wu2006-rd},
    nonlinear-input-models (NIM) methods comprised of stacks of additive
    layers \cite{McFarland2013-ui,Mineault2012-hu}, or
    deep learning models
    \cite{Wu2015-rg,McIntosh2016-zh,Cadena2019-ay,Lurz2020-cs,Cobos2022-lo,Willeke2023-ae, franke2022state}
\item
    Attributing visual decisions to particular parts of a stimulus
    through methods involving random masking, such as Bubbles
    \cite{Gosselin2001-wp}
  \end{itemize}
\item
  Characterizing the tuning curves of neurons using parameterized
  stimuli
  \cite{Hubel1968-uo,Pasupathy2002-wv,Butts2006-zo}\item
  Using active stimulus design to characterize a system

  \begin{itemize}
  \item
    Finding preferred, anti-preferred, or diverse stimuli using
    inception loops
    \cite{Walker2019-bx,Bashivan2019-ha,Ponce2019-ll}\item
    Characterizing iso-response curves
    \cite{DiMattina2013-xt}\item
    Finding stimuli that maximize the delta between predictions of
    different models to facilitate model comparisons
    \cite{Golan2020-iv,Feather2023-pm}
  \end{itemize}
\item
  Relating an organism's behavior in decision-making tasks to the
  responses of single neurons, for example through neurometric curves
  \cite{Gold2007-lx}
\end{itemize}

These methods aim to characterize the input-output functions of single
neurons through a rich set of tools. Although these methods can
exhaustively characterize the response properties of neurons, they tend
not to directly address the relationship of single neurons to perception
and behavior \cite{Cohen2009-lq}. Causal manipulation
tools, including micro-stimulation, optogenetics, and thermogenetic
tools can all be used to activate or deactivate a neuron or a set of
neurons in situ, thus establishing a causal link between single neurons
and behavior \cite{Bernstein2012-tv,Emiliani2022-jl}.

\hypertarget{subsubsec-population-level-hopfieldian-view}{%
\subsubsection{Population-level (Hopfieldian
view)}\label{subsubsec-population-level-hopfieldian-view}}

In contrast to the single-neuron view, the Hopfieldian view
\cite{Barack2021-br} seeks to understand neural
representation from the perspective of how multiple neurons together
coordinate to create and transform a representation. The population
level view has been popularized by the rise of large-scale recordings of
single neurons--Utah arrays, multi-electrode arrays, flexible probes,
V-probes, Neuropixels, and multi-photon calcium imaging
\cite{Urai2022-eh}. This shift in perspective was first
popularized in the study of the motor cortex
\cite{Churchland2012-cp,Vyas2020-qx}. The view that
populations of neurons should be the primary object of interest in
neuroscience has been referred to as the \emph{neural population doctrine}
\cite{Saxena2019-xw,Ebitz2021-zy} or \emph{neural manifolds}
\cite{Gallego2017-fa,Gao2017-ze,Chung2021-jb}. In this
view, the brain's state is viewed as evolving over time in a phase space
with a lower dimensionality than the ambient dimension
\cite{Jazayeri2021-rk}, tracing out manifolds; neurons
are viewed as (linear) projections of the phase space.

In parallel, a large body of work in noninvasive human measurements, in
particular in functional magnetic resonance imaging (fMRI), has focused
on the geometry of representations
\cite{Norman2006-cd,Kriegeskorte2008-id}. This is partly
due to methodological concerns: fMRI and other noninvasive modalities
average over the responses of hundreds of thousands of neurons, making
single-neuron analysis impractical. However, some aspects of the geometry of the
representations, and the distance between stimuli in voxel space, are
preserved by random projections by the Johnson-Lindenstrauss lemma
\cite{Kriegeskorte2016-qh}. This allows fruitful
investigation of the geometry of representations despite the
methodological limitations caused by measuring neural activity
indirectly.

Population-level analyses are an integral part of the toolkit of the
neuroscientist seeking to understand the brain. These methods include:

\begin{itemize}
\item
  \textbf{Methods that seek to find sparse representations of neural
  and/or artificial data.} These are often, strictly speaking, machine
  learning methods, but they've often been motivated from a neuroscience
  perspective, either from first principles or from a use-case
  perspective:

  \begin{itemize}
  \item
    ICA \cite{Bell1994-gq}\item
    Sparse coding \cite{Olshausen1996-ug}\item
    Nonnegative Matrix Factorization \cite{Lee1999-jk}\item
    Tensor component analysis and their variants
    \cite{Williams2018-hq,Pellegrino2024-tg}
  \end{itemize}
\item
  \textbf{Methods that seek to perform decoding of neural activity}

  \begin{itemize}
  \item
    Multivariate pattern analysis \cite{Haxby2012-gc}\item
    Encoding and decoding models \cite{Naselaris2011-wu}
  \end{itemize}
\item
  \textbf{Methods that compare representations}

  \begin{itemize}
  \item
    Representational Similarity Analysis
    \cite{Kriegeskorte2008-id}\item
    Canonical Correlation Analysis \cite{Wang2020-gy}\item
    Neural shape metrics
    \cite{Williams2021-uu,Duong2023-ej}
  \end{itemize}
\item
  \textbf{Methods adapted from dynamical systems analysis}, including
  phase-space analysis and the analysis of attractors and bifurcations
  \cite{Sussillo2014-hr,Strogatz1994-hw}

  \begin{itemize}
  \item
    Estimating latent dimensions from noisy neural data
    \cite{Hurwitz2021-ml}\item
    Methods based on delay embeddings, e.g. Empirical Dynamic Modelling
    \cite{Watanakeesuntorn2020-pj}\item
    Dynamical Similarity Analysis
    \cite{Ostrow2024-dt}
  \end{itemize}
\end{itemize}

\hypertarget{subsubsec-alternative-views-on-neural-computation}{%
\subsubsection{Alternative views on neural
computation}\label{subsubsec-alternative-views-on-neural-computation}}

Intermediate between the single-neuron view and population view is that
of axis-aligned coding. In axis-aligned coding, single neurons have
non-accidental selectivity, but population activity nevertheless has a
central role \cite{Khosla2024-xa,Prince2024-do}. A
number of metrics have been proposed to characterize axis-aligned coding
\cite{Williams2021-uu}, although no consensus metric has
thus far emerged.

An additional line of research has focused on characterizing the
connections between neurons as opposed to their activity
\cite{Sporns2005-ge,Seung2012-va,Bassett2017-or,Barabasi2023-hv}.
This edge-centric viewpoint has been popularized in the subfield of
connectomics. Connectomics has made remarkable progress in explaining
neural circuits, including in particular in the fly, where reference
connectomes have been published, first of the fly hemibrain
\cite{Shan_Xu2020-nc,Scheffer2020-ck}, and more
recently of the entire brain of larvae and adults
\cite{Dorkenwald2023-xt,Winding2023-re}.
Connectomics-centric approaches have been used to explain the mushroom
body \cite{Li2020-xg}, ring attractors for navigation
\cite{Hulse2021-wq}, functional properties of the visual
system
\cite{Shinomiya2022-pe,Lappalainen2023-ot,Pospisil2024-nq},
and sensorimotor transformations \cite{Shiu2024-sq}.

\newpage
\setlength\LTleft{-.5in}
\setlength\LTright{-.1in}
\begin{longtable}{p{0.2\textwidth}p{0.4\textwidth}p{0.4\textwidth}}
\toprule
\textbf{Method} & \textbf{Neuroscience} & \textbf{Mechanistic Interpretability} \\
\midrule
\endfirsthead

\multicolumn{3}{l}{\small\textit{Continued from previous page}} \\
\toprule
\textbf{Method} & \textbf{Neuroscience} & \textbf{Mechanistic Interpretability} \\
\midrule
\endhead

\midrule
\multicolumn{3}{r}{Continues on next page} \\
\endfoot

\endlastfoot

Tuning Curves & Characterizing neuron responses to parameterized stimuli \cite{Hubel1968-uo, Pasupathy2002-wv, Butts2006-zo} & Probing artificial neurons with parameterized inputs \cite{Cammarata2021-gt} \\

\hline

Receptive Fields / Preferred Stimuli & 
Manual characterization of excitation and inhibition areas \cite{Hartline1940-ti, Hubel1959-zo}; Finding preferred stimuli using inception loops \cite{Walker2019-bx, Bashivan2019-ha, Ponce2019-ll}; Systems identification methods including:
  \begin{itemize}[nosep]
  \item Spike-triggered analysis \cite{De_Boer1968-eo, Marmarelis1978-dh, Ringach2004-ay, Schwartz2006-du}
  \item Linear-Nonlinear-Poisson models \cite{Paninski2004-yv, Wu2006-rd}
  \item Nonlinear input models \cite{McFarland2013-ui, Mineault2012-hu}
  \item Deep learning models \cite{Wu2015-rg, McIntosh2016-zh, Cadena2019-ay, Lurz2020-cs}
  \end{itemize}
 & 
Calculating maximizing stimuli for artificial neurons \cite{Zeiler2013-hz,Cammarata2020-ld}\\

\hline

Causal Manipulations & Micro-stimulation, optogenetics, and thermogenetic tools for neural manipulation \cite{Bernstein2012-tv, Emiliani2022-jl} & Ablations \cite{Wang2022-bp}; Activation patching \cite{Heimersheim2024-qt}; Steering \cite{Templeton2024-br, Zou2023-qo}; Causal mediation analysis \cite{Mueller2024-it} \\

\hline

Circuit Analysis & 
Connectomics studies \cite{Shan_Xu2020-nc, Scheffer2020-ck, Dorkenwald2023-xt, Winding2023-re}; Functional circuit analysis \cite{Li2020-xg, Hulse2021-wq, Shinomiya2022-pe, Lappalainen2023-ot, Pospisil2024-nq, Shiu2024-sq}
 & 
Identifying circuits across layers in neural networks \cite{Cammarata2020-ld, Elhage2021-vv} \\

\hline

Population-level Analysis & Analyzing coordinated activity of multiple neurons \cite{Saxena2019-xw, Ebitz2021-zy}; Neural manifolds \cite{Gallego2017-fa, Gao2017-ze, Chung2021-jb} & Examining distributed representations in neural networks \cite{Zou2023-qo} \\

\hline

Dimensionality Reduction & PCA and ICA \cite{Bell1994-gq}; NMF \cite{Lee1999-jk}; Tensor component analysis \cite{Williams2018-hq, Pellegrino2024-tg} & Similar techniques applied to artificial neural network activations \\

\hline

Sparse Representations & 
Sparse coding analysis \cite{Olshausen1996-ug}; Tensor component analysis \cite{Williams2018-hq, Pellegrino2024-tg}
 & 
Sparse autoencoders for interpreting neural network activations \cite{Elhage2022-fp, Cunningham2023-yb, Gao2024-do} \\

\hline

Representation Comparison & 
Representational Similarity Analysis \cite{Kriegeskorte2008-id}; Canonical Correlation Analysis \cite{Wang2020-gy}; Neural shape metrics \cite{Williams2021-uu, Duong2023-ej}
& 
Centered Kernel Alignment (CKA) \cite{Kornblith2019-dy} \\

\hline

Decoding & 
Multivariate pattern analysis \cite{Haxby2012-gc}; Encoding and decoding models \cite{Naselaris2011-wu}
 & 
Using classifier probes to understand representations in artificial networks \cite{Alain2016-cv} \\

\hline

Dynamical Systems Analysis & 
Analysis of attractors and bifurcations \cite{Sussillo2014-hr, Strogatz1994-hw}; Empirical Dynamic Modelling \cite{Watanakeesuntorn2020-pj} & 
Applying similar concepts to understand the dynamics of artificial neural networks \cite{Maheswaranathan2019-hy} \\

\hline

\caption{Parallel analysis methods in neuroscience and mechanistic interpretability.}
\label{tab-parallel-methods}
\end{longtable}

\hypertarget{subsubsec-mechanistic-interpretability}{%
\subsubsection{Mechanistic
interpretability}\label{subsubsec-mechanistic-interpretability}}

Interpreting the inner workings of artificial neural networks has a long
history. Early ANNs were often hand-designed, in a bottom-up fashion,
and the relationship between weights, architectures and capabilities was
clear \cite{Rumelhart1987-cp}. During the first wave of
neural networks trained through backpropagation, it became common
practice to visualize the weights of neural networks (i.e. Hinton
diagrams). During the early deep learning resurgence, it was common to
visualize the weights of networks trained on visual tasks or to
visualize samples from their learned generative models
\cite{Hinton2006-pn,Lee2009-ro,Krizhevsky2012-uz}.
Later, methods were devised to attribute decisions to particular
elements in inputs such as images or text
\cite{Barredo_Arrieta2020-xi}. These methods can
potentially give insights into what drives a decision and identify when
algorithms use non-robust features to perform a task--shortcut learning
\cite{Geirhos2020-uk}.

Mechanistic interpretability \cite{Bereska2024-ry}
aspires to reverse-engineer the algorithms implemented by artificial
neural networks \cite{Olah2020-yy,Elhage2021-vv}. This
approach has a broader scope than earlier methods of interpretability,
in that it seeks to build mechanistic (e.g. pseudocode) breakdowns of
computations in artificial neural networks by the bottom-up analysis of
neurons, connections, and activations. This is often complemented by an
analysis of the mathematics of neural networks
\cite{Elhage2021-vv}. The analogy between mechanistic
interpretability--or MechInt--and neuroscience is not lost on the
field; leaders in the field of mechanistic interpretability have
sometimes referred to it as ``neuroscience of artificial neural
networks'' \cite{Lindsay2023-zr,Kar2022-wx}.

Early work on mechanistic interpretability on image classification
models \cite{Olah2020-yy} cut across many of the
concepts and tools from decades of visual neuroscience, and no doubt
would be familiar to Hubel \& Wiesel. This included:

\begin{itemize}
\item
  Calculating tuning curves to parametrized stimuli
\item
  Identifying maximizing stimuli (i.e. receptive fields)
\item
  Identifying circuits across layers (across visual areas, respectively)
  to implement specific functions
\end{itemize}

An example of this line of work is the identification of a circuit for
the detection of curves in images in a large-scale convolutional neural
network trained on image classification. A combination of estimating
preferred stimuli using gradient descent
\cite{Simonyan2013-pa,Olah2018-er} and visualizing
natural image patches causing high activations
\cite{Zimmermann2024-mm} identified a subset of neurons
that were putatively selective for curves. These neurons were then
probed using parameterized stimuli to confirm their specific tuning to
orientation and curvature, thus forming tuning curves. This identified a
family of curve detectors in a layer of the network, selective for
similar curvature, but at different orientations. A bottom-up circuit
with hand-tuned, programmatically-generated weights was then built,
recapitulating the main findings from the backpropagation-trained model
\cite{Cammarata2021-gt}. This work could be seen to
recapitulate long-standing work in neuroscience mapping the selectivity
of contours in primate areas V1, V2, and V4
\cite{Hubel1959-zo,Freeman2013-jb,Pasupathy2002-wv}.

This approach toward mechanistic interpretability has since been adapted
to the analysis of large language models
\cite{Rauker2023-pn}. Specialized circuits implementing
important atomic functions have been identified in LLMs, such as
induction heads \cite{Olsson2022-bb}, indirect object
identification \cite{Wang2022-bp}, factual recall
\cite{Geva2023-jw}, and addition
\cite{Nanda2023-mc,Quirke2024-md}. A standard
mechanistic interpretability toolkit has slowly formed
\cite{Lindsay2023-zr,Bereska2024-ry} around
visualization techniques, classifier probes, causal methods that patch
activations between circuits, and automated circuit discovery methods
\cite{Meng2023-ub,Syed2023-wa,Heimersheim2024-qt,Conmy2023-ix}.
These analyses are complemented by the explicit construction of
transformer micro-circuits with domain-specific languages
\cite{Weiss2021-ni,Zhou2023-zd,Lindner2023-px}.

\hypertarget{subsubsec-distributed-representations-and-representation-engineering}{%
\subsubsection{Distributed representations and representation
engineering}\label{subsubsec-distributed-representations-and-representation-engineering}}

Early mechanistic interpretability work was focused on a
neuron-and-circuits-centric view of artificial neural networks. Even at
this stage, however, one could see the cracks in the underlying
assumptions: convolutional neural networks for image classification
contain polysemantic units, which respond to a variety of concepts
\cite{Olah2018-er}. Polysemanticity cast doubt on the
feasibility of finding crisp explanations for single neurons, e.g. a
curve or cat detector. By contrast, the first generation of neural
networks was motivated by the idea of distributed representation as the
locus of conceptual information
\cite{Kohonen1977-ru,Hopfield1982-dm,Hinton1989-wr,Rumelhart1987-cp,Thorpe1989-eu}.
Later parallel work in computational neuroscience showed that mixed
representations were both ubiquitous in the brain and advantageous in
terms of coding capacity \cite{Rigotti2013-yd}. If
representations are distributed rather than localized, where does this
leave the project of understanding neural networks from the bottom up?

These conceptual issues became particularly relevant as the field
shifted to analyzing large-language models, where polysemanticity is the
norm \cite{Elhage2022-fp}. The exact proportion of
monosemantic units in LLMs can vary by architecture, capacity, and
criterion, but it has been estimated at anywhere between 0\%
\cite{Bricken2023-bg} and 35\%
\cite{Elhage2022-fp}. That is, most units represent a
variety of concepts and defy simple explanations. A related phenomenon
is the infeasibility of removing undesirable behavior by the ablation of
the units involved in it because of the redundancy of representations
and mechanisms \cite{McDougall2023-dr}.

Just as the field of neuroscience has moved from thinking mostly in
terms of single neurons to thinking about populations of neurons (i.e.
manifolds) over the 2010s, so has the field of mechanistic
interpretability. An example of this trend is the recent work on sparse
autoencoders (SAEs), which attempts to explain the functioning of LLMs
in terms of multiple partially overlapping representations. Activations
of neural networks are decomposed using a sparse autoencoder
\cite{Cunningham2023-yb,Bricken2023-bg}, similar to
sparse component analysis in neuroscience
\cite{Olshausen1996-ug}. The activations of the
discovered factors tend to be more interpretable than the activations of
the underlying network, as evidenced by human judges and automated
labeling and verification by other large language models
\cite{Templeton2024-br}. There is now an entire
menagerie of variants of sparse autoencoders: classic, top-k, jump,
gated, etc. with different sparseness and efficiency in
training
\cite{Gao2024-do,Rajamanoharan2024-rt,Rajamanoharan2024-xk}.
As training SAEs requires significant computational resources
\cite{Lieberum2024-vs}, it has become fashionable
to release pretrained latent factors in the form of atlases for
community-based investigation. The discovered features, labeled
automatically by other large language models, can be perused and
visualized, much like neurophysiology lab monitors are filled with
visualizations of neurons, tuning curves, and manifolds. SAE features
can also form a new basis for automated circuit discovery mechanisms
\cite{Marks2024-bh}.

An important tool for the verification of these discovered latent
factors is steering. By manipulating the activations of networks in the
(linear) direction of discovered latent factors, one can change the
output of networks toward desired ends. An evocative example of this
method is Golden Gate Claude, a version of the Claude large language
model that consistently steers conversations back to discussions of the
Golden Gate bridge \cite{Templeton2024-br}. These
methods can also potentially be used for safety applications,
suppressing undesirable behaviors such as sycophancy, repeating
imitative falsehoods, or complying with instructions to generate
dangerous, biased, or toxic outputs.

Latent feature steering is just one of a large family of techniques for
causally intervening in large language models, known as activation
patching. These allow one to measure the causal role of a set of neurons
and activations in a particular behavior. These include techniques that
graft activations from one sentence or token to another
\cite{Meng2023-ub}; measure contrasts between
two sets of related inputs to determine a steering direction
\cite{Zou2023-qo,Turner2024-ww}; and determine
potential steering dimensions through a self-consistency criterion
\cite{Burns2024-co}. Thanks to the vast amount of linear
structure in modern transformers
\cite{Marks2023-dk}, linear steering is
surprisingly potent in leading models toward desirable behaviors.

\hypertarget{subsec-evaluation-mechint}{%
\subsection{Evaluation}\label{subsec-evaluation-mechint}}

Neuroscience was and continues to be a rich source of inspiration for AI
interpretability. Neuroscience and AI interpretability have a common goal:
understanding the mechanisms by which black-box systems form
representations and display flexible, adaptive behaviors. AI
interpretability has an important role in building safer systems and
facilitating the assurance of existing AI systems. However, not every
tool in neuroscience applies to AI safety (see Box \ref{box-anns-vs-bnns}). Much of
neuroscience tooling is designed to overcome the limits of noisy and
partial recordings, building efficient estimators. In addition, some of
the most powerful tools in neuroscience interpretability are dedicated
to understanding recurrent dynamical systems, which don't have clear
analogs in many popular AI systems. It may be that many of the insights
from neuroscience have already been incorporated into the AI interpretability
toolbox, which continues to evolve on its own.

The opposite arrow of influence --- moving from AI interpretability to
neuroscience --- is underexplored and likely to benefit neuroscience. AI
interpretability is, in a sense, an ideal neuroscience \cite{Olah2021-su}: a neuroscience
where one can measure every neuron, every weight, every activation,
control the flow of information, causally mediate, edit, and ablate, all
without resorting to difficult and slow biological experiments. Finding
a minimal set of tools that have proven useful in mechanistic
interpretability and bringing them back to neuroscience, where they can
accelerate discoveries--some of which may have a potential impact on AI
safety--is a promising avenue of research, particularly when applied to digital twins~\cite{Walker2019-bx, Bashivan2019-ha}.

\hypertarget{subsec-opportunities-mechint}{%
\subsection{Opportunities}\label{subsec-opportunities-mechint}}

In summary, we highlight several promising avenues for neuroscience to
influence AI safety from the interpretability standpoint:

\begin{itemize}
\item
  Adapt interpretability methods from neuroscience to AI

  \begin{itemize}
  \item
    Focus on methods that have been seldom applied in AI
    interpretability, including dynamical systems analysis and methods from connectomics.
  \item
    Apply interpretability to model types better adapted to neuroscience
    tools, e.g. state-space models
    \cite{Dao2024-gj,Sharma2024-nf,Wang2024-du} and
    newer variants of LSTMs \cite{Beck2024-ja}.
  \end{itemize}
\item
  Build models, inspired by neuroscience, which are transparent
  by design

  \begin{itemize}
  \item
    Build modular
    \cite{Liu2023-xm,Jarvis2024-ng,Clune2013-bk},
    spatially embedded
    \cite{Achterberg2022-nc,Margalit2024-dq}, and sparse
    models \cite{Elhage2022-fp} to facilitate
    interpretation.
  \item
    Build models with learnable activation functions on edges
    \cite{Kouh2008-nh,Liu2024-kz,McFarland2013-ui}.
  \end{itemize}
\item
  Build tools to facilitate mechanistic interpretability in
  neuroscience

  \begin{itemize}
  \item
    Build tools to record from
    \cite{Urai2022-eh,Steinmetz2021-ag,Manley2024-xn,Norman2021-ib}
    and stimulate
    \cite{Bernstein2012-tv,Adesnik2021-xr,Emiliani2022-jl,Meng2020-ak}
    large populations of neurons to facilitate causal mediation analysis
    \cite{Haspel2023-zf}.
  \end{itemize}
\item
  Define new criteria for interpretability based on insights
  from cognitive science

  \begin{itemize}
  \item
    Find operationalizations anchored in cognitive science for what it
    means to be ``interpretable'', e.g. Minimum Description Length
    (MDL), Kolmogorov complexity, etc.
   \item
    Develop new benchmarks for interpretability based on insights from
    neuroscience and cognitive science
  \end{itemize}
\end{itemize}

\hypertarget{sec-discussion}{\section{Discussion}\label{sec-discussion}}
This roadmap has evaluated several promising approaches for how neuroscience could positively impact AI safety, including:

\begin{enumerate}
    \item Building digital twins of sensory systems and reverse engineering their robust representations
    \item Developing embodied digital twins through large-scale neural recordings and virtual embodiment 
    \item Pursuing biophysically detailed models through connectomics and biophysical modeling
    \item Developing better cognitive architectures
    \item Using brain data to fine-tune AI systems
    \item Training AI systems using an evolutionary curriculum
    \item Inferring loss functions from neural data and behavior
    \item Leveraging neuroscience methods for mechanistic interpretability
\end{enumerate}

Several key themes have emerged from this analysis:

\paragraph{Focus on safety over capabilities} 

Much of NeuroAI has historically been focused on increasing capabilities: creating systems that leverage reasoning, agency, embodiment, compositional representations, etc., that display adaptive behavior over a broader range of circumstances than conventional AI. We highlighted several different ways in which NeuroAI could also enhance safety without dramatically increasing capabilities. This is a promising and potentially impactful niche for NeuroAI as AI systems develop more autonomous capabilities.

\paragraph{Data and tooling bottlenecks}

Some of the most impactful ways in which neuroscience could affect AI safety are infeasible today because of a lack of tooling and data. Neuroscience is more data-rich than at any time in the past (Box \ref{box-available-neural-data}), but it remains fundamentally data-poor. Recording technologies are advancing exponentially (Box \ref{box-scaling-trends}), doubling every 5.2 years for electrophysiology and 1.6 years for imaging, but this is dwarfed by the pace of progress in AI. For example, AI compute is estimated to double every 6-10 months \cite{Sevilla2022-lc}. Being serious about neuroscience affecting AI safety will require large-scale investments in data and tooling to record neural data in animals and humans under high-entropy natural tasks, measure structure and its mapping to function, and access frontier-model-scale compute.

\paragraph{Need for theoretical frameworks} 

While we have identified promising empirical approaches, stronger theoretical frameworks are needed to understand when and why brain-inspired approaches enhance safety. This includes better understanding when robustness can be transferred from structural and functional data to AI models; the range of validity of simulations of neural systems and their ability to self-correct; and the role of evolution and curriculum in building robust behavior.

\paragraph{Breaking down research silos}

When we originally set out to write a technical overview of neuroscience for AI safety, 
we did not foresee that our work would balloon to a 100 page manuscript. What we 
found is that much of the relevant research lives in different silos: AI safety 
research has a different culture than AI research as a whole; neuroscience has 
only recently started to engage with scaling law research; structure-focused and 
representation-focused work rarely overlap, with the recent exception of structure-to-function
 enabled by connectomics \cite{Vaxenburg2024-cv,markiewicz_openneuro_2021}; insights 
 from mechanistic interpretability have yet to shape much research in neuroscience. 
 Some prescient proposals have synthesized these areas of
 research into programs, including Byrnes' brain-like AGI safety proposal 
 (Section \ref{subsubsec-brain-like-agi-safety}) and
 the framework of Sarma et al. for AGI safety based on top-down neuropsychology and bottom-up biophysical simulations \cite{Sarma2016-fx, Sarma2018-td, Sarma2019-ui}. 
 We hope to catalyze more positive exchanges between these fields by building a strong 
 common base of knowledge from which AI safety and neuroscience researchers 
 can have productive interactions.

\subsection{Moving forward: recommendations}

We've identified several distinct neuroscientific approaches which could positively impact AI safety. Some of the approaches, which are focused on building tools and data, would benefit from coordinated execution within a national lab, a focused research organization \cite{Marblestone2022-ei}, a research non-profit, or a moonshot startup. Well-targeted tools and data serve a dual purpose: a direct shot-on-goal of improving AI safety, and an indirect benefit of accelerating neuroscience research and neurotechnology translation. Projects that we identify as being good targets for a coordinated effort within the next 7 years include:

\begin{itemize}
    \item Development of high-bandwidth neural interfaces, including next-generation chronic electrophysiology and functional ultrasound imaging recording capabilities in animals and humans. We believe that decreasing the doubling time for electrophysiology capabilities to 2 years is structurally feasible provided sufficient funding and a concerted effort
    \item Large-scale naturalistic neural recordings during rich behavior in animals and humans, including the aggregation of data collected in humans in a distributed fashion
    \item Development of detailed virtual animals with bodies and environments with the aim of a shot-on-goal at the embodied Turing test \cite{Zador2022-hd}, and expansion of that environment to simulate an evolutionary curriculum
    \item Bottom-up reconstruction of circuits underlying robust behavior, including simulation of the whole mouse cortex at the point neuron level
    \item Development of multimodal foundation models for neuroscience to simulate neural activity at the level of representations and dynamics across a broad range of target species
\end{itemize}

Other approaches are focused on building knowledge and insight, and could be addressed through conventional and distributed academic research:

\begin{itemize}
    \item Improving robustness through neural data augmentation
    \item Developing better tools for mechanistic interpretability
    \item Creating benchmarks for human-aligned representation learning
\end{itemize}

Nothing about safer AI is inevitable - progress requires sustained investment and focused research effort. By thoughtfully combining insights from neuroscience with advances in AI, we can work toward systems that are more robust, interpretable, and aligned with human values. However, this field is still in its early stages. Many of the approaches we've evaluated remain speculative and will require significant advances in both neuroscience and AI to realize their potential. Success will require close collaboration between neuroscientists, AI researchers, and the broader scientific community.

Our review suggests that neuroscience has unique and valuable contributions to make to AI safety, particularly in understanding how biological systems implement robust, safe, and aligned intelligence. The challenge now is to translate these insights into practical approaches for developing safer AI systems.

\hypertarget{sec-acknowledgements}{\subsection{Acknowledgements}\label{sec-acknowledgements}}

We would like to thank reviewers who provided critical feedback on early versions of this manuscript: Ed Boyden, Bing Brunton, Milan Cvitkovic, Jan Leike, Grace Lindsay, Alex Murphy, Sumner Norman, Bence Ölveczky, Raiany Romanni, Jarod Rutledge and Paul Scotti, as well as dozens of neuroscientists and AI researchers who shaped the narrative of this manuscript. Finally, we would like to thank James Fickel for his unwavering support and vision in advancing neuroscience and AI safety.

\hypertarget{sec-methods}{%
\section{Methods}\label{sec-methods}}

\hypertarget{scaling-laws-digital-twins}{%
\subsection{Scaling laws for digital twins}\label{scaling-laws-digital-twins}}

To compile empirical scaling laws for digital twins (Figure \ref{fig-visual-scaling-laws}), we manually captured the mean test set performance reported in the main text or tables of papers; where relevant, we used \href{https://automeris.io/}{webplotdigitizer} to capture the data in graphs. In cases where correlation coefficients were reported, we squared them to obtain a measure that approximates \emph{FEVE}. Although recording length was often reported in terms of trials, repeats, and number images, we standardized everything to report recording time in seconds; for example, 12 repeats of 300 images presented in one second trials would add up to an hour. 

We fit scaling laws by minimizing the squared loss between the points and a sigmoid as a function of log recording time, using the \texttt{curve\_fit} function in \texttt{scipy.optimize}. For the V4 data, we only used the data from action potentials to fit the scaling law, as the other dataset used noisier single photon calcium imaging.

\paragraph{Simulations}

We simulated how the number of datapoints used to train the readout weights from a core affect the prediction accuracy. We generated data from a linear-nonlinear-Poisson (LNP) model:

$$\boldsymbol{\mu} = \exp(a\mathbf{Xw} + b)$$
$$\mathbf{y} \sim \textrm{Poisson}(\boldsymbol{\mu})$$

Where the weights of the model were taken to be $\mathbf{w} \sim \textrm{Normal}(0, 1/M)$, and the design matrix was iid Gaussian, $\mathbf{X} \sim \textrm{Normal}(0, 1)$. We set $a = 0.4$ and $b = 0.1$.

$\mathbf{X}$ has size $N$x$M$, where $N$ is the number of simulated trials or timepoints, and $M$ is the number of simulated readout weights. We estimated the parameters of this model using MAP with a prior matched to the weights, using the iteratively reweighted least squares algorithm. We evaluated the accuracy of the predictions on newly generated data to estimate the validated FEVE. We then fit scaling laws to this data as above.

\paragraph{Wrong core} We simulated a scenario where a neuron is driven by two components of a core: one part which is accounted for, and another part which is incorrect. For example, imagine a visual neuron in area V1 halfway between a simple cell and a complex cell: it displays some phase sensitivity as well as tolerance for the position of an oriented bar. Fitting that cell with a linear model means that no matter how much data we use to fit the model, predictions will be imperfect: the linear model cannot account for the (quadratic) nonlinearity inherent in the phase insensitivity. Thus, the mean response was modeled as:

$$\boldsymbol{\mu} = \exp(a\alpha^2 \mathbf{X}_c\mathbf{w}_c + a(1-\alpha^2) \mathbf{X}_i\mathbf{w}_i + b)$$

Since we assumed the weights $\mathbf{w}_i$ couldn't be estimated, in practice, $a(1-\alpha^2) \mathbf{X}_i\mathbf{w}_i$ reduced to a source of normal noise:

$$\boldsymbol{\mu} = \exp(a\alpha^2 \mathbf{X}_c\mathbf{w}_c + a(1-\alpha^2) \epsilon + b)$$

Where $\epsilon\sim\textrm{Normal}(0, 1)$. $\alpha$ was varied simulate scenarios varying from the core being completely incorrect ($\alpha=0$) to the core being completely correct ($\alpha=1$).

\hypertarget{simulations-of-adversarial-robustness}{%
\subsection{Simulations of adversarial
robustness}\label{simulations-of-adversarial-robustness}}

To investigate the feasibility of transferring robustness through neural
data, we took adversarially trained networks and attempted to distill them into student networks based on either their behavioral output or their internal representations. We first conducted proof-of-concept experiments using MNIST-1D, a
simplified, algorithmically generated one-dimensional digit
classification task \cite{Greydanus2020-hd} consisting of 40 element vectors. We
trained a robust teacher network (a 3-layer convolutional neural network
with 64 channels per layer) using $L_\infty$-norm adversarial
training with $\epsilon=0.3$ with a PGD attack with 50 iterations. The teacher
was trained using the Adam optimizer with an initial learning rate of
0.01 (decreased by 10x halfway through training) for 250 epochs with
data augmentation. For student training, we leveraged
MNIST-1D\textquotesingle s algorithmic generator to create multiple
training datasets with different random seeds. Students of identical
architecture were trained for 256 epochs using Adam with learning rate
0.01 (decreased by 10x halfway through training). The training
objective, $L = (1-\beta)L_{CLASS} + \beta L_{RSA}$,
combined a teacher prediction matching loss ($L_{CLASS}$) and
a representation matching one ($L_{RSA}$), with $\beta$, ranging
from 0 to 300, controlling the strength of representation matching. To
simulate training with different amounts of data while controlling for
training time, we always ran 256 epochs of 5,000 examples, varying the
number of \emph{distinct} training examples from 5,000 (labeled 1X) to
640,000 (128X); we recycled examples appropriately across epochs.

For CIFAR-10, we used a pre-trained $L_\infty$-adversarially
robust WideResNet-28-10 teacher
\cite{Wang2023-by}. Student networks of identical architecture were trained for 200 epochs
using SGD with momentum 0.9, weight decay 5e-4, an initial learning rate
of 0.01 (decreased by 10x halfway through training) and the same
training objective as in MNIST-1D experiments. Teacher features were
precomputed on clean images, while student training utilized standard
CIFAR-10 augmentation (random crops and horizontal flips). The RSA loss
was only computed over middle block group features and was rescaled
dynamically during training to maintain consistent magnitude relative to
the classification loss. To simulate neural recording noise, we added
Gaussian noise of varying magnitude (0\%, 5\%, or 10\% of the feature
magnitude) to the teacher\textquotesingle s features before computing
the RSA loss. Models were evaluated against 10-step PGD attacks with
$\epsilon=8\/255$.

\hypertarget{scaling-trends-for-neural-recordings}{%
\subsection{Scaling trends for neural
recordings}\label{scaling-trends-for-neural-recordings}}

To extend the results of \cite{Stevenson2011-sr} and \cite{Urai2022-eh}, we
retrieved all neuroscience abstracts from bioRxiv since its inception
(44,000 abstracts, cutoff date of Sep 1st, 2024) and filtered them using
an LLM (gpt-4o-2024-08-06) to obtain 513 promising papers. We used a
query that started with:

\begin{verbatim}
You are tasked with analyzing an abstract from a scientific paper to
determine if the full paper is likely to contain useful information about
state-of-the-art neural recording methods. Focus only on invasive methods, 
including penetrating electrodes, ECoG arrays, and calcium imaging.
Also consider functional ultrasound, which is sometimes referred to as
non-invasive but typically requires a craniotomy.

Here is the abstract you need to analyze:

<abstract>{{abstract}}</abstract>

Carefully read through the abstract and consider the following criteria
to determine if the paper is promising:

1. Does the abstract mention methods development--specifically methods
that enable large-scale recordings, for example new hardware or
indicators--as its primary goal?

2. Is there any indication of a large or massive dataset being recorded?

3. Does the abstract suggest that the research would only have been
possible with large datasets?

4. Are there mentions of advancements in terms of:

a) Size of the dataset
b) Number of probes used
c) Number of neurons recorded per session
d) The percentage of the brain that can be recorded, for instance
whole-brain imaging or near whole-brain imaging

If the paper meets one or more of these criteria, consider it promising.
<further instructions for formatting in JSON>
\end{verbatim}

We then fed the full text of these papers using the same LLM, using a
query that started like this:

\begin{verbatim}

You are tasked with analyzing a scientific paper to extract information
about neural recording techniques and the number of simultaneously
recorded neurons. Here\textquotesingle s how to proceed:

1. Carefully read through the following PDF content:

<pdf_content>
{{PDF_content}}
</pdf_content>

2. As you read, look for information related to:

- The species studied
- The neural recording technology used
- The number of neurons, unit, multi-units, channels, electrodes,
animals, ROIs, probes, sessions, voxels, arrays.
- Whether the recording was chronic or acute
- Any details about the recording sessions or animals used

3. Extract relevant quotes from the PDF content and list them using
<quote> tags. For example:

<quote>We recorded from 10,000 neurons across 5
mice using two-photon calcium imaging.</quote>

4. After listing the relevant quotes, analyze the information to
determine:

- The number of simultaneously recorded neurons per session
- Whether the recording was chronic or acute
- The specific neural recording technology used

<further instructions for formatting results in JSON>
\end{verbatim}

One of the authors (PM) then manually curated the results. After joining
with previous databases \cite{Stevenson2011-sr, Urai2022-eh}, we deduplicated and filtered out papers with a lower number of neurons per recording than 10 prior papers, as suggested by \cite{Stevenson2011-sr}. This process resulted in a total of 151 papers. We then fit the following Bayesian linear
regression using PyMC \cite{Abril-Pla2023-jg},
$log(\textrm{neurons}) \sim at + b$ separately for electrophysiology
and calcium imaging recorded after 1990.

\hypertarget{dimension-neural-data}{%
\subsection{Dimensionality of neural data}\label{dimension-neural-data}}

To estimate the dimensionality of neural data across model organisms, we retrieved six representative datasets: two for C. elegans \cite{Yemini2021-fd, Suzuki2024-ar}, one for larval zebrafish \cite{Chen2018-um}, and three for mice \cite{Stringer2018-gi, Stringer2021-yg, Manley2024-xn}. The data were converted to HDF5 format for analysis, and included both neural activity recordings and three-dimensional spatial coordinates of each recorded neuron, allowing us to take spatial organization of neurons into account.

For all datasets, we employed Shared Variance Component Analysis (SVCA) to estimate the number of dimensions of the data which could be reliably estimated \cite{Stringer2019-uz, Manley2024-xn}. We split the data in train and test sets that did not overlap along either time or neuron axes. SVCA then calculated the covariance between the two groups of neurons for time points belonging to the train set, applied randomized singular value decomposition (SVD) to calculate orthogonal bases for both sets of neural activity, and assessed how well this basis explained the variance in the test set. This amount of variance, normalized by the total variance, was termed reliable variance, and dimensions were considered significantly reliable only if their percentage of reliable variance was greater than four standard deviations above the mean of the reliable variance for two shuffled datasets, following \cite{Manley2024-xn}.

To assess how dimensionality scaled with the number of recorded neurons, we randomly sampled subsets of neurons from each dataset, z-scored their activity, and applied all three methods described above. The number of neurons sampled was varied either linearly or logarithmically, depending on the dataset’s total neuron count. To prevent spurious correlations, neurons were divided into spatially alternating bins before splitting into training and testing sets. We then fit a linear regression in log-log space to estimate the exponent of a scaling law for neural dimensionality as a function of the recording size, $\textrm{dim}\sim \alpha (\# \text{neurons})^\beta$.

\hypertarget{estimating-free-data}{%
\subsection{Estimating freely available data}\label{estimating-free-data}}

To estimate the total amount of freely available neural data, we analyzed datasets available on three major data repositories: DANDI \cite{halchenko_dandi/dandi-cli:_2024}, OpenNeuro \cite{markiewicz_openneuro_2021} and \href{https://ieeg.org}{iEEG.org}. For iEEG, we used a script to scrape recording length information from the HTML source of website. For DANDI datasets, we used the DANDI Python client to access Neurodata Without Borders \cite{rubel_neurodata_2022} files, leveraging partial downloads to minimize data transfer. Extracted metadata included recording modality (e.g. calcium imaging or electrophysiology), subject species, number of recorded neurons, and recording duration for each file. The resulting data was then processed to standardize species and recording technology nomenclature, as well as to compute subject- and dataset-level statistics.

For OpenNeuro datasets, we used the OpenNeuro GraphQL API endpoint to identify and analyze datasets containing neuroimaging data. We then retrieved session metadata for BIDS-formatted \cite{gorgolewski_brain_2016} recordings, extracting metadata such as recording modality (e.g. MRI, EEG, MEG etc) and number of subjects, using the accompanying event files to estimate the recording's duration. The resulting data was then filtered to exclude likely artifacts, such as single sessions lasting more than four hours, and to compute dataset-level statistics.

Given their outsized contribution to the total amount of available data, we manually validated the largest datasets in each repository, cross-referencing them against published papers, technical documentation and code repositories when available, and manually correcting them in case of discrepancies. Since many large-scale neuroimaging datasets are not hosted on these repositories, we supplemented this data with the UK Biobank \cite{miller_multimodal_2016}, the 1000 Functional Connectomes Project \cite{kalcher_fully_2012}, the Human Connectome Project \cite{van_essen_human_2012}, the Adolescent Brain Cognitive Development Study \cite{casey_adolescent_2018}, the Healthy Brain Network \cite{alexander_open_2017}, the Natural Scenes Dataset \cite{Allen2022-bq}, the Courtois NeuroMod project \cite{boyle_courtois_2023, seeliger_large_2019,siegel_psilocybin_2024}, incorporating this information into our analysis.

\hypertarget{trends-brain-score}{%
\subsection{Trends in brain score}\label{trends-brain-score}}

We pulled the results of the vision leaderboard from \href{https://brainscore.org}{brainscore.org} using a script. We filtered the data on those models for which we had scores on 9 neural datasets: one from V1, one from V2, three from V4, and four from IT. The submission dates for the models are not listed, and we instead used the order in which each model was submitted as a proxy for the date of submission. We used seaborn to plot the results, using a quadratic model to fit the temporal trends separately.

\printbibliography

\section{Appendix}

\subsection{Log-sigmoid scaling laws}

We were surprised to find log-sigmoid scaling curves for digital in both empirical data and simulations; to the best of our knowledge, this has not been reported in the literature. We thus searched for a regime where a log-sigmoid scaling law follows naturally from asymptotics.

Consider a linear regression model--similar to the LNP model considered in the main text (Section \ref{scaling-laws-digital-twins}), but more analytically tractable:
\[
y = X w + \epsilon
\]
where:
\begin{itemize}
    \item \( X \in \mathbb{R}^{N \times M} \) with entries \( X_{ij} \sim \mathcal{N}(0, 1) \).
    \item \( w \sim \mathcal{N}\left(0, \frac{1}{M} I_M\right) \).
    \item \( \epsilon \sim \mathcal{N}\left(0, \sigma_N^2 I_N\right) \).
    \item The validation data \( X_{\text{val}} \in \mathbb{R}^{N_{\text{val}} \times M} \) with entries \( X_{\text{val}, ij} \sim \mathcal{N}(0, 1) \).
\end{itemize}

Note that with this definition, $\text{var}(y) = 1 + \sigma_N^2$. We use a Maximum A Posteriori (MAP) estimate of \( w \) with a prior matching its sampling distribution:
\[
w_{\text{MAP}} = \left( X^\top X + M \sigma_N^2 I_M \right)^{-1} X^\top y
\]

The Fraction of Explained Variance Estimator (FEVE) on the validation set is defined as:
\[
\text{FEVE} = 1 - \frac{\mathbb{E}\left[ \left( y_{\text{val}} - X_{\text{val}} w_{\text{MAP}} \right)^2 \right] - \sigma_N^2}{\mathbb{E}\left[ y_{\text{val}}^2 \right] - \sigma_N^2}
\]
where \( y_{\text{val}} = X_{\text{val}} w + \epsilon_{\text{val}} \).

Equivalently, the FEVE is the $R^2$ on the validation set if there's no noise in the validation set.

The denominator of FEVE is:
\[
\mathbb{E}\left[ y_{\text{val}}^2 \right] - \sigma_N^2 = 1
\]

Similarly, the numerator is:

\[
\mathbb{E}\left[ \left( X_{\text{val}} \left( w - w_{\text{MAP}} \right) \right)^2 \right]
\]

Approximating \( X_{\text{val}}^\top X_{\text{val}} \approx N_{\text{val}} I_M \) (since entries are i.i.d. standard normal), the design matrix falls out, and we find that the numerator is:

\[
\approx \mathbb{E}\left[ \left\| w - w_{\text{MAP}} \right\|^2 \right]
\]

The posterior covariance matrix is:
\[
\Sigma_{\text{post}} = \left( X^\top X + M \sigma_N^2 I_M \right)^{-1} M \sigma_N^2 I_M
\]
Again using the expected value of the covariance matrix, we find that:
\[
\Sigma_{\text{post}} \approx  \left( N I_M + M \sigma_N^2 I_M \right)^{-1} M \sigma_N^2 I_M = \frac{M \sigma_N^2}{N + M \sigma_N^2} I_M
\]
Therefore:
\[
\mathbb{E}\left[ \left\| w - w_{\text{MAP}} \right\|^2 \right] = \operatorname{Tr}\left( \Sigma_{\text{post}} \right) = \frac{M^2 \sigma_N^2}{N + M \sigma_N^2}
\]

Finally, the FEVE is:
\[
\text{FEVE} \approx  \frac{ N }{ N + M \sigma_N^2 }
\]

We note that this can be expressed as the log sigmoid:

$$\text{FEVE} = \sigma(\log(N) - \log(M) - \log(\sigma^2_N))$$

Thus, under the assumptions we use, the scaling law for linear regression is a log-sigmoid. We note that our expression is valid when $M > N$ and potentially high noise, the so-called classic regime. By contrast, Canatar et al. (2023) \cite{Canatar2023-ya} tackle a different regime where there are far more regressors than observations, $N > M$, and no noise, the so-called modern or interpolation regime.

We believe that the scenario we consider is a better reflection of practices in digital twins, as opposed as the analysis in \cite{Canatar2023-ya}, which better reflects practices in comparing neural networks and brains with task-driven neural networks. Neural data is noisy, and in the limit of high observation noise, we don't always get better fits in the interpolation regime. It is common for digital twins to be fit in the classic regime; for example, Lurz et al. (2021) \cite{Lurz2020-cs} have fewer than a hundred parameters in their readout, but thousands of observations. Future work should address how these two frameworks can be merged to cover the full range of scenarios in which digital twins and task-driven neural networks are fit.

\end{document}